\documentclass[sigconf,nonacm]{cea-preprint}
\DeclareUnicodeCharacter{266B}{\mbox{\fontencoding{U}\fontfamily{wasy}\selectfont\char15}}
\usepackage{amsmath,mathtools}
\usepackage{graphicx,booktabs,colortbl,multirow,tabularx}
\usepackage{float,placeins}
\newcommand{\appref}[1]{\hyperref[app:#1]{Appendix~\ref*{app:#1}}}
\newcommand{\apprefs}[2]{Appendices~\ref{app:#1} and~\ref{app:#2}}
\ifhmode\space\fi
\definecolor{openairow}{HTML}{EAF4EE}
\definecolor{googlerow}{HTML}{EAF1FC}
\definecolor{anthropicrow}{HTML}{FFF0E5}
\definecolor{qwenrow}{HTML}{F1EAFB}
\definecolor{mistralrow}{HTML}{FFF7D9}
\definecolor{methodrow}{HTML}{E3F3F3}
\definecolor{ink}{HTML}{223244}
\newcommand{\Self}{\mathrm{SR}}
\newcommand{\Mas}{\mathrm{MAS}}
\newcommand{\Enon}{E_\Delta^{\mathrm{non}}}
\newcommand{\Arel}{A_{\mathrm{rel}}}

\newcommand{\R}{\mathcal{R}}
\newcommand{\Sset}{\mathcal{S}}
\graphicspath{{figures/}}

\definecolor{fullrow}{HTML}{EAF1FC}
\definecolor{transferrow}{HTML}{E3F3F3}
\definecolor{softrow}{HTML}{FFF6DD}
\definecolor{hardrow}{HTML}{FBEDE7}

\usepackage{pifont}
\definecolor{scopeyes}{HTML}{238443}
\definecolor{scopeno}{HTML}{C43C39}
\definecolor{tablegray}{HTML}{F1F2F3}
\newcommand{\scopeyes}{\textcolor{scopeyes}{\ding{51}}}
\newcommand{\scopeno}{\textcolor{scopeno}{\ding{55}}}

\definecolor{topologyband}{HTML}{E8F0FA}
\definecolor{factband}{HTML}{E8F5ED}
\definecolor{referenceband}{HTML}{F1F2F3}
\definecolor{relatedband}{HTML}{F4F6F8}
\definecolor{oursband}{HTML}{E8F5ED}

\definecolor{sectionband}{HTML}{F4F4F4}
\definecolor{correlationband}{HTML}{F4F4F4}
\definecolor{recoveryband}{HTML}{F4F4F4}
\definecolor{taskband}{HTML}{F4F4F4}
\definecolor{pooledband}{HTML}{F4F4F4}
\definecolor{homoband}{HTML}{F4F4F4}
\definecolor{heteroband}{HTML}{F4F4F4}
\newcommand{\datatablefont}{\normalsize}

\newcommand{\vbest}[1]{\textbf{#1}}
\newcommand{\vsecond}[1]{\underline{#1}}
\newcommand{\vnegative}[1]{#1}
\newcommand{\vband}[3]{\multicolumn{#1}{@{}l@{}}{\strut\textit{#3}}\\}
\newcommand{\vhead}[1]{\shortstack{#1}}

\newcommand{\mainband}[3]{\rowcolor{#2}[0pt][0pt]\multicolumn{#1}{@{}l@{}}{\strut\textit{#3}}\\}

\usepackage{tikz}
\definecolor{tierW}{HTML}{D69532}
\definecolor{tierM}{HTML}{377FB5}
\definecolor{tierS}{HTML}{8062A6}
\DeclareRobustCommand{\tierdisc}[2]{%
  \tikz[baseline=(tier.base)]{\node[circle,draw=tier#1,line width=.4pt,
    fill=tier#1!22,inner sep=0pt,minimum size=9.3pt,text=black] (tier)
    {\fontsize{5.7}{6}\selectfont\bfseries #1};}%
  \if\relax\detokenize{#2}\relax\else\textsubscript{\scriptsize #2}\fi}
\newcommand{\tierlegend}{\tierdisc{W}{} weak\quad\tierdisc{M}{} medium\quad\tierdisc{S}{} strong}

\ifhmode\space\fi
\definecolor{covblue}{HTML}{4E79A7}
\definecolor{f1green}{HTML}{59A14F}
\definecolor{emergeorange}{HTML}{F28E2B}
\definecolor{deltapurple}{HTML}{9C6ADE}
\definecolor{peerred}{HTML}{E76F51}

\definecolor{softgray}{HTML}{EEF0F2}
\definecolor{zerogray}{HTML}{8B8F94}
\definecolor{negred}{HTML}{C94C4C}

\definecolor{closedband}{HTML}{E9E6F2}
\definecolor{closedstripe}{HTML}{F7F5FA}
\definecolor{closedtext}{HTML}{665A7A}

\definecolor{openband}{HTML}{DDEFE8}
\definecolor{openstripe}{HTML}{F3FAF7}
\definecolor{opentext}{HTML}{2E6F5B}

\newcommand{\miniposbar}[4]{%
\begin{tikzpicture}[baseline=(num.base)]
  \def\BW{0.82}\def\BH{0.12}
  \path[use as bounding box] (0,-0.10) rectangle (2.00,0.23);
  \pgfmathsetmacro{\ratio}{min(max(#1/#2,0),1)}
  \fill[softgray,rounded corners=0.45pt] (0,0) rectangle (\BW,\BH);
  \fill[#3!78,rounded corners=0.45pt] (0,0) rectangle ({\BW*\ratio},\BH);
  \node[anchor=east,inner sep=0pt,font=\normalsize] (num) at (2.00,0.06) {#4};
\end{tikzpicture}%
}
\newcommand{\minidivbar}[4]{%
\begin{tikzpicture}[baseline=(num.base)]
  \def\BW{0.82}\def\BH{0.12}\def\MID{0.41}\def\HALF{0.39}
  \path[use as bounding box] (0,-0.10) rectangle (2.00,0.23);
  \fill[softgray,rounded corners=0.45pt] (0,0) rectangle (\BW,\BH);
  \pgfmathsetmacro{\ratio}{min(abs(#1)/#2,1)}
  \pgfmathparse{#1 < 0}
  \ifnum\pgfmathresult=1
    \fill[negred!78] ({\MID-\HALF*\ratio},0) rectangle (\MID,\BH);
  \else
    \fill[#3!78] (\MID,0) rectangle ({\MID+\HALF*\ratio},\BH);
  \fi
  \draw[zerogray!75,line width=0.25pt] (\MID,-0.01)--(\MID,\BH+0.01);
  \node[anchor=east,inner sep=0pt,font=\normalsize] (num) at (2.00,0.06) {#4};
\end{tikzpicture}%
}
\newcommand{\finalpair}[6]{%
  \shortstack[c]{\miniposbar{#1}{#2}{covblue}{#3}\\[-2.5pt]
                 \miniposbar{#4}{#5}{f1green}{#6}}%
}
\newcommand{\effectpair}[6]{%
  \shortstack[c]{\minidivbar{#1}{#2}{emergeorange}{#3}\\[-2.5pt]
                 \minidivbar{#4}{#5}{deltapurple}{#6}}%
}
\newcommand{\peerbar}[3]{%
\begin{tikzpicture}[baseline=(num.base)]
  \def\BW{0.95}\def\BH{0.13}
  \path[use as bounding box] (0,-0.10) rectangle (2.00,0.23);
  \pgfmathsetmacro{\ratio}{min(max(#1/#2,0),1)}
  \fill[softgray,rounded corners=0.45pt] (0,0) rectangle (\BW,\BH);
  \fill[peerred!76,rounded corners=0.45pt] (0,0) rectangle ({\BW*\ratio},\BH);
  \node[anchor=east,inner sep=0pt,font=\normalsize] (num) at (2.00,0.065) {#3};
\end{tikzpicture}%
}
\ifhmode\space\fi
\setcopyright{none}
\acmSubmissionID{}
\submissionType{}
\fancypagestyle{preprint}{%
  \fancyhf{}
  \fancyhead[L]{\small\itshape Preprint}
  \fancyhead[R]{\small Collective Emergent Alignment}
  \fancyfoot[C]{\small\thepage}

}

\ifhmode\space\fi

\title[Collective Emergent Alignment]{Collective Emergent Alignment: Measuring and Amplifying Interaction-Induced Reasoning Alignment in Multi-Agent LLMs}
\author{Boxuan Wang}
\affiliation{\institution{University of Liverpool}\city{Liverpool}\country{United Kingdom}}
\author{Zhuoyun Li}
\affiliation{\institution{University of Liverpool}\city{Liverpool}\country{United Kingdom}}
\author{Sarvapali D. Ramchurn}
\affiliation{\institution{University of Southampton}\city{Southampton}\country{United Kingdom}}
\author{Xiaowei Huang}
\affiliation{\institution{University of Liverpool}\city{Liverpool}\country{United Kingdom}}
\author{Yi Dong}
\authornote{Corresponding author: Yi Dong (\href{mailto:yi.dong@liverpool.ac.uk}{yi.dong@liverpool.ac.uk}).}
\email{yi.dong@liverpool.ac.uk}
\affiliation{\institution{University of Liverpool}\city{Liverpool}\country{United Kingdom}}
\renewcommand{\shortauthors}{Wang et al.}

\makeatletter
\renewcommand{\@mkauthors@iii}{%
  \global\setbox\mktitle@bx=\vbox{%
    \noindent\unvbox\mktitle@bx\par\medskip
    \centering
    {\fontsize{11}{14}\selectfont\sffamily
      Boxuan Wang\textsuperscript{1}\quad
      Zhuoyun Li\textsuperscript{1}\quad
      Sarvapali D. Ramchurn\textsuperscript{2}\quad
      Xiaowei Huang\textsuperscript{1}\quad
      Yi Dong\textsuperscript{1,*}\par}
    \vspace{5pt}
    {\small\normalfont
      \textsuperscript{1}School of Computer Science and Informatics, University of Liverpool, UK\par
      \textsuperscript{2}School of Electronics and Computer Science, University of Southampton, UK\par}
    \medskip}}
\makeatother

\begin{abstract}
Multi-agent large language model (LLM) systems can exhibit emergent behaviors through interaction, yet whether a similar phenomenon exists for \emph{reasoning alignment} remains largely unexplored. We study whether agent interaction produces additional reference-grounded reasoning alignment beyond matched individual self-refinement, and introduce \textbf{REMAP} (\textbf{R}eference--\textbf{E}ntailment--\textbf{M}atrix for \textbf{A}gentic \textbf{P}rocess), a framework that quantifies this effect through Semantic Coverage and interaction-specific alignment gains. Across diverse model families and two multi-hop reasoning benchmarks, we observe broadly positive interaction gains, particularly on reasoning content that depends on integrating evidence distributed across agents. These gains are consistently associated with independently measured peer-information recovery, while further analysis reveals composition-dependent heterogeneity effects and asymmetric information transfer across agent capability levels. Motivated by these findings, we further introduce a fact-level multi-agent communication framework that replaces full reasoning-trace exchange with compact, gap-aware, and selectively routed information. Our experiments show that compact fact exchange improves reasoning alignment and task performance while reducing communication overhead, whereas more selective communication exposes a trade-off between acquiring distributed information, retaining local content, and communication efficiency. Overall, our results show that collective emergent alignment can be quantified, linked to information-transfer patterns, and used to guide more effective multi-agent communication.
\ifhmode\space\fi
\end{abstract}

\usepackage{listings}
\lstdefinestyle{ceaprompt}{
  basicstyle=\ttfamily\footnotesize,
  columns=fullflexible, keepspaces=true,
  breaklines=true, breakatwhitespace=false,
  showstringspaces=false, frame=single,
  framesep=4pt, xleftmargin=4pt, xrightmargin=4pt,
  aboveskip=5pt, belowskip=6pt,
  rulecolor=\color{black!35}
}
\hypersetup{hidelinks,keeppdfinfo}
\begin{document}
\flushbottom
\setcounter{dbltopnumber}{1}
\maketitle
\hypersetup{pdfauthor={Boxuan Wang, Zhuoyun Li, Sarvapali D. Ramchurn, Xiaowei Huang, Yi Dong},pdfkeywords={Multi-agent LLMs, alignment, reasoning, emergent behavior, evaluation}}
\pagestyle{preprint}
\thispagestyle{preprint}


\begin{figure}[t]
\centering
\includegraphics[width=\columnwidth]{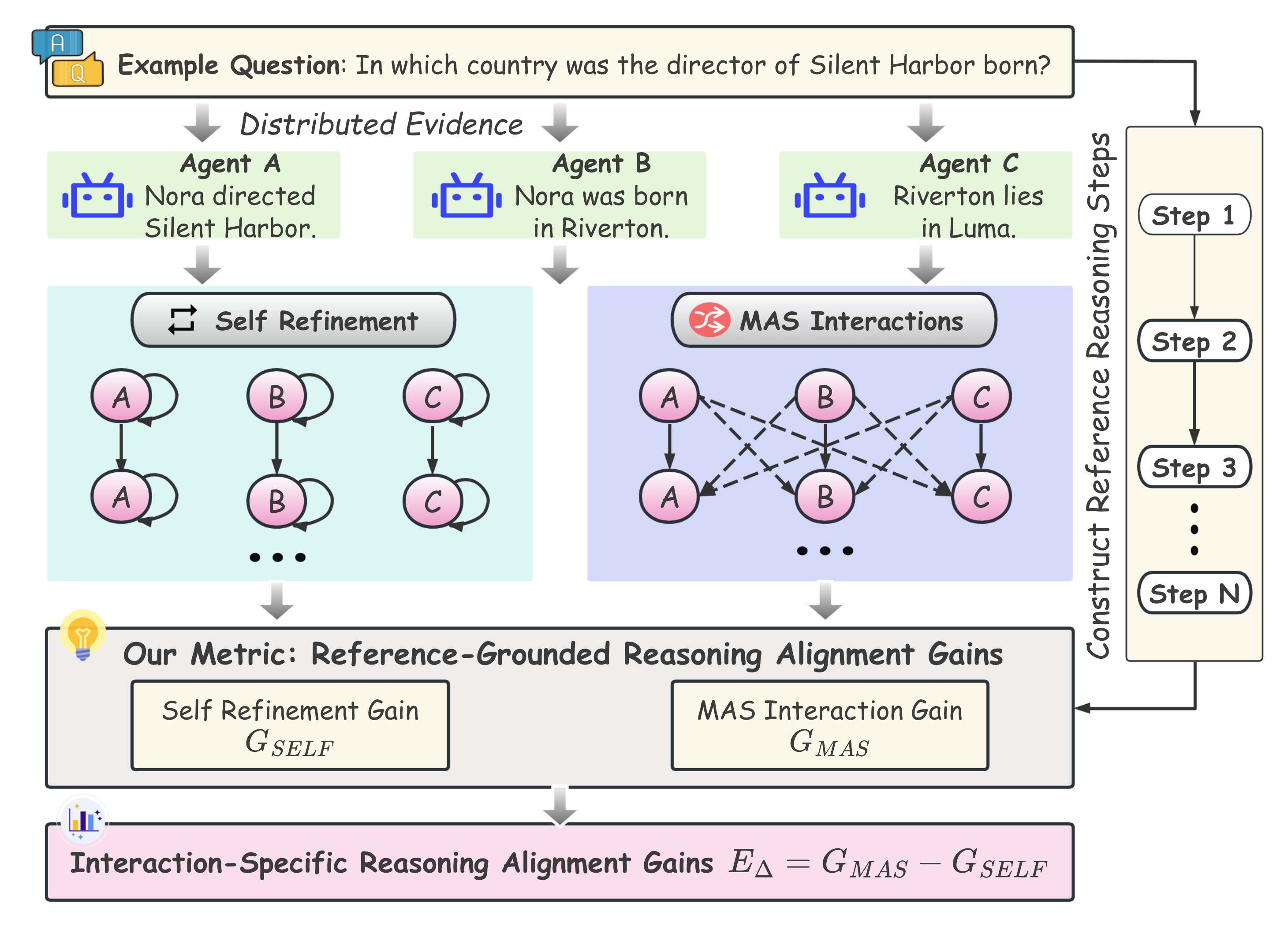}
\caption{\textbf{Measuring collective emergent alignment.} Agents with distributed evidence follow matched self-refinement and MAS trajectories. Their reference-grounded reasoning gains quantify the contribution of interaction.}
\label{fig:concept}
\Description{Three agents receive complementary facts about a director, a birthplace, and a country. They follow parallel self-refinement and multi-agent interaction trajectories. A common reference supplies the reasoning steps for evaluating both trajectories. Subtracting their gains yields interaction-specific reasoning alignment.}
\end{figure}
\ifhmode\space\fi
\section{Introduction}
\label{sec:introduction}

Large language model (LLM) agents increasingly work in multi-agent systems (MAS)~\cite{wu2023autogen,hong2024metagpt}, where they exchange information, discuss candidate solutions, and revise their reasoning~\cite{li2023camel,qian2024chatdev}. A central motivation is \emph{emergent collective behavior}: interaction can help a group combine partial information into outcomes that isolated agents may not reach~\cite{chen2023agentverse}. Previous work has observed emergent coordination in generative-agent simulations~\cite{park2023generative} and improved reasoning and factual accuracy through multi-agent debate~\cite{du2024debate,liang2024debate}. These findings motivate a closer look at how the interaction changes reasoning and whether the agentic process contributes to emergent behavior in reasoning alignment.

Reasoning alignment and final-answer quality capture complementary aspects of collaboration~\cite{smit2024mad,wang2026consistency,turpin2023unfaithful}. Gains may also come from voting~\cite{li2024more}, repeated sampling~\cite{wang2023selfconsistency}, or individual refinement~\cite{madaan2023selfrefine,shinn2023reflexion}. Existing reasoning-alignment metrics mainly evaluate individual traces~\cite{golovneva2023roscoe,prasad2023receval,xia2024reasoneval}, while MAS evaluation commonly focuses on task accuracy, consensus, or communication efficiency~\cite{zhu2025multiagentbench,li2024sparse}. A quantitative gap remains: how can we determine whether interaction adds reference-aligned reasoning beyond what the same agents obtain through independent self-refinement? We define this additional, reference-grounded gain in reasoning alignment as \textbf{collective emergent alignment}.

As shown in Figure~\ref{fig:concept}, we introduce \textbf{REMAP}, a Reference Entailment Matrix framework for Agentic Process to quantify this effect. REMAP constructs a bidirectional natural language inference (NLI) matrix between reference units and the reasoning units generated by each agent. Each reference unit contributes its best generated match, yielding a \emph{Semantic Coverage} score $C$. The matched initial-to-final coverage changes give the interaction-specific contrast:
\[
E_\Delta = G_{\mathrm{MAS}}-G_{\mathrm{SELF}},
\qquad G_X=C_{\mathrm{final}}^X-C_{\mathrm{init}}^X.
\]
A positive $E_\Delta$ means that MAS gains more reference coverage, or loses less, than matched self-refinement. An ownership-aware variant, $\Enon$, evaluates only reference units supported by evidence initially assigned to peers. This separates changes on peer-supported information from changes on an agent's local content. We validate the measurements against answer quality and deterministic gold-item recovery computed independently of the NLI scorer.

REMAP then connects measurement to the design of communication. We examine peer-information recovery, heterogeneous composition, and owner--receiver directionality. The analyses show associations between collective gains and recovered peer information, together with composition-dependent gains and exploratory directional differences. These observations motivate a fact-level MAS framework that organizes communication around the information available to each receiver. 

We organize the study around three research questions:
\begin{itemize}
    \item \textbf{RQ1:} How can we quantify whether reference-grounded reasoning alignment emerges from multi-agent interaction beyond matched self-refinement?
    \item \textbf{RQ2:} What information-transfer patterns are associated with collective emergent alignment, particularly peer information recovery, heterogeneous/homogeneous composition, and owner--receiver directionality?
    \item \textbf{RQ3:} Can these findings guide more effective fact-level communication that improves reasoning alignment and task performance while reducing communication overhead?
\end{itemize}

Our contributions are threefold: \textbf{(1)} We introduce \textbf{REMAP}, a reference-entailment-matrix framework for measuring reasoning alignment throughout agent trajectories. \textbf{(2)} We empirically characterize collective emergent alignment across seven model families and two multi-hop reasoning benchmarks. \textbf{(3)} We introduce and evaluate a fact-level communication framework guided by these findings, connecting peer-information acquisition, task performance, and communication efficiency.
\ifhmode\space\fi
\section{Related Work}\label{sec:related}
\noindent\textbf{Evaluating Reasoning Alignment.} 
The Structured Alignment Score~\cite{wang2026cotlens} compares generated and reference reasoning through semantic-relation matrices. ROSCOE~\cite{golovneva2023roscoe} includes reference-based and reference-free diagnostics. ReCEval~\cite{prasad2023receval} evaluates step correctness and informativeness using claims, prior steps, and context. REVEAL~\cite{jacovi2024reveal} benchmarks relevance, evidence attribution, and logical correctness. The NLI-hypergraph audit of Kohli et al.~\cite{kohli2026referencefree} checks how support relations compose within a trace. REMAP builds on reference-based evaluation and applies it separately to each agent's states along matched self-refinement and MAS trajectories.

\noindent\textbf{Evaluating Multi-Agent Reasoning.} 
MultiAgentBench~\cite{zhu2025multiagentbench} evaluates milestone progress and coordination. Who\&When~\cite{zhang2025whowhen} locates failure-responsible agents and steps, while MAST~\cite{cemri2025mast} classifies system-design, inter-agent, and verification failures. CARA assesses inter-agent consistency. Its CARA-GT variant measures expert-reference alignment and coverage across debate rounds~\cite{wang2026consistency}. Together, these studies support process-level analysis~\cite{zhu2026raffles}, including reference-based MAS evaluation.

Table~\ref{tab:related-positioning} compares six capabilities: reference-based evaluation (Ref.), multi-agent evaluation (MAS), matched self-refinement (Self), evidence ownership (Own.), quality evaluation (Qual.), and emergent alignment (Emerg.). REMAP combines these capabilities to quantify interaction gains and distinguish acquired peer information from retained local content. Its matched trajectory comparison isolates interaction-associated alignment gains beyond individual refinement. The ownership-aware decomposition further reveals how agents acquire missing evidence and preserve existing reasoning, providing actionable insights for MAS communication design.

\begin{table}[!htb]

\caption{\textbf{Positioning of REMAP.}
Checks indicate capabilities supported by each method.
SAS: Structured Alignment Score.
NLI-HG: NLI-hypergraph.
MAB: MultiAgentBench.
W\&W: Who\&When.
ROSCOE refers to its reference-based variants.}

\label{tab:related-positioning}
\centering
\footnotesize
\setlength{\tabcolsep}{2.0pt}
\renewcommand{\arraystretch}{1.10}

\begin{tabularx}{\columnwidth}{@{}>{\raggedright\arraybackslash}Xcccccc@{}}
\toprule
\textbf{Method/Framework} &
\textbf{Ref.} &
\textbf{MAS} &
\textbf{Self} &
\textbf{Own.} &
\textbf{Qual.} &
\textbf{Emerg.}\\
\midrule

\mainband{7}{relatedband}{Reasoning-trace evaluation}

SAS~\cite{wang2026cotlens}
& \scopeyes & \scopeno & \scopeno & \scopeno & \scopeyes & \scopeno\\

ROSCOE~\cite{golovneva2023roscoe}
& \scopeyes & \scopeno & \scopeno & \scopeno & \scopeyes & \scopeno\\

ReCEval~\cite{prasad2023receval}
& \scopeno & \scopeno & \scopeno & \scopeno & \scopeyes & \scopeno\\

REVEAL~\cite{jacovi2024reveal}
& \scopeno & \scopeno & \scopeno & \scopeno & \scopeyes & \scopeno\\

NLI-HG~\cite{kohli2026referencefree}
& \scopeno & \scopeno & \scopeno & \scopeno & \scopeyes & \scopeno\\

\midrule
\mainband{7}{topologyband}{Multi-agent evaluation}

MAB~\cite{zhu2025multiagentbench}
& \scopeno & \scopeyes & \scopeno & \scopeno & \scopeyes & \scopeno\\

W\&W~\cite{zhang2025whowhen}
& \scopeno & \scopeyes & \scopeno & \scopeno & \scopeyes & \scopeno\\

MAST~\cite{cemri2025mast}
& \scopeno & \scopeyes & \scopeno & \scopeno & \scopeyes & \scopeno\\

CARA-GT~\cite{wang2026consistency}
& \scopeyes & \scopeyes & \scopeno & \scopeno & \scopeyes & \scopeno\\

\midrule
\rowcolor{oursband}
\textbf{REMAP (Ours)}
& \scopeyes & \scopeyes & \scopeyes & \scopeyes & \scopeyes & \scopeyes\\

\bottomrule
\end{tabularx}
\end{table}

\ifhmode\space\fi

\ifhmode\space\fi
\section{Methodology}
\label{sec:method}
REMAP scores each agent's response against the same question reference, then averages the agent scores and compares matched self-refinement and MAS trajectories. Figure~\ref{fig:calculation} shows these steps.

\subsection{Preliminaries}
REMAP builds on the reference-based NLI approach of the Structured Alignment Score~\cite{wang2026cotlens}. It compares two types of units:
\begin{figure*}[t]
\centering
\includegraphics[width=0.96\textwidth]{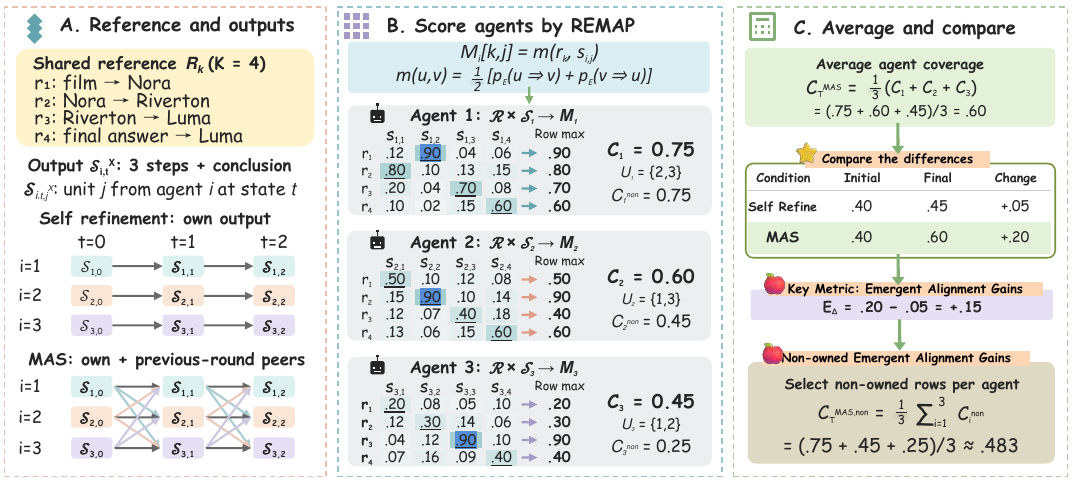}

\caption{\textbf{Computing collective alignment gains.}
Left: a fixed reference $\R$ and agent outputs $\Sset_{i,t}^{X}$ under self-refinement and MAS.
Middle: row maxima of each NLI matching matrix yield agent coverage $C_{i,t}^{X}$. The set $U_i$ indexes reference facts not initially owned by agent $i$ (e.g., $U_1=\{2,3\}$).
Right: average agent scores and contrast initial-to-final changes to obtain $E_\Delta$. Non-owned coverage averages rows indexed by $U_i$, excluding the conclusion. The matrices illustrate the final MAS state.}

\label{fig:calculation}
\Description{A shared four-unit reference is compared separately with the four generated units from each of three agents. Each matrix uses bidirectional entailment for all sixteen pairs. Agent 1 matches r1 to its second generated unit and r2 to its first. The three row-maximum vectors are (.90,.80,.70,.60), (.50,.90,.40,.60), and (.20,.30,.90,.40), giving agent coverage .75,.60,.45 and mean .60. Non-owned sets are {2,3}, {1,3}, and {1,2}. Their agent scores .75,.45,.25 average to .483. Matched temporal changes give overall E Delta .15.}
\end{figure*}

\noindent\textbf{(1) Reference units $r_k$.}
For a question $q$, one fixed reference $\R=(r_1,\ldots,r_K)$ is shared by all agents, states, and conditions. Its first $L$ units express the annotated intermediate answers or evidence triples. The last unit $r_K$ states the final gold conclusion, so $K=L+1$. MuSiQue units are built from resolved subquestions and gold subanswers. Corrected 2Wiki units use rendered evidence triples. References are constructed deterministically based on the questions and used only for offline evaluation. Full details of the reference construction are provided in \appref{protocol}.

\noindent\textbf{(2) Generated units $s_{i,t,j}^{X}$.}
Agents are indexed by $i\in\{1,\ldots,N\}$, states by $t\in\{0,\ldots,T\}$, and conditions by $X\in\{\Self,\Mas\}$, where $\Self$ denotes self refinement. Agent $i$'s current scored output is $\Sset_{i,t}^{X}=(s_{i,t,1}^{X},\ldots,s_{i,t,J}^{X})$, where $j$ indexes a unit within that response and $J=|\Sset_{i,t}^{X}|$. Scored units comprise the emitted reasoning steps and conclusion. Local-fact metadata, memory, gaps, and the separate answer field serve the communication and answer-aggregation procedures. The main protocol requests three steps plus a conclusion, giving four scored units.

\subsection{Semantic Coverage}
For each agent and state, compare \emph{every} reference unit with \emph{every} generated unit. This gives a separate $K\times J$ matrix $M_{i,t}^{X}$, whose entry can be described as:
\begin{equation}
\begin{split}
 M_{i,t}^{X}[k,j]&=m(r_k,s_{i,t,j}^{X}),\\
 m(u,v)&=\tfrac12[p_E(u\Rightarrow v)+p_E(v\Rightarrow u)].
\end{split}
\label{eq:match}
\end{equation}
Here $p_E(u\Rightarrow v)$ is the NLI entailment probability for premise $u$ and hypothesis $v$. Each reference row $k$ retains its best match in the scored agent response:
\begin{equation}
 c_{ik,t}^{X}=\max_{1\le j\le J}M_{i,t}^{X}[k,j].
\label{eq:rowmax}
\end{equation}
Matching follows semantic content: $r_1$ may match $s_{i,t,2}^{X}$, and one generated unit may cover several reference rows. Each row selects its match independently.

Average the row maxima per agent and then across agents:
\begin{equation}
 C_{i,t}^{X}=\frac1K\sum_{k=1}^{K}c_{ik,t}^{X},\qquad
 C_t^{X}=\frac1N\sum_{i=1}^{N}C_{i,t}^{X}.
\label{eq:coverage}
\end{equation}
Thus three agents produce three matrices and three agent scores at each state. Their arithmetic mean is the system score, in $[0,1]$. Each agent contributes equally, preserving the attribution of reference content to individual responses. Full worked examples are provided in \appref{formal}.

\noindent\textbf{What the system average measures.}
The team may initially hold all source facts collectively, while each response covers only part of the reference. When peer information appears in a receiver's reasoning, its row maximum can increase. The fixed reference tracks the same content across agents and states, and equal weighting measures its spread through the team. Answer voting gives the separate task-quality measures.

\noindent\textbf{Score properties and computation.}
Permuting generated units or adding an exact duplicate preserves every row maximum. The fixed output schema controls the number and type of scored units. Each agent-state matrix requires at most $2KJ$ directional NLI evaluations, with repeated text pairs retrieved from a cache. Overall and ownership scores reuse this matrix. A uniform perturbation of at most $\varepsilon$ to local matches changes coverage by at most $\varepsilon$ and the interaction contrast by at most $4\varepsilon$. This bound describes sensitivity to a specified matrix perturbation. Full proofs and caching details are provided in \appref{formal}.

\begin{figure*}[t]
\centering
\includegraphics[width=\textwidth]{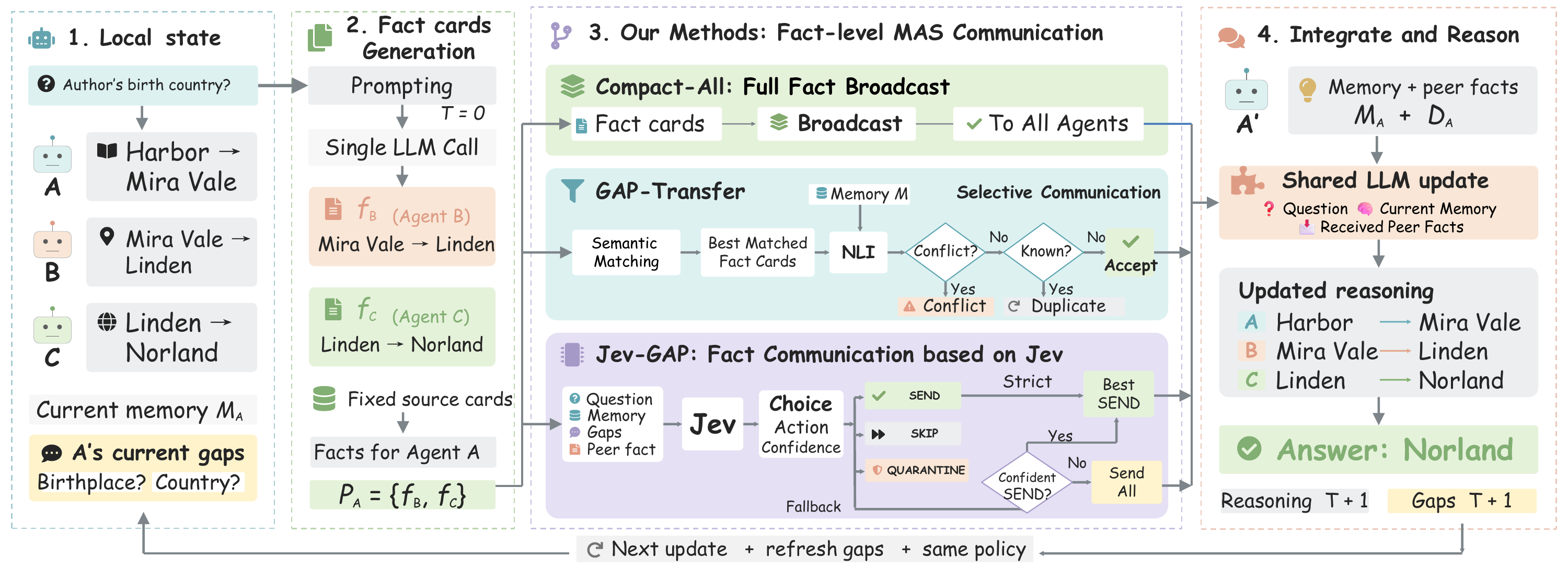}
\caption{\textbf{Overview of fact-level MAS communication.}
Three agents generate reasoning, gaps, and fact cards from complementary evidence. For receiver A, $P_A=\{f_B,f_C\}$ contains B's and C's cards. Compact-All broadcasts all cards, GAP-Transfer uses semantic matching and NLI screening, and Jev-GAP uses action and confidence decisions with optional fallback. A integrates delivered facts $D_A$ with memory $M_A$, then updates its reasoning, answer, and gaps.}
\label{fig:gap}
\Description{Four stages show local reasoning, initial fact-card construction, three fact-level communication policies, and integration. Agents hold facts about an author, a birthplace, and a country. The receiver integrates delivered peer facts with its memory to answer Norland and refresh its reasoning gaps.}
\end{figure*}

\subsection{Interaction-Specific Alignment Gain}
Under self refinement, each agent revises using local evidence and its own previous response. MAS also receives previous-state peer responses. Updates are synchronous, with matched questions, models, evidence allocations, update counts, and answer aggregation. For $G_X=C_T^X-C_0^X$, define
\begin{equation}
E_\Delta = G_{\mathrm{MAS}}-G_{\mathrm{SELF}}.
\label{eq:emergence}
\end{equation}
Positive $E_\Delta$ means that MAS gains more coverage, or loses less, than self refinement. Reporting $G_{\Mas}$ distinguishes absolute improvement from reduced loss. The architecture study shares exact initial responses across conditions. The earlier cross-model study uses separate initial generations. Both conditions use the same initial evidence allocation. Computing each trajectory change before subtraction retains improvements and degradations in either condition, so the contrast follows their relative change during communication.

\subsection{Evidence Ownership}
Let $O_i\subseteq\{1,\ldots,L\}$ index units supported by agent $i$'s initial evidence and $U_i=\{1,\ldots,L\}\setminus O_i$ its non-owned units. Both sets are fixed. Non-owned coverage averages the row maxima indexed by $U_i$ for each agent:
\begin{equation}
\begin{split}
 C_{i,t}^{X,\mathrm{non}}&=\frac1{|U_i|}\sum_{k\in U_i}c_{ik,t}^{X},\quad
 C_t^{X,\mathrm{non}}=\frac1N\sum_i C_{i,t}^{X,\mathrm{non}},\\
 \Enon&=G_{\Mas}^{\mathrm{non}}-G_{\mathrm{SELF}}^{\mathrm{non}}.
\end{split}
\label{eq:nonowned}
\end{equation}
Here $G_X^{\mathrm{non}}=C_T^{X,\mathrm{non}}-C_0^{X,\mathrm{non}}$. The main split-evidence settings have nonempty $U_i$. Substituting $O_i$ gives owned coverage and the local-retention diagnostic $G_X^{\mathrm{own}}$. Overall coverage includes the conclusion row. Ownership scores use the intermediate rows because the conclusion can depend on several sources. Ownership denotes the initial source allocation throughout the trajectory. Full derivations are provided in \appref{formal}.

\noindent\textbf{Reading overall and ownership gains together.}
The final conclusion remains a separate contribution to overall coverage. Writing $o_i=|O_i|$, $u_i=|U_i|$, and $C_i^{\mathrm{out}}$ for its row score gives
\begin{equation}
 C_i=\frac{o_i C_i^{\mathrm{own}}+u_i C_i^{\mathrm{non}}+C_i^{\mathrm{out}}}{L+1}.
 \label{eq:ownership-decomposition-main}
\end{equation}
The fixed allocation preserves this identity for temporal changes and MAS-minus-self-refinement contrasts. Gains on peer-supported content can be offset by losses on local content or the conclusion. Joint reporting keeps acquisition and retention visible. With unequal allocations, apply the identity within each agent before averaging. Full derivations appear in \appref{formal}.
\ifhmode\space\fi

\subsection{Fact-Level Communication}
\label{sec:architecture}
\begin{table*}[t]
\caption{\textbf{Collective emergent alignment across model families.}
Three-seed means on MuSiQue and corrected 2Wiki. Paired bars show $C$/F1 and $E_\Delta$/$\Delta$F1 (top/bottom), and single bars show $\Enon$. Lengths are normalized within each benchmark and metric. Bold/underline mark the largest/second-largest model values. F1 is in \%, and $\Delta$F1 is in percentage points. RQ1 explains the saturated 2Wiki scores. Full results appear in \appref{cross-family}.}
\label{tab:main}

\centering
\datatablefont
\setlength{\tabcolsep}{2.8pt}
\renewcommand{\arraystretch}{1.06}

\renewcommand{\tabularxcolumn}[1]{m{#1}}
\begin{tabularx}{\textwidth}{@{}>{\raggedright\arraybackslash}m{.205\textwidth}|*{3}{>{\centering\arraybackslash}X}|*{3}{>{\centering\arraybackslash}X}@{}}

\toprule

&
\multicolumn{3}{c|}{\textbf{MuSiQue}}
&
\multicolumn{3}{c}{\textbf{2Wiki}}
\\[-1pt]

\cmidrule(lr){2-4}
\cmidrule(l){5-7}

\textbf{Model}
&
\textcolor{covblue}{$C$}\,/\,\textcolor{f1green}{F1}
&
\textcolor{emergeorange}{$E_\Delta$}\,/\,%
\textcolor{deltapurple}{$\Delta$F1}
&
\textcolor{peerred}{$E_\Delta^{\mathrm{non}}$}
&
\textcolor{covblue}{$C$}\,/\,\textcolor{f1green}{F1}
&
\textcolor{emergeorange}{$E_\Delta$}\,/\,%
\textcolor{deltapurple}{$\Delta$F1}
&
\textcolor{peerred}{$E_\Delta^{\mathrm{non}}$}
\\

\midrule


\rowcolor{closedband}
\multicolumn{7}{l}{
\textcolor{closedtext}{
\textbf{\textsf{Closed / API models}}
}
}
\\[-1pt]


GPT-5.6 Terra
&
\finalpair
    {.2758}{.3303}{.2758}
    {23.60}{45.75}{23.60}
&
\effectpair
    {.0422}{.0422}{\vbest{.0422}}
    {-2.52}{8.53}{\textcolor{negred}{\ensuremath{-}2.52}}
&
\peerbar{.0831}{.0944}{.0831}
&
\finalpair
    {.5078}{.6222}{.5078}
    {100.00}{100.00}{\vbest{100.00}}
&
\effectpair
    {.2021}{.3025}{.2021}
    {100.00}{100.00}{\vbest{+100.00}}
&
\peerbar{.4039}{.5938}{.4039}
\\


\rowcolor{closedstripe}
Gemini 3.1 Flash-Lite
&
\finalpair
    {.3303}{.3303}{\vbest{.3303}}
    {45.62}{45.75}{\vsecond{45.62}}
&
\effectpair
    {.0217}{.0422}{.0217}
    {.21}{8.53}{+.21}
&
\peerbar{.0484}{.0944}{.0484}
&
\finalpair
    {.3898}{.6222}{.3898}
    {94.83}{100.00}{94.83}
&
\effectpair
    {-0.0088}{.3025}
    {\textcolor{negred}{\ensuremath{-}.0088}}
    {42.53}{100.00}{+42.53}
&
\peerbar{.1779}{.5938}{.1779}
\\


Claude Haiku 4.5
&
\finalpair
    {.2408}{.3303}{.2408}
    {22.78}{45.75}{22.78}
&
\effectpair
    {.0236}{.0422}{.0236}
    {1.49}{8.53}{+1.49}
&
\peerbar{.0475}{.0944}{.0475}
&
\finalpair
    {.4550}{.6222}{.4550}
    {99.43}{100.00}{\vsecond{99.43}}
&
\effectpair
    {.1710}{.3025}{.1710}
    {97.70}{100.00}{\vsecond{+97.70}}
&
\peerbar{.4592}{.5938}{.4592}
\\

\addlinespace[1.5pt]


\rowcolor{openband}
\multicolumn{7}{l}{
\textcolor{opentext}{
\textbf{\textsf{Open-weight models}}
}
}
\\[-1pt]


Qwen3.5-35B-A3B
&
\finalpair
    {.3023}{.3303}{.3023}
    {45.75}{45.75}{\vbest{45.75}}
&
\effectpair
    {.0421}{.0422}{\vsecond{.0421}}
    {8.53}{8.53}{\vbest{+8.53}}
&
\peerbar{.0836}{.0944}{\vsecond{.0836}}
&
\finalpair
    {.6222}{.6222}{\vbest{.6222}}
    {88.51}{100.00}{88.51}
&
\effectpair
    {.3025}{.3025}{\vbest{.3025}}
    {64.94}{100.00}{+64.94}
&
\peerbar{.5938}{.5938}{\vbest{.5938}}
\\


\rowcolor{openstripe}
Mistral Small 3.2
&
\finalpair
    {.3299}{.3303}{\vsecond{.3299}}
    {17.25}{45.75}{17.25}
&
\effectpair
    {.0371}{.0422}{.0371}
    {7.21}{8.53}{+7.21}
&
\peerbar{.0944}{.0944}{\vbest{.0944}}
&
\finalpair
    {.6201}{.6222}{\vsecond{.6201}}
    {67.38}{100.00}{67.38}
&
\effectpair
    {.2821}{.3025}{.2821}
    {36.99}{100.00}{+36.99}
&
\peerbar{.5499}{.5938}{\vsecond{.5499}}
\\


DeepSeek V3.2
&
\finalpair
    {.2635}{.3303}{.2635}
    {32.55}{45.75}{32.55}
&
\effectpair
    {.0316}{.0422}{.0316}
    {7.34}{8.53}{\vsecond{+7.34}}
&
\peerbar{.0789}{.0944}{.0789}
&
\finalpair
    {.5739}{.6222}{.5739}
    {100.00}{100.00}{\vbest{100.00}}
&
\effectpair
    {.2977}{.3025}{\vsecond{.2977}}
    {100.00}{100.00}{\vbest{+100.00}}
&
\peerbar{.5208}{.5938}{.5208}
\\


\rowcolor{openstripe}
GLM-5
&
\finalpair
    {.2779}{.3303}{.2779}
    {37.20}{45.75}{37.20}
&
\effectpair
    {.0411}{.0422}{.0411}
    {.82}{8.53}{+.82}
&
\peerbar{.0698}{.0944}{.0698}
&
\finalpair
    {.5301}{.6222}{.5301}
    {82.18}{100.00}{82.18}
&
\effectpair
    {.2539}{.3025}{.2539}
    {82.18}{100.00}{+82.18}
&
\peerbar{.4776}{.5938}{.4776}
\\

\bottomrule

\end{tabularx}
\end{table*}
\ifhmode\space\fi

REMAP reveals that peer interaction helps agents acquire missing information but can also weaken their existing reasoning. This motivates fact-level communication to better control information acquisition and local retention. We compare three communication policies using the shared fact interface in Figure~\ref{fig:gap}. In the following, we will introduce the details of the proposed MAS architecture.

\noindent\textbf{Fact cards from the initial generation.}
At state zero, one LLM call produces a local-evidence fact $f_i$, its source identifier, gaps $Q_{i,0}$, three reasoning steps, a conclusion, and an answer. A deterministic parser reads the fact field directly. All architectures reuse these initial records, with source cards fixed across updates. A gap is a self-identified information need.

\noindent\textbf{Complete compact broadcast.}
Receiver $i$ has two initial peer cards, $\mathcal F_{-i}=\{f_j:j\ne i\}$. \textbf{Compact-All} inserts both into its update prompt at every communication round. The receiver then makes its normal reasoning-update call. Delivery uses deterministic field selection. \textbf{Random-K} uses the same cards and integration prompt, but samples a subset matching the number delivered by Jev-GAP-Fallback for the same receiver and update as a control group.

\noindent\textbf{Gap-aware selection.}
\textbf{GAP-Transfer} selects the highest-MiniLM-cosine peer fact for each current gap. An NLI gate compares candidates with receiver memory: contradictions are quarantined, entailments are treated as duplicates, and other selected facts are accepted. The agent then updates its reasoning and gaps. Hard GAP-MAS and GAP-SOFT add local-retention instructions.
\noindent\textbf{Jev-GAP.} Recent studies have demonstrated Jev's potential for lightweight semantic decisions in agent memory management~\cite{deußer2026evaluatingbenchmarkingmodeljev,jiang2026jevmemsystemonecontrolledagenticmemory}. Jev-GAP replaces this selector and gate with Jev model~\cite{typesafe2026jev}. Given the question, gap, receiver facts, candidate, and source, Jev returns relevance, novelty, and conflict probabilities, plus a Choice action $a\in\{\mathrm{SEND},\mathrm{SKIP},\mathrm{QUARANTINE}\}$, option probabilities $\pi$, and confidence $c$. SEND candidates are ranked by $\pi(\mathrm{SEND})$, with at most two distinct cards delivered to each receiver. 
 
\noindent\textbf{Confidence-guided fallback.}
\textbf{Jev-GAP-Fallback} broadens sharing when a gap has no sufficiently confident SEND candidate. Let $S_{i,t}(g)$ contain SEND candidates for gap $g$, and let $H_{i,t}(g)=1$ when at least one candidate has confidence $c\geq\theta$, where $\theta$ is calibrated on a development set. The payload is
\begin{equation}
 P_{i,t}(g)=\begin{cases}
 \operatorname{Top2}_{\pi(\mathrm{SEND})}S_{i,t}(g),&H_{i,t}(g)=1,\\
 \mathcal F_{-i},&H_{i,t}(g)=0.
 \end{cases}
\label{eq:jev-route}
\end{equation}
Fallback supplies the same initial compact cards. Cards are deduplicated across gaps. An empty gap set leaves the selected payload empty. Compact-All broadcasts on every update. The threshold uses official Choice confidence~\cite{typesafe2026confidence}. All policies use fixed rounds, reference-free runtime inputs, and self-refinement controls. Full communication-policy details are provided in \apprefs{gap-protocol}{jev-protocol}.
\ifhmode\space\fi
\setlength{\parskip}{0pt plus 2pt}
\section{Experiments and Results}\label{sec:experiments}
We first introduce the experimental setup, followed by results that address each research question (RQ).

\subsection{Experimental Setup}\label{sec:setup}

\noindent\textbf{Datasets and models.}
We evaluate seven model families on held-out three-hop MuSiQue and corrected 2Wiki questions~\cite{trivedi2022musique,ho2020twowiki}. Communication experiments span diverse model families and capability levels, including 3 closed-source models and 4 open-weight models.

\noindent\textbf{References and protocol.}
References come from annotated subanswers or evidence triples for offline scoring. Three agents receive distributed evidence, generate initial responses, and perform two synchronous updates under self refinement or MAS, using seeds 7/42/123. Architecture conditions share the initial outputs and the fact cards produced alongside them. Execution details and prompt templates are provided in \apprefs{protocol}{prompts}.

\noindent\textbf{Baselines.}
We compare Full-MAS, AgentPrune, AgentDropout, and PEAR with Compact-All, Random-K, GAP-Transfer, and Jev variants. Full adaptations are provided in \appref{comm-protocol}.

\noindent\textbf{Evaluation and reporting.}
Frozen \texttt{roberta-large-mnli}~\cite{liu2019roberta,williams2018mnli} scores reasoning steps and conclusions. Final plurality answers use EM (exact match) and token-level F1. $\Delta$F1 is final MAS F1 minus matched self-refinement F1. Gold recovery $T_{\rm gold}$ is the fraction of annotated subanswers (MuSiQue) or triple values (2Wiki) matched in each agent's emitted steps and conclusion using a separate normalized lexical matcher. $\Delta T_{\rm gold}$ subtracts the self-refinement initial-to-final recovery change from the corresponding MAS change, averaged over agents. We report three-seed means and paired question-bootstrap intervals using 10,000 resamples. \apprefs{protocol}{validity} detail the evaluation protocol and scoring procedures.
\ifhmode\space\fi

\subsection{RQ1: Positive Interaction Gains}\label{sec:models}
\subsubsection{Cross-model alignment gains}\label{sec:cross-model-gains}
Table~\ref{tab:main} shows positive overall gains throughout MuSiQue and for most models on 2Wiki, with positive non-owned gains in every reported setting. The pattern extends to DeepSeek and GLM, and gains are generally larger on corrected 2Wiki.

Gemini gains peer-supported content on 2Wiki despite a slightly negative overall contrast. DeepSeek's MuSiQue contrast reflects less loss than self refinement, while its 2Wiki coverage increases absolutely. Positive interaction gains can therefore reflect acquisition or better retention. Coverage and answer-quality rankings differ across models. Full states and intervals appear in \appref{cross-family}.

\subsubsection{Evidence gaps and answer saturation}\label{sec:saturation}
Complementary splitting makes the 2Wiki comparison questions difficult for isolated agents: each receives one of three evidence facts, while the comparison requires information held by peers. Self refinement adds no evidence and instructs agents to answer incorrectly when their information is insufficient. MAS exposes the missing facts through peer responses, enabling the comparison to be completed.

Terra and DeepSeek reach 100\% MAS F1 and 0\% self-refinement F1 in all 174 question--seed units, yielding $+100$ percentage points in $\Delta$F1. This is \emph{floor--ceiling saturation} under the complementary-evidence protocol. The question-level audit attributes the contrast to self-refinement abstention and correct MAS answers, with no detected scaling or answer-parsing error. For example, agents that each know one director can resolve a comparison of director counts after sharing their facts.

\noindent\textbf{What the saturated results establish.}
These outcomes support successful \emph{controlled evidence integration} and peer-information acquisition. Emergent reasoning alignment is assessed through $E_\Delta$, the ownership decomposition, and independent gold recovery, which track changes within the reasoning process. A correct short answer can coexist with partial Semantic Coverage because $C$ scores intermediate reference content as well as the conclusion. This interpretation concerns the controlled evidence split. Audit and saturation statistics are provided in \appref{cross-family}.

\suppressfloats[t]
\begin{table}[t]
\caption{\textbf{Metric validity.} Within-model Pearson $r$ for Gemini (G) and Qwen (Q). Bold/underline mark the largest/second-largest values per panel and column. \appref{validity} gives full correlations and intervals.}
\label{tab:validity}
\centering\datatablefont\setlength{\tabcolsep}{3pt}
\renewcommand{\arraystretch}{1.28}
\begin{tabular*}{\columnwidth}{@{\extracolsep{\fill}}lrrrr@{}}
\toprule
 & \multicolumn{2}{c}{MuSiQue} & \multicolumn{2}{c}{2Wiki}\\
\cmidrule(lr){2-3}\cmidrule(l){4-5}
Measure / scorer & G & Q & G & Q\\
\midrule
\mainband{5}{topologyband}{A. Final-state alignment $\leftrightarrow$ F1}
Structured $\Arel$ & \vsecond{\ensuremath{-}.006} & \vsecond{.111} & \vbest{\ensuremath{-}.021} & \vsecond{\ensuremath{-}.028}\\
Semantic Coverage $C$ & \vbest{.602} & \vbest{.663} & \vsecond{\ensuremath{-}.106} & \vbest{.127}\\
\midrule
\mainband{5}{factband}{B. Trajectory change $\leftrightarrow$ $\Delta$F1}
$E_{\rm rel,\Delta}$ (structured) & .183 & \vnegative{\ensuremath{-}.034} & \vnegative{\ensuremath{-}.368} & \vnegative{\ensuremath{-}.220}\\
$G_{\Mas}$ & .229 & \vbest{.405} & \vsecond{.210} & \vsecond{.156}\\
$E_\Delta$ & \vsecond{.237} & \vsecond{.352} & .204 & \vbest{.176}\\
$\Enon$ & \vbest{.299} & .294 & \vbest{.214} & .132\\
\midrule
\mainband{5}{topologyband}{C. Trajectory change $\leftrightarrow$ $\Delta T_{\rm gold}$}
$G_{\Mas}$ & .399 & .541 & \vsecond{.592} & \vsecond{.602}\\
$E_\Delta$ & \vsecond{.471} & \vsecond{.562} & .551 & .591\\
$\Enon$ & \vbest{.525} & \vbest{.664} & \vbest{.612} & \vbest{.606}\\
\midrule
\mainband{5}{factband}{D. Non-owned scorer $\leftrightarrow$ $\Delta T_{\rm gold}$}
Token-F1 & .428 & .379 & \vnegative{\ensuremath{-}.162} & .456\\
ROUGE-L & .447 & .383 & \vsecond{\ensuremath{-}.040} & \vsecond{.582}\\
MiniLM cosine & \vsecond{.485} & \vsecond{.562} & \vnegative{\ensuremath{-}.223} & .569\\
NLI coverage (REMAP) & \vbest{.525} & \vbest{.664} & \vbest{.612} & \vbest{.606}\\
\bottomrule\end{tabular*}\end{table}
\ifhmode\space\fi
\begin{table*}[t]
\caption{\textbf{Heterogeneous gains over homogeneous controls.} Entries subtract each control mean from the mixed-team mean. $\Delta$F1 differences are in percentage points. Bold/underline mark the largest/second-largest difference per task and metric. Coloured circles identify capability tiers, and subscripts identify models. Full paired results and source-trace audits appear in \appref{heterogeneity}.}
\label{tab:heterogeneity}
\centering\datatablefont\setlength{\tabcolsep}{3pt}\renewcommand{\arraystretch}{1.08}
\begin{tabular*}{\textwidth}{@{\extracolsep{\fill}}ll|rrr|rrr@{}}
\toprule
 & & \multicolumn{3}{c}{\textbf{MuSiQue}} & \multicolumn{3}{c}{\textbf{2Wiki}}\\
\cmidrule(lr){3-5}\cmidrule(l){6-8}
Mixed team & Homogeneous control & $E_\Delta$ & $\Enon$ & $\Delta$F1 & $E_\Delta$ & $\Enon$ & $\Delta$F1\\
\midrule
\mainband{8}{topologyband}{Weak-tier mixtures}
\tierdisc{W}{1}\,\tierdisc{W}{1}\,\tierdisc{W}{2} & \tierdisc{W}{1}\,\tierdisc{W}{1}\,\tierdisc{W}{1} & +0.0120 & +0.0070 & +5.24 & \ensuremath{-}0.0461 & \ensuremath{-}0.0479 & \vsecond{+29.17}\\
\tierdisc{W}{1}\,\tierdisc{W}{1}\,\tierdisc{W}{2} & \tierdisc{W}{2}\,\tierdisc{W}{2}\,\tierdisc{W}{2} & \vbest{+0.0254} & \vbest{+0.0539} & \vbest{+10.95} & +0.0650 & +0.0428 & \ensuremath{-}31.54\\
\tierdisc{W}{1}\,\tierdisc{W}{2}\,\tierdisc{W}{2} & \tierdisc{W}{1}\,\tierdisc{W}{1}\,\tierdisc{W}{1} & \ensuremath{-}0.0062 & \ensuremath{-}0.0244 & +3.34 & \ensuremath{-}0.0864 & \ensuremath{-}0.0774 & \vbest{+49.21}\\
\tierdisc{W}{1}\,\tierdisc{W}{2}\,\tierdisc{W}{2} & \tierdisc{W}{2}\,\tierdisc{W}{2}\,\tierdisc{W}{2} & +0.0072 & +0.0225 & \vsecond{+9.06} & +0.0246 & +0.0132 & \ensuremath{-}11.49\\
\addlinespace[2pt]
\mainband{8}{factband}{Medium-tier mixtures}
\tierdisc{M}{1}\,\tierdisc{M}{1}\,\tierdisc{M}{2} & \tierdisc{M}{1}\,\tierdisc{M}{1}\,\tierdisc{M}{1} & \ensuremath{-}0.0038 & \ensuremath{-}0.0092 & \ensuremath{-}6.02 & \ensuremath{-}0.0333 & \ensuremath{-}0.0389 & +0.00\\
\tierdisc{M}{1}\,\tierdisc{M}{1}\,\tierdisc{M}{2} & \tierdisc{M}{2}\,\tierdisc{M}{2}\,\tierdisc{M}{2} & \ensuremath{-}0.0143 & \ensuremath{-}0.0134 & +3.83 & +0.0623 & +0.0780 & +0.00\\
\tierdisc{M}{1}\,\tierdisc{M}{2}\,\tierdisc{M}{2} & \tierdisc{M}{1}\,\tierdisc{M}{1}\,\tierdisc{M}{1} & +0.0034 & \ensuremath{-}0.0063 & \ensuremath{-}5.03 & \ensuremath{-}0.0695 & \ensuremath{-}0.0862 & +0.00\\
\tierdisc{M}{1}\,\tierdisc{M}{2}\,\tierdisc{M}{2} & \tierdisc{M}{2}\,\tierdisc{M}{2}\,\tierdisc{M}{2} & \ensuremath{-}0.0072 & \ensuremath{-}0.0105 & +4.82 & +0.0261 & +0.0307 & +0.00\\
\addlinespace[2pt]
\mainband{8}{topologyband}{Strong-tier mixtures}
\tierdisc{S}{1}\,\tierdisc{S}{1}\,\tierdisc{S}{2} & \tierdisc{S}{1}\,\tierdisc{S}{1}\,\tierdisc{S}{1} & \ensuremath{-}0.0031 & \ensuremath{-}0.0063 & \ensuremath{-}0.47 & \ensuremath{-}0.1162 & \ensuremath{-}0.1682 & \ensuremath{-}6.44\\
\tierdisc{S}{1}\,\tierdisc{S}{1}\,\tierdisc{S}{2} & \tierdisc{S}{2}\,\tierdisc{S}{2}\,\tierdisc{S}{2} & \vsecond{+0.0174} & \vsecond{+0.0289} & +7.85 & \vbest{+0.1951} & \vbest{+0.2477} & +15.98\\
\tierdisc{S}{1}\,\tierdisc{S}{2}\,\tierdisc{S}{2} & \tierdisc{S}{1}\,\tierdisc{S}{1}\,\tierdisc{S}{1} & \ensuremath{-}0.0159 & \ensuremath{-}0.0289 & \ensuremath{-}7.66 & \ensuremath{-}0.2188 & \ensuremath{-}0.3011 & \ensuremath{-}16.09\\
\tierdisc{S}{1}\,\tierdisc{S}{2}\,\tierdisc{S}{2} & \tierdisc{S}{2}\,\tierdisc{S}{2}\,\tierdisc{S}{2} & +0.0046 & +0.0063 & +0.65 & \vsecond{+0.0925} & \vsecond{+0.1147} & +6.32\\
\bottomrule\end{tabular*}
\par\smallskip{\small\tierlegend\quad Full-spread teams: \tierdisc{W}{1}\,\tierdisc{M}{1}\,\tierdisc{S}{1} and \tierdisc{W}{2}\,\tierdisc{M}{2}\,\tierdisc{S}{2}.}
\end{table*}
\ifhmode\space\fi
\ifhmode\space\fi
\subsection{RQ2: Peer Recovery and Transfer}\label{sec:validity}
\subsubsection{Validity and peer recovery}\label{sec:validity-results}
Table~\ref{tab:validity} compares reference coverage, answer quality, and peer-information recovery. EM/F1 score the final plurality answer, while independent gold recovery tracks intermediate facts in each agent's emitted reasoning.

\noindent\textbf{Task outcomes.}
Final Semantic Coverage tracks F1 more closely than the Structured Alignment Score on MuSiQue, but this relationship is weak on 2Wiki. Interaction gains have positive mean correlations with matched $\Delta$F1 on both tasks, although the 2Wiki intervals include zero. Reasoning change and answer improvement provide complementary evidence. The constant 2Wiki F1 profiles of Terra and DeepSeek have undefined within-model F1 correlations.

\noindent\textbf{Peer recovery.}
The most consistent association links $\Enon$ with independently measured gold-item recovery. NLI coverage has the largest correlation estimates among the tested lexical and embedding scorers. They share gold targets but use distinct matchers, providing convergent evidence of peer recovery. Full correlations, scorer comparisons, answer transitions, and length/headroom analyses appear in \appref{validity}.

\noindent\textbf{Association strength.}
For Gemini and Qwen, correlations between $\Enon$ and gold recovery are $0.525$ and $0.664$ on MuSiQue, and $0.612$ and $0.606$ on 2Wiki. Each exceeds its corresponding correlation with $\Delta$F1. The ownership decomposition follows intermediate-evidence recovery, while F1 summarizes the final answer. Joint reporting connects information transfer to both reasoning and task outcomes.

\noindent\textbf{Recovery checks.}
The lexical matcher searches MuSiQue gold subanswers and valid aliases or annotated 2Wiki triple values. Since a mention may lack the correct relation, the strict 2Wiki check requires subject and object in the same generated unit. Positive associations persist under this check. Alternative scorers preserve the row-wise maximum and trajectory contrast while changing only the local match function. Full sensitivity results are in \appref{validity}.
\ifhmode\space\fi
\subsubsection{Heterogeneous team gains}\label{sec:heterogeneity}
We define weak (W), medium (M), and strong (S) tiers from the main experiment's frozen MuSiQue self-refinement F1. Among the six selected models, $W_1/W_2$ denote Mistral Small 3.2 24B/Claude Haiku 4.5, $M_1/M_2$ denote DeepSeek V3.2/GPT-5.6 Terra, and $S_1/S_2$ denote Qwen3.5-35B-A3B/Gemini 3.1 Flash-Lite. These assignments are retained on 2Wiki. Table~\ref{tab:heterogeneity} compares each same-tier mixture with both homogeneous endpoints.
\begin{table}[!t]
\caption{\textbf{Owner--receiver directionality.} Mean forward-minus-reverse differences. Positive values favour the displayed direction. Bold/underline mark the largest/second-largest difference per panel and task. Tier colours follow Table~\ref{tab:heterogeneity}. Intervals and audits appear in \appref{heterogeneity}.}
\label{tab:directionality}
\centering\datatablefont\setlength{\tabcolsep}{2.5pt}\renewcommand{\arraystretch}{1.35}
\begin{tabular*}{\columnwidth}{@{\extracolsep{\fill}}lrr@{}}
\toprule
Direction contrast & MuSiQue & 2Wiki\\
\midrule
\mainband{3}{factband}{A. self-refinement-adjusted gold-recovery difference}
\tierdisc{M}{}\,$\to$\,\tierdisc{W}{}\,$-$\,\tierdisc{W}{}\,$\to$\,\tierdisc{M}{} & +0.0817 & \ensuremath{-}0.0259\\
\tierdisc{S}{}\,$\to$\,\tierdisc{W}{}\,$-$\,\tierdisc{W}{}\,$\to$\,\tierdisc{S}{} & \vbest{+0.1383} & \vsecond{+0.1408}\\
\tierdisc{S}{}\,$\to$\,\tierdisc{M}{}\,$-$\,\tierdisc{M}{}\,$\to$\,\tierdisc{S}{} & \vsecond{+0.1283} & \vbest{+0.2615}\\
\midrule
\mainband{3}{topologyband}{B. Non-owned coverage-gain difference}
\tierdisc{M}{}\,$\to$\,\tierdisc{W}{}\,$-$\,\tierdisc{W}{}\,$\to$\,\tierdisc{M}{} & \vsecond{+0.0329} & +0.0156\\
\tierdisc{S}{}\,$\to$\,\tierdisc{W}{}\,$-$\,\tierdisc{W}{}\,$\to$\,\tierdisc{S}{} & \vbest{+0.0707} & \vbest{+0.0847}\\
\tierdisc{S}{}\,$\to$\,\tierdisc{M}{}\,$-$\,\tierdisc{M}{}\,$\to$\,\tierdisc{S}{} & +0.0183 & \vsecond{+0.0476}\\
\bottomrule\end{tabular*}
\end{table}
\ifhmode\space\fi
\begin{table*}[t]
\caption{\textbf{MAS architecture comparison.} Matched Qwen/Gemini means over three seeds and two communication rounds. Bold/underline mark the best/second-best value per task and column. F1 is in \%, and $\Delta$F1 is the gain over self refinement in percentage points. Peer counts measure delivered tokens, cost is USD$\times10^{-3}$, and time is in seconds. $\Delta$ rows subtract Full-MAS from the named method. Paired results and budgets appear in \apprefs{comm-protocol}{comm-results}.}
\label{tab:gapmain}
\centering
\datatablefont
\setlength{\tabcolsep}{2.35pt}
\renewcommand{\arraystretch}{1.07}
\setlength{\arrayrulewidth}{0.35pt}

\begin{tabular*}{\textwidth}{@{\extracolsep{\fill}}cl|rrrrrr|rrr@{}}
\toprule
 & &
 \multicolumn{3}{c}{\textbf{Alignment change}} &
 \multicolumn{3}{c}{\textbf{Recovery and answer}} &
 \multicolumn{3}{c}{\textbf{Resource use}}\\

\cmidrule(lr){3-5}
\cmidrule(lr){6-8}
\cmidrule(l){9-11}

\textbf{Task}
& \textbf{Method}
& $E_\Delta\uparrow$
& $\Enon\uparrow$
& $G^{\rm own}\uparrow$
& $\Delta T_{\rm gold}\uparrow$
& F1$\uparrow$
& $\Delta$F1$\uparrow$
& \vhead{Peer tok.$\downarrow$}
& \vhead{Cost$\downarrow$\\(m\$)}
& \vhead{Time$\downarrow$\\(s)}\\

\midrule

\multirow{14}{*}{\rotatebox[origin=c]{90}{\textbf{MuSiQue}}}
& \multicolumn{10}{c}{\cellcolor{referenceband}\strut\textit{Original full-MAS baseline}}\\

& Full-MAS~\cite{du2024debate}
& .0530
& .10907
& \ensuremath{-}.0709
& .2017
& 37.08
& +2.64
& 1,725.7
& 3.92
& 12.93\\

& \multicolumn{10}{c}{\cellcolor{topologyband}\strut\textit{Topology-level routing baselines}}\\

& AgentPrune~\cite{zhang2025agentprune}
& .0213
& .06009
& \ensuremath{-}.0808
& .0953
& 31.00
& \ensuremath{-}3.44
& 289.5
& 3.71
& 13.14\\

& AgentDropout~\cite{wang2025agentdropout}
& .0362
& .08286
& \ensuremath{-}.0680
& .1103
& 32.09
& \ensuremath{-}2.36
& \vsecond{273.0}
& \vbest{2.90}
& \vbest{9.38}\\

& PEAR~\cite{feng2026pear}
& .0073
& .05131
& \ensuremath{-}.1186
& .0844
& 32.02
& \ensuremath{-}2.42
& 428.0
& 5.70
& 24.88\\

& \multicolumn{10}{c}{\cellcolor{factband}\strut\textit{Fact-level communication (ours)}}\\

& Compact-All
& \vbest{.1039}
& \vbest{.17283}
& \vsecond{\ensuremath{-}.0075}
& \vbest{.3778}
& \vbest{54.49}
& \vbest{+20.04}
& 407.4
& 3.81
& \vsecond{12.35}\\

& Random-K
& \vsecond{.0874}
& \vsecond{.14547}
& \ensuremath{-}.0097
& .2989
& \vsecond{50.04}
& \vsecond{+15.60}
& 390.6
& 3.79
& 12.54\\

& GAP-Transfer
& .0517
& .10046
& \ensuremath{-}.0490
& .1781
& 36.00
& +1.55
& 435.4
& \vsecond{3.63}
& 12.50\\

& Jev-GAP
& .0601
& .09717
& \ensuremath{-}.0127
& .1769
& 45.29
& +10.85
& \vbest{133.8}
& 4.04
& 17.55\\

& Jev-GAP-FB
& .0855
& .14243
& \vbest{\ensuremath{-}.0063}
& \vsecond{.3014}
& 49.20
& +14.76
& 399.8
& 4.11
& 17.30\\

\cmidrule(l){2-11}

& $\Delta$ Compact-All
& +.0509
& +.06376
& +.0635
& +.1761
& +17.41
& +17.41
& \ensuremath{-}1,318.3
& \ensuremath{-}0.11
& \ensuremath{-}0.57\\

& $\Delta$ Jev-GAP-FB
& +.0325
& +.03336
& +.0647
& +.0997
& +12.12
& +12.12
& \ensuremath{-}1,325.9
& +0.19
& +4.37\\

\midrule

\multirow{14}{*}{\rotatebox[origin=c]{90}{\textbf{2Wiki}}}
& \multicolumn{10}{c}{\cellcolor{referenceband}\strut\textit{Original full-MAS baseline}}\\

& Full-MAS~\cite{du2024debate}
& .0432
& .19360
& \ensuremath{-}.0701
& .3309
& 27.51
& +12.73
& 1,421.9
& 3.39
& 11.24\\

& \multicolumn{10}{c}{\cellcolor{topologyband}\strut\textit{Topology-level routing baselines}}\\

& AgentPrune~\cite{zhang2025agentprune}
& .0238
& .14263
& \ensuremath{-}.0464
& .2227
& 31.33
& +16.55
& 236.5
& 3.09
& 9.95\\

& AgentDropout~\cite{wang2025agentdropout}
& .0297
& .14866
& \ensuremath{-}.0319
& .2131
& 27.53
& +12.75
& 227.0
& \vbest{2.41}
& \vbest{7.10}\\

& PEAR~\cite{feng2026pear}
& .0081
& .13173
& \ensuremath{-}.0955
& .2337
& \vbest{45.66}
& \vbest{+30.87}
& 421.7
& 4.96
& 20.26\\

& \multicolumn{10}{c}{\cellcolor{factband}\strut\textit{Fact-level communication (ours)}}\\

& Compact-All
& \vbest{.1132}
& \vbest{.34778}
& \ensuremath{-}.0845
& \vbest{.6260}
& \vsecond{32.74}
& \vsecond{+17.95}
& 279.6
& 3.14
& 10.11\\

& Random-K
& .0923
& .20165
& .1232
& \vsecond{.3851}
& 27.21
& +12.43
& \vsecond{202.7}
& 3.13
& 10.19\\

& GAP-Transfer
& .0350
& .12829
& .0290
& .1638
& 31.12
& +16.34
& 275.7
& \vsecond{3.02}
& \vsecond{9.87}\\

& Jev-GAP
& .1011
& .17179
& \vbest{.2138}
& .2854
& 30.11
& +15.33
& \vbest{144.5}
& 3.25
& 13.71\\

& Jev-GAP-FB
& \vsecond{.1065}
& \vsecond{.20173}
& \vsecond{.1790}
& .3549
& 28.65
& +13.86
& 207.3
& 3.29
& 13.97\\

\cmidrule(l){2-11}

& $\Delta$ Compact-All
& +.0699
& +.15418
& \ensuremath{-}.0144
& +.2950
& +5.23
& +5.22
& \ensuremath{-}1,142.3
& \ensuremath{-}0.25
& \ensuremath{-}1.12\\

& $\Delta$ Jev-GAP-FB
& +.0632
& +.00813
& +.2491
& +.0240
& +1.13
& +1.13
& \ensuremath{-}1,214.5
& \ensuremath{-}0.10
& +2.73\\

\bottomrule
\end{tabular*}
\end{table*}
\ifhmode\space\fi

\noindent\textbf{Mixed-team gains.}
On MuSiQue, the weak-tier $W_1W_1W_2$ mixture improves overall alignment gain and matched task gain over both homogeneous endpoints. Strong-tier mixing also improves non-owned gain over one endpoint on both tasks, but remains below the other. The full-spread teams likewise behave differently from one another. The gains depend on the particular model combination and the homogeneous endpoint used for comparison.

\subsubsection{Directional information recovery}\label{sec:directional-results}
Table~\ref{tab:directionality} compares recovery in each direction with its reverse. After the self refinement adjustment, recovery of strong-owned facts by weak receivers exceeds the reverse on both tasks. The strong-to-medium gold-recovery contrast follows the same pattern, while the medium-to-weak contrast varies across tasks. Direction identifies the initial evidence owner and the receiving agent, providing a source-to-receiver view of content recovery across the full interaction.

\noindent\textbf{Reading reciprocal contrasts.}
Positive adjusted contrasts indicate greater interaction-associated recovery in the named direction. Coverage and lexical recovery provide complementary evidence.

\noindent\textbf{Source-trace audit.}
The reused traces contain context mismatches, and Mistral mixtures also change serving backend. We report these composition and directional results as exploratory. Sensitivity testing on a shared subset of context-consistent questions is pending. Complete contrasts and audit details appear in \appref{heterogeneity}.
\ifhmode\space\fi
\subsection{RQ3: Efficient Fact Exchange}\label{sec:gapresults}
Table~\ref{tab:gapmain} compares full reasoning exchange, topology-level routing, and fact-level communication. The following sections report information recovery, answer quality, and resource use together.

\subsubsection{Compact-exchange gains}\label{sec:compact-gains}
Compact-All and Jev-GAP-Fallback yield higher non-owned gains than the evaluated topology baselines on both tasks. Compact-All also improves overall alignment, gold recovery, and pooled F1 over Full-MAS while delivering substantially fewer peer tokens. These results favour sharing compact source facts in the evaluated distributed-evidence setting. The policies reuse fact cards from the shared initial outputs. Full paired comparisons are provided in \appref{comm-results}.

\noindent\textbf{Magnitude of compact-exchange gains.}
Relative to Full-MAS, Compact-All raises MuSiQue F1 from $37.08$ to $54.49$ and non-owned gain from $0.10907$ to $0.17283$. On 2Wiki, F1 rises from $27.51$ to $32.74$ and non-owned gain from $0.19360$ to $0.34778$. Delivered peer tokens fall from $1{,}725.7$ to $407.4$ and from $1{,}421.9$ to $279.6$, respectively. The compact interface improves source-information recovery while reducing the delivered payload on both datasets.

\subsubsection{Selection and local retention}\label{sec:selection-retention}
Jev-GAP delivers the fewest peer tokens. On 2Wiki it retains more local content than Compact-All, while acquiring less peer-supported content. Fallback recovers some acquisition by broadening compact sharing when no confident SEND is available. Random-K also performs strongly, with pooled F1 close to fallback and paired non-owned-gain intervals spanning zero for their contrast. Complete compact broadcast therefore remains an important control for judging the value of selection. Full controller diagnostics are provided in \appref{jev-results}.

\subsubsection{Cost and communication controls}\label{sec:resource-controls}
Compact-All has the highest F1 on MuSiQue and in the pooled comparison, while PEAR leads on 2Wiki. AgentDropout has the lowest observed fees and latency. Jev's total resource use includes the controller overhead as well as the delivered payload. The Full-MAS comparison also changes the message source from evolving peer traces to fixed initial cards and uses a factual integration interface. This comparison evaluates the complete communication design. Hard and soft retention ablations further show that protecting local content can restrict peer acquisition. Full retention and resource results are provided in \apprefs{gap-results}{comm-results}.

\makeatletter
\let\cea@originalbalancecol\@BAlancecol
\patchcmd{\@BAlancecol}{\unvbox1\vfil}{\dimen2=\dp1\unvbox1\kern-\dimen2\vfil}{}{}
\makeatother
\balance
\noindent\textbf{What the communication controls isolate.}
Compact-All and Random-K share the fact pool and integration prompt. Random-K matches fallback's card count for each receiver and update, controlling content choice at a fixed request schedule. Two cards reproduce compact broadcast, while fewer cards test selection. Topology baselines use their reported realized budgets. Message-count and budget details appear in \appref{comm-protocol}. The shared-start design fixes initial evidence and reasoning, linking source-fact selection to peer recovery and local retention under a common integration interface.
\looseness=2\par
\ifhmode\space\fi
\ifhmode\space\fi

\section{Conclusion}\label{sec:conclusion}
We study whether interaction induces reasoning alignment beyond matched self-refinement. \textbf{REMAP} measures alignment changes across agent trajectories and separates local retention from peer-information gains. Across model families, interaction yields broadly positive gains on non-owned content, supported by independent gold recovery. The 2Wiki saturation highlights controlled evidence integration, while alignment gains describe changes within reasoning. We then introduce fact-level communication with compact broadcast and selective routing. Compact exchange improves alignment and pooled answer quality with fewer peer tokens, while selection reveals acquisition, retention, and cost trade-offs. Together, these findings show that collective emergent alignment can be quantified and used to guide more effective multi-agent communication.
\ifhmode\space\fi

\clearpage
\nobalance
\setlength{\parskip}{0pt}
\section*{Acknowledgements}
This work is partially funded by the European Union (under grant agreement ID 101212818). Views and opinions expressed are however those of the author(s) only and do not necessarily reflect those of the European Union or European Health and Digital Executive Agency (HADEA). Neither the European Union nor the granting authority can be held responsible for them.

This work is partially supported by Innovate UK through AI-PASSPORT under Grant 10126404. This work was awarded a grant by the AI Security Institute (AISI) via the Alignment Project (Rare-Event Estimation in Large Language Models via Subset Simulation) and funded by EPSRC.

Yi's contribution is partially supported through the Royal Society international exchanges programme and in part by the Engineering and Physical Sciences Research Council, through funding from RAi UK [EP/Y009800/1].

\nocite{li2026fragileflow,wang2025smallworld,hu2025responsiblemas,li2026scope,wang2026ambiguity,li2026evidenceorder,li2026segmentrobustness,azulay2026jailbreaking,wang2026topologycontrol}
\bibliographystyle{ACM-Reference-Format}
\bibliography{references}

\raggedbottom
\balance
\clearpage
\nobalance
\makeatletter
\let\@BAlancecol\cea@originalbalancecol
\makeatother
\raggedbottom
\appendix
\setcounter{dbltopnumber}{2}
\renewcommand{\dbltopfraction}{0.9}
\renewcommand{\textfraction}{0.05}
\section{Notation and Measurement Properties}
\label{app:formal}
\subsection{Notation Reference}
Table~\ref{tab:notation} collects the quantities used in the main paper. Indices are suppressed where a single question, state, or agent is fixed. The main score averages agents equally. Each agent's non-owned score averages that agent's non-owned reference units equally.
\begin{table}[h]
\caption{Notation. All averages are computed at question level before dataset aggregation.}
\label{tab:notation}\centering\small
\begin{tabularx}{\columnwidth}{@{}lX@{}}\toprule
Symbol & Meaning\\\midrule
$q$ & A question instance.\\
$N,i$ & Number of agents and agent index.\\
$T,t$ & Final state index and current state. Main runs have $T=2$.\\
$X$ & Experimental condition, self refinement or MAS.\\
$\R,r_k$ & One fixed question reference shared across agents and states. Its $k$th unit.\\
$L,K$ & Evidence-linked units. Total units including conclusion, $K=L+1$.\\
$\Sset_{i,t}^X$ & Scored steps and conclusion of agent $i$ at state $t$ in condition $X$.\\
$s_{i,t,j}^X,J$ & A single generated unit and the unit count in that response.\\
$M_{i,t}^X[k,j]$ & Match of reference $r_k$ to generated unit $s_{i,t,j}^X$.\\
$m(r,s)$ & Mean of the two directional NLI entailment probabilities.\\
$c_{ik,t}^X$ & Maximum match to reference unit $r_k$.\\
$C_{i,t}^X,C_t^X$ & Mean of rows for one agent. Subsequent mean across agents.\\
$O_i,U_i$ & Initially owned and non-owned evidence-linked reference indices.\\
$G_X$ & Initial-to-final coverage change.\\
$E_\Delta,\Enon$ & Overall and non-owned MAS-minus-self-refinement change contrasts.\\
$\Delta T_{\rm gold}$ & Analogous contrast in deterministic gold-item recovery.\\
\bottomrule\end{tabularx}
\end{table}

\subsection{Reference--Output Matching: A Worked Example}
\label{app:matching-example}
The indices represent different objects: $k$ selects a reference unit, $i$ selects an agent, $t$ selects a state, and $j$ selects a generated unit \emph{inside} that agent's current response. The condition $X$ selects self refinement or MAS. A collection $\Sset_{i,t}^X$ is therefore not a single sentence and is not a concatenation of the team. The reference $\R$ is the same at every comparison for that question.

\noindent\textbf{Construct the reference.}
Consider the fictional question, ``In which country was the director of Silent Harbor born?'' A MuSiQue-style decomposition resolves three subquestions using gold subanswers. The deterministic reference strings are:
\begin{quote}\small
$r_1$: The answer to `Who directed Silent Harbor?' is `Nora'.\\
$r_2$: The answer to `Where was Nora born?' is `Riverton'.\\
$r_3$: The answer to `Which country is Riverton in?' is `Luma'.\\
$r_4$: The final answer is `Luma'.
\end{quote}
These are an illustrative instantiation of the template, not dataset records. Corrected 2Wiki instead renders its annotated evidence triples into reference units and appends the final conclusion. Three evidence units may contain parallel facts, so unit count alone does not establish sequential reasoning depth. Full benchmark-specific construction details are provided in \appref{protocol}.

\noindent\textbf{Extract each agent's current units.}
Agent 1 might emit (1) ``Nora was born in Riverton'', (2) ``Nora directed Silent Harbor'', (3) ``Riverton is in Luma'', and the conclusion (4) ``The director was born in Luma''. These four texts form $\Sset_{1,t}^X$. Agents 2 and 3 provide their own collections $\Sset_{2,t}^X$ and $\Sset_{3,t}^X$. Peers may have informed the response, but the scorer reads each agent's emitted steps and conclusion separately. Raw source paragraphs, message metadata, gap fields, memory, and the separate short-answer field are not added to these collections.

\noindent\textbf{Compute all pairs within one agent.}
At a fixed state and condition, abbreviate $s_{i,t,j}^X$ to $s_j$ only inside that agent's matrix. In the following synthetic example, agent 1's row maxima are $\mathbf c_1=(.90,.80,.70,.60)$:
\[
\begin{array}{c|rrrr|r}
 &s_1&s_2&s_3&s_4&\max_j\\\hline
 r_1&.12&\mathbf{.90}&.04&.06&.90\\
 r_2&\mathbf{.80}&.10&.13&.15&.80\\
 r_3&.20&.04&\mathbf{.70}&.08&.70\\
 r_4&.10&.02&.15&\mathbf{.60}&.60
\end{array}
\]
Each cell is the arithmetic mean of the two NLI entailment probabilities. For example, directional probabilities .92 and .88 yield the cell value .90. The numbers are constructed to explain the arithmetic, not measured NLI outputs for the example text. The best matches are $r_1\mapsto s_2$, $r_2\mapsto s_1$, $r_3\mapsto s_3$, and $r_4\mapsto s_4$. This is row-wise maximum matching, without positional alignment, step-order constraints, or a global one-to-one assignment. The same generated unit may be the maximum in multiple reference rows. Ties have the same coverage value regardless of which tied unit is named.

\noindent\textbf{Average agents after scoring.}
Suppose the other agents have row-max vectors $\mathbf c_2=(.50,.90,.40,.60)$ and $\mathbf c_3=(.20,.30,.90,.40)$. The agent means are .75, .60, and .45, and the system mean is
\[
 C_t^X=\frac{.75+.60+.45}{3}=.60.
\]
If agents 1, 2, and 3 own $r_1$, $r_2$, and $r_3$, respectively, then $U_1=\{2,3\}$, $U_2=\{1,3\}$, and $U_3=\{1,2\}$. Reusing the same matrices gives agent non-owned means .75, .45, and .25, so $C_t^{X,\mathrm{non}}=1.45/3\approx.4833$. The conclusion $r_4$ contributes to overall $C$ but not to any ownership score.

\noindent\textbf{Why the aggregation order matters.}
Pooling all agents' statements before taking row maxima would answer whether the \emph{union} of team outputs covers the reference. REMAP instead measures average coverage expressed by each agent. With idealized row-max vectors $(1,0,0,0)$, $(0,1,0,0)$, and $(0,0,1,0)$, REMAP gives $C=.25$. A pooled maximum would give .75 even though no agent expresses a peer's fact. The frozen score is
\[
 \frac1N\sum_i\frac1K\sum_k\max_j M_{i,t}^X[k,j],
\]
not a maximum over $i$ and $j$. No agent-level answer voting or choice of a final spokesperson enters this score. Answer aggregation is evaluated separately.

\noindent\textbf{Repeat across trajectories.}
Compute the same agent-specific matrices at initial and final states of self refinement and MAS. In main-paper Figure 2, self refinement moves from .40 to .45 and MAS from .40 to .60. The changes are .05 and .20, respectively, giving $E_\Delta=.15$. The non-owned contrast applies the same subtraction to the non-owned state means. This is a clarification of the existing computation, with no change to stored experimental scores.

\subsection{Ranges and Invariances}
\noindent\textbf{Boundedness.}
Each local NLI match lies in $[0,1]$. A row maximum and any arithmetic average of row maxima therefore lie in $[0,1]$. It follows that $G_X\in[-1,1]$ and $E_\Delta\in[-2,2]$, with the same ranges for the non-owned versions whenever their reference sets are nonempty. No normalization of the change contrast is applied in the primary results.

\noindent\textbf{Permutation and exact-duplicate invariance.}
Let a generated state contain a finite multiset $S$ of reasoning units. Permuting its elements preserves $\max_{s\in S}m(r_k,s)$ for every $k$. Adding an exact duplicate of an existing unit preserves the same maximum. Both operations therefore preserve state coverage and all contrasts derived from it. This property concerns exact duplication. Adding a semantically different unit can change a row maximum.

\noindent\textbf{Local-score sensitivity.}
Suppose two local scoring functions $m$ and $\widetilde m$ satisfy $|m(r,s)-\widetilde m(r,s)|\leq\varepsilon$ on all scored pairs. For any fixed reference row,
\begin{equation}
 \left|\max_s m(r,s)-\max_s\widetilde m(r,s)\right|\leq\varepsilon.
\end{equation}
Averaging preserves this bound, giving $|C-\widetilde C|\leq\varepsilon$. The triangle inequality gives $|G-\widetilde G|\leq2\varepsilon$ and $|E_\Delta-\widetilde E_\Delta|\leq4\varepsilon$. This is a deterministic sensitivity result under a specified local-score perturbation. The experiments do not assume a known bound on NLI model error.

\noindent\textbf{Shared offsets.}
Adding a fixed question-specific offset to both initial and final scores in a condition cancels from $G_X$. If that operation is applied in both conditions, it also cancels from $E_\Delta$. Sampling variation and changes in reference representation need separate empirical controls, because they need not behave as fixed offsets.

\subsection{Ownership Decomposition}
For agent $i$, write $o_i=|O_i|$ and $u_i=|U_i|$, where $o_i+u_i=L$. Let $C_i^{\rm out}$ denote coverage of the conclusion row. The all-unit state score decomposes as
\begin{equation}
 C_i=\frac{o_i C_i^{\rm own}+u_i C_i^{\rm non}+C_i^{\rm out}}{L+1}.
 \label{eq:s-decomp}
\end{equation}
For an empty owned set, the weighted owned contribution is defined as zero and the standalone owned mean is undefined. The analogous convention applies to an empty non-owned set. This identity also holds after each initial-to-final and MAS-minus-self-refinement difference, because the ownership sets and weights are frozen. With three evidence units and one unit owned by each agent, the system-level identity is
\begin{equation}
 E_\Delta=\tfrac14 E_\Delta^{\rm own}+\tfrac12 E_\Delta^{\rm non}+\tfrac14 E_\Delta^{\rm out}.
\end{equation}
Thus a positive non-owned contrast can be offset by the combined owned and conclusion components. Their separate contributions require their own row-level measurements. This is important when interpreting Gemini's near-zero overall 2Wiki result. For unequal ownership, as in the four-hop allocation experiments, Equation~\eqref{eq:s-decomp} is applied per agent before the equal-agent average.

\subsection{Computation and Caching}
For a trace with $J$ generated units and $K$ reference units, the local matrix needs at most $2KJ$ directional NLI evaluations before caching. With two conditions, $N$ agents, and $T+1$ states, the uncached upper count is $4N(T+1)KJ$ per question. The three-agent, three-state, four-unit setting therefore needs at most 576 directional pair evaluations per question. Repeated identical pairs are retrieved from the local cache. Matrix construction dominates scoring. Row maxima and averaging are linear in the matrix size. Evidence ownership changes only the row aggregation and reuses the same scores.

For the initial three-hop semantic analysis, the local cache contained 27,554 directional pairs across the frozen traces. This reported count reflects deduplication and the specific run. It is not an asymptotic complexity claim. The scoring stage performs no language-model generation requests.

\section{Data, Protocol, and Execution Details}
\label{app:protocol}
\subsection{MuSiQue Reference Construction}
MuSiQue-Ans development examples provide the question, answer, aliases, paragraph context, and question decomposition~\cite{trivedi2022musique}. The primary filter keeps exactly three decomposition steps with valid distinct supporting paragraphs, resolvable references, and nonempty answers. The 100-question held-out set excludes the 30 pilot IDs. Dataset selection uses seed 42 and a frozen ID manifest.

For each decomposition step, a placeholder such as \texttt{\#1} is replaced with the corresponding earlier gold subanswer. The deterministic reference unit is:
\begin{quote}\small
The answer to `\emph{resolved subquestion}' is `\emph{gold subanswer}'.
\end{quote}
A final conclusion states the question's gold answer. These reference strings are constructed deterministically. Each agent receives the original question and its assigned source paragraph, with decomposition metadata and reference text excluded from the generation prompt.

The paragraph-to-agent mapping is deterministically shuffled per question, stored, and reused across self refinement and MAS. Later model-family experiments reuse the original 100 question IDs and evidence allocations.

\subsection{Corrected 2Wiki Construction and Audit}
The source 2Wiki development data supplies contextual support and explicit evidence triples~\cite{ho2020twowiki}. Filtering for exactly three distinct gold triples yields 88 valid examples. The original implementation mixed supporting sentences and rendered triples. An audit found seven incorrect or insufficient mappings among 60 inspected evidence assignments. All 88 exact-gold answer-evaluation invariants passed. 20 inspected reference chains were internally consistent. The initial zero-answer result was therefore retained as a pipeline diagnostic, outside the scientific generalization results.

The corrected protocol uniformly renders the official triples as simple factual text. Each agent owns one rendered fact, explicitly reports its local fact in its response, and may use peer-reported local facts during integration. Multiple legitimate values for a relation are retained. No gold final answer, reference-chain ordering, or ownership labels are shown to the agents. An initial corrected smoke attempt and a subsequent scoped prompt clarification are preserved in the audit provenance. The successful ten-question smoke is included in the corrected 30-question development/validation set.

The subsequent model-family experiment generates all 88 corrected questions and separately reports the frozen 58-question subset excluded from that 30-question set. The reported primary results use the frozen held-out subset. These three-evidence comparison questions can contain parallel facts. Their evidence count should not be described as a three-hop sequential dependency chain. Individual agents have only their assigned fact and are instructed to answer \texttt{unknown} when the available evidence is insufficient. MAS can recover the missing comparison facts from peer responses. The corresponding answer saturation and audit are described in \appref{cross-family}.

\subsection{Condition Semantics}
Self refinement retains the original question, the agent's local source, and its own last response at each update. MAS adds the other agents' last responses. The synchronous barrier is placed after every generated state. An update consumes only messages from the immediately preceding state. The standard output contains \texttt{steps}, \texttt{conclusion}, and \texttt{answer}. Exactly three steps are required in the main three-unit runs.

The prompt descriptions here specify the protocol. The full verbatim prompts and parameter manifests remain in the experimental archive. They should be attached with the final code release. Independent refinement can change answer content and semantic coverage in either direction, which is why both conditions are measured at their own initial and final states.

\subsection{Provider and Repetition Registry}
Table~\ref{tab:providers} records the reported inference profiles. Provider settings are recorded explicitly. Prompts and evidence semantics remain fixed within each benchmark. Provider fallbacks are disabled. Supported formatting parameters are adapted at the API layer, and format-only retries do not receive new task evidence or gold feedback. The same designated seeds are used for local sampling and bookkeeping. Provider support determines whether a request-level seed is effective.

\begin{table*}[t]
\caption{Reported provider and reasoning profiles for the seven-model, three-seed cross-family study. IDs are the requested run identifiers. Provider fallback is disabled.}
\label{tab:providers}\centering\small\setlength{\tabcolsep}{4pt}
\begin{tabular}{@{}lll l@{}}\toprule
Model & Requested identifier & Provider & Reasoning setting\\\midrule
Terra & \texttt{openai/gpt-5.6-terra} & OpenAI & disabled\\
Gemini & \texttt{google/gemini-3.1-flash-lite} & Google & low\\
Haiku & \texttt{anthropic/claude-haiku-4.5} & Amazon Bedrock & disabled\\
Qwen & \texttt{qwen/qwen3.5-35b-a3b} & Darkbloom & disabled\\
Mistral & \texttt{mistralai/mistral-small-3.2-24b-instruct} & DeepInfra & not requested\\
DeepSeek V3.2 & \texttt{deepseek/deepseek-v3.2} & StreamLake & disabled\\
GLM-5 & \texttt{z-ai/glm-5} & StreamLake & disabled\\
\bottomrule\end{tabular}\end{table*}

Qwen's selected provider exhausted the output budget with low/bounded reasoning, so its structured-output runs use disabled reasoning. Gemini uses the reported low setting. DeepSeek V3.2 and GLM-5 use StreamLake with reasoning disabled. Temperature is provider default. Request seeds are sent. Output is capped at 1200 tokens. Fallback is disabled. Their combined scientific-run OpenRouter cost is \$5.797287, excluding health checks and technical dry runs. Mistral's seed-7 logs include transport failures and retries. Haiku includes format retries. The analysis retains the latest successful logical record per slot.

\subsection{Answer Aggregation and Uncertainty}
The final agent answers are normalized with the frozen evaluation code and combined by the same deterministic plurality rule in self refinement and MAS. EM checks normalized exact agreement, and token F1 provides partial-overlap credit. MuSiQue evaluation uses valid gold aliases. The difference $\Delta$F1 compares the two final system outcomes for a matched question and seed.

For repeated runs, the primary estimand averages each question's contrast over the three seeds. Bootstrap samples draw question IDs and preserve the associated matched conditions and seed-average values. This describes uncertainty across sampled questions conditional on the realized runs. Seed means and their standard deviation describe run variation separately. Matched-model comparisons resample shared question IDs together. Independent depth sets use independent bootstrap differences in means. Intervals are pointwise and describe their specified comparison. The primary model and architecture results use the held-out question sets defined above.

\section{State, Interval, and Seed Results}
\label{app:cross-family}\label{sec:s-cross-family}
\begin{table*}[t]
\caption{\textbf{Complete cross-family interaction-gain intervals.} All seven models use three seeds on the same MuSiQue and corrected 2Wiki sets. Cells report mean interaction-specific gain with 95\% question-bootstrap intervals. \textbf{Bold}/\underline{underline} mark the largest/second-largest mean in each benchmark and metric.}
\label{tab:s-model-intervals}
\centering\datatablefont
\setlength{\tabcolsep}{4.0pt}
\renewcommand{\arraystretch}{1.08}
\begin{tabular*}{\textwidth}{@{\extracolsep{\fill}}llcc|cc@{}}
\toprule
&& \multicolumn{2}{c}{\cellcolor{topologyband}\textbf{MuSiQue}} & \multicolumn{2}{c}{\cellcolor{factband}\textbf{2Wiki}}\\
\cmidrule(lr){3-4}\cmidrule(l){5-6}
Model & Family & $E_\Delta\uparrow$ & $E_\Delta^{\rm non}\uparrow$ & $E_\Delta\uparrow$ & $E_\Delta^{\rm non}\uparrow$\\
\midrule
GPT-5.6 Terra & OpenAI & \vbest{.0422} [.0286,.0559] & .0831 [.0606,.1070] & .2021 [.1885,.2154] & .4039 [.3794,.4284]\\
Gemini 3.1 Flash-Lite & Google & .0217 [.0087,.0349] & .0484 [.0293,.0684] & $-.0088$ [$-.0231$,.0058] & .1779 [.1495,.2064]\\
Claude Haiku 4.5 & Anthropic & .0236 [.0145,.0332] & .0475 [.0330,.0622] & .1710 [.1590,.1834] & .4592 [.4395,.4788]\\
Qwen3.5-35B-A3B & Qwen & \vsecond{.0421} [.0301,.0542] & \vsecond{.0836} [.0657,.1014] & \vbest{.3025} [.2850,.3206] & \vbest{.5938} [.5589,.6289]\\
Mistral Small 3.2 & Mistral & .0371 [.0248,.0500] & \vbest{.0944} [.0754,.1143] & .2821 [.2572,.3060] & \vsecond{.5499} [.5043,.5951]\\
DeepSeek V3.2 & DeepSeek & .0316 [.0169,.0464] & .0789 [.0589,.0992] & \vsecond{.2977} [.2832,.3125] & .5208 [.4985,.5440]\\
GLM-5 & Z.ai & .0411 [.0307,.0522] & .0698 [.0538,.0863] & .2539 [.2420,.2658] & .4776 [.4562,.4989]\\
\bottomrule
\end{tabular*}
\end{table*}

\begin{table*}[t]
\caption{Absolute MuSiQue state scores and cost. Runs indicate repeated generation runs. $\Arel$ is the canonical Structured Alignment Score. Cost is the reported self refinement+MAS generation cost per question and run. Reported costs follow the frozen three-seed experiment logs.}
\label{tab:s-states}\centering\small\setlength{\tabcolsep}{3.2pt}
\begin{tabular}{@{}lrrrrrrrrr@{}}\toprule
Model & Runs & $A_{\mathrm{rel},T}^{\Mas}$ & $C_0^{\Mas}$ & $C_T^{\Mas}$ & $G_{\Mas}$ & $G_{\Self}$ & EM (\%) & F1 (\%) & USD\\\midrule
Claude Haiku 4.5 & 3 & 0.8576 & 0.2362 & 0.2408 & +0.0046 & -0.0190 & 15.33 & 22.78 & 0.0311 \\
GPT-5.6 Terra & 3 & 0.8847 & 0.2814 & 0.2758 & -0.0056 & -0.0478 & 17.00 & 23.60 & 0.0372 \\
Gemini 3.1 Flash-Lite & 3 & 0.8880 & 0.3219 & 0.3303 & +0.0085 & -0.0132 & 33.00 & 45.62 & 0.0098 \\
Mistral Small 3.2 & 3 & 0.8575 & 0.3109 & 0.3299 & +0.0190 & -0.0181 & 5.33 & 17.25 & 0.0012 \\
Qwen3.5-35B-A3B & 3 & 0.8596 & 0.2846 & 0.3023 & +0.0177 & -0.0245 & 35.67 & 45.75 & 0.0047 \\
\bottomrule\end{tabular}\end{table*}

\begin{table*}[t]
\caption{MuSiQue seed-level outcomes for Haiku, Mistral, and Terra. $\Delta$F1 is shown on the 0--1 scale here. $\Delta T_{\rm gold}$ is the matched recovery-change contrast. These runs reuse question identities, so the main intervals bootstrap seed-averaged questions.}
\label{tab:s-seeds}\centering\small\setlength{\tabcolsep}{6pt}
\begin{tabular}{@{}llrrrrrr@{}}\toprule
Model & Seed & $G_{\Mas}$ & $G_{\Self}$ & $E_\Delta$ & $\Enon$ & $\Delta$F1 & $\Delta T_{\rm gold}$\\\midrule
\multicolumn{8}{l}{\textbf{MuSiQue}}\\
Claude Haiku 4.5 & 7 & +0.0101 & -0.0207 & 0.0308 & 0.0538 & -0.0191 & +0.2100\\
Claude Haiku 4.5 & 42 & +0.0021 & -0.0183 & 0.0204 & 0.0439 & -0.0058 & +0.1817\\
Claude Haiku 4.5 & 123 & +0.0017 & -0.0180 & 0.0197 & 0.0448 & +0.0696 & +0.1883\\
Mistral Small 3.2 & 7 & +0.0272 & -0.0186 & 0.0457 & 0.1143 & +0.0577 & +0.2817\\
Mistral Small 3.2 & 42 & +0.0206 & -0.0210 & 0.0416 & 0.0992 & +0.0841 & +0.2667\\
Mistral Small 3.2 & 123 & +0.0091 & -0.0147 & 0.0238 & 0.0698 & +0.0744 & +0.2467\\
GPT-5.6 Terra & 7 & +0.0000 & -0.0534 & 0.0534 & 0.1016 & -0.0254 & +0.1433\\
GPT-5.6 Terra & 42 & -0.0055 & -0.0415 & 0.0360 & 0.0714 & -0.0416 & +0.1333\\
GPT-5.6 Terra & 123 & -0.0115 & -0.0485 & 0.0371 & 0.0763 & -0.0085 & +0.1683\\
\bottomrule\end{tabular}\end{table*}

\subsection{Absolute Scores and Task Response}
Table~\ref{tab:s-states} gives the complete MuSiQue absolute-state and cost fields for the five-model state analysis. The largest final coverage and the largest interaction-specific gain can belong to different models. Gemini has final coverage 0.3303 and $E_\Delta=0.0217$, while Terra has final coverage 0.2758 and $E_\Delta=0.0422$. Their initial states and self-refinement trajectories contribute to this difference.

Table~\ref{tab:s-model-intervals} reports all seven model families on both primary datasets, with question-bootstrap intervals for the 14 model--task settings. The main-paper cross-family table pairs coverage with F1 and alignment gain with matched task gain. The added DeepSeek and GLM three-seed means and standard deviations are preserved in \nolinkurl{data/model_expansion_deepseek_glm_v14.csv}.

\subsection{Seed-Level Variation}
Table~\ref{tab:s-seeds} reports the archived MuSiQue seed runs for Haiku, Terra, and Mistral. Overall and non-owned contrasts have positive means in all displayed runs. The task responses vary: Haiku has negative $\Delta$F1 in two seeds and a positive value in the third, Terra has negative task gain in all three, and Mistral has positive task gain in all three. Semantic contrast and answer response are reported as separate outcomes. The full primary-set model intervals average seeds within each question before resampling question identities, as specified in \appref{protocol}.

\subsection{Answer Saturation on Corrected 2Wiki}
\label{sec:s-saturation}
The question-level audit covers 58 held-out questions and three seeds, giving 174 question--seed units for each model. Terra and DeepSeek have MAS F1 $=1$ and self-refinement F1 $=0$ in every unit. Consequently, the main table reports 100\% MAS F1 and $+100$ percentage points for $\Delta$F1. Their seed standard deviations are zero, and the empirical question-bootstrap intervals for these saturated means collapse to a single value. These intervals describe the observed protocol and question set. Within-model correlations involving their constant F1 or $\Delta$F1 profiles are undefined.

\noindent\textbf{Information gaps and abstention.}
Each agent receives one of the three complementary evidence facts. The retained comparison questions require facts from multiple sources, while the self-refinement prompt requests \texttt{unknown} when evidence is insufficient. The audited self-refinement outputs predominantly abstain, and the final MAS plurality answers are all correct. For the comparison between \emph{Two Evil Eyes} and \emph{The Marquise of Clermont}, isolated agents each know one director. Interaction combines the director facts, yielding two directors for the former and one for the latter, which resolves the comparison.

\noindent\textbf{Scoring and runtime audit.}
The audit found no evidence of a percentage-scaling error, empty outputs receiving correct-answer credit, or a plotting artifact. DeepSeek reproduced the outcome with a strictly label-free generation runtime. Terra's earlier runner added \texttt{gold\_answer} to the stored record after the API response for offline scoring. Prompt construction excluded this field, so model inputs did not contain the label. This record-handling distinction is relevant to reproductions that also require the generation process itself to avoid reading gold labels.

\noindent\textbf{Interpretation.}
The floor--ceiling contrast supports controlled evidence integration and recovery of peer information. Final-answer F1 evaluates a short answer, while Semantic Coverage assesses intermediate reference facts and the conclusion within each agent's reasoning. Full answer credit can therefore coexist with partial coverage. Interaction-specific alignment is assessed through $E_\Delta$, the ownership decomposition, and independently computed gold recovery. The audit findings summarized here were supplied with the experimental revision. The underlying question-level trajectory archive is maintained separately from this typesetting project.

\section{Gold-Recovery Validation and Alternative Scorers}
\label{app:validity}
Table~\ref{tab:s-validity-intervals} reports the complete intervals for the trajectory-validity correlations summarized in the main paper.
\begin{table*}[t]
\caption{\textbf{Full intervals for trajectory-validity comparisons.} Pearson $r$ and 95\% question-bootstrap intervals for the primary sets. Scoring definitions and sample membership match the merged main-paper table. $E_{\rm rel,\Delta}$ applies the matched trajectory contrast to the Structured Alignment Score.}
\label{tab:s-validity-intervals}
\centering\small\setlength{\tabcolsep}{3.5pt}
\begin{tabular*}{\textwidth}{@{\extracolsep{\fill}}llcccc@{}}
\toprule
\multicolumn{6}{@{}l}{\textbf{A. Correlation with matched answer improvement $\Delta$F1}}\\
Dataset & Model & $E_{\rm rel,\Delta}$ & $G_{\Mas}$ & $E_\Delta$ & $\Enon$\\
\midrule
MuSiQue & Gemini & $.183\ [-.033,.367]$ & $.229\ [.015,.437]$ & $.237\ [.033,.424]$ & $.299\ [.053,.507]$\\
MuSiQue & Qwen & $-.034\ [-.232,.165]$ & $.405\ [.250,.543]$ & $.352\ [.186,.510]$ & $.294\ [.132,.449]$\\
2Wiki & Gemini & $-.368\ [-.555,-.134]$ & $.210\ [-.025,.439]$ & $.204\ [-.021,.429]$ & $.214\ [-.006,.449]$\\
2Wiki & Qwen & $-.220\ [-.485,.031]$ & $.156\ [-.087,.372]$ & $.176\ [-.058,.395]$ & $.132\ [-.107,.346]$\\
\addlinespace
\multicolumn{6}{@{}l}{\textbf{B. Correlation with independently computed gold-recovery change $\Delta T_{\rm gold}$}}\\
Dataset & Model & \multicolumn{2}{c}{$G_{\Mas}$} & $E_\Delta$ & $\Enon$\\
\midrule
MuSiQue & Gemini & \multicolumn{2}{c}{$.399\ [.229,.551]$} & $.471\ [.286,.630]$ & $.525\ [.343,.666]$\\
MuSiQue & Qwen & \multicolumn{2}{c}{$.541\ [.367,.681]$} & $.562\ [.384,.708]$ & $.664\ [.515,.782]$\\
2Wiki & Gemini & \multicolumn{2}{c}{$.592\ [.378,.760]$} & $.551\ [.326,.728]$ & $.612\ [.415,.773]$\\
2Wiki & Qwen & \multicolumn{2}{c}{$.602\ [.427,.742]$} & $.591\ [.419,.734]$ & $.606\ [.443,.744]$\\
\bottomrule
\end{tabular*}
\end{table*}

\subsection{Deterministic Gold-Item Recovery}
The independent lexical diagnostic operates on each agent's emitted \texttt{steps + conclusion}. For MuSiQue it searches canonical subanswers and metadata aliases with at least three alphanumeric characters, after deterministic normalization. For 2Wiki it searches the official triple object/value. Initial and final recovery are measured under both conditions, and the primary $\Delta T_{\rm gold}$ is the MAS-minus-self-refinement difference in recovery changes.

A gold value mention is a recovery proxy. It can undercount paraphrases and can overcount a value appearing with the wrong subject, relation, or polarity. The strict diagnostic additionally requires the subject and object to occur in the same structured unit. This reduces the ambiguity while retaining a fully deterministic scorer. The resulting measure records subject-context co-mention in the emitted response. Table~\ref{tab:s-recovery-sensitivity} reports the canonical and subject-context recovery sensitivities.

\begin{table}[t]
\caption{Canonical/strict recovery sensitivity for $r(\Enon,\Delta T_{\rm gold})$. MuSiQue canonical matching excludes aliases. 2Wiki strict matching adds a subject-context requirement.}
\label{tab:s-recovery-sensitivity}
\centering\small\setlength{\tabcolsep}{3pt}
\begin{tabular}{@{}lll@{}}\toprule
Setting & Variant & Pearson [95\% CI]\\\midrule
MuSiQue, Gemini & canonical & $.506\ [.315,.653]$\\
MuSiQue, Qwen & canonical & $.614\ [.477,.739]$\\
2Wiki, Gemini & strict & $.304\ [.021,.551]$\\
2Wiki, Qwen & strict & $.582\ [.426,.718]$\\
\bottomrule\end{tabular}
\end{table}

The NLI-based transfer diagnostic and coverage changes share local semantic comparisons. Their high correlations are therefore reported as internal convergent evidence. The deterministic recovery diagnostic contributes a separate measurement procedure while sharing the same gold targets and matched trajectory design. Together, these diagnostics assess expressed recovery of the reference targets along the matched trajectories.

\subsection{Task-Gain Criteria and Alternative Scorers}
Each alternative scorer averages the best generated-unit match for each reference row, following the primary coverage operation. We substitute token F1, ROUGE-L, or a locally cached MiniLM cosine score and preserve the same time and condition contrasts. The main paper shows the non-owned gold-recovery criterion. Table~\ref{tab:s-taskbaseline} gives the task-gain criterion. The table reports every evaluated scorer in each setting. Point-estimate rankings summarize these task associations, with pairwise significance requiring the corresponding joint resampling analysis.

\begin{table}[t]
\caption{Alternative change scores against $\Delta$F1. Primary sets only. Bold marks the largest point estimate in each column.}
\label{tab:s-taskbaseline}\centering\small\setlength{\tabcolsep}{3.2pt}
\begin{tabular}{@{}lrrrr@{}}\toprule
 & \multicolumn{2}{c}{MuSiQue} & \multicolumn{2}{c}{2Wiki}\\
Scorer & Gemini & Qwen & Gemini & Qwen\\\midrule
Canonical $\Arel$ & .183 & $-.034$ & $-.368$ & $-.220$\\
Token F1 & \textbf{.303} & \textbf{.389} & $-.013$ & $-.078$\\
ROUGE-L & .269 & .362 & $-.080$ & $-.080$\\
MiniLM cosine & .259 & .382 & .053 & .034\\
NLI coverage & .237 & .352 & \textbf{.204} & \textbf{.176}\\
\bottomrule\end{tabular}\end{table}

\subsection{Answer Transitions and Initial-State Diagnostics}
\noindent\textbf{Answer transitions.}
Within each seed, exact match classifies the final self-refinement and MAS answers as correct or wrong. This defines the four transitions C$\to$C, C$\to$W, W$\to$C, and W$\to$W. A question-level majority label is assigned only when one transition has a unique majority across seeds. All questions remain in the continuous question-level coverage and recovery analyses, including those with mixed transition labels.

\noindent\textbf{Initial state and headroom.}
Table~\ref{tab:s-states} gives initial and final coverage under both conditions. The paired trajectory contrast accounts for each condition's initial value. Initial coverage also determines the available room for an absolute increase, while the ownership decomposition separates local retention from recovery of initially non-owned content.

\noindent\textbf{Visible length.}
The fixed output schema specifies the number of reasoning steps. Visible word counts provide an additional description of response length alongside provider-reported tokens. The observed correlations between visible-length change and $\Enon$ are $-0.239$ for Gemini and $-0.156$ for Qwen on MuSiQue, and $-0.032$ and $-0.501$ on 2Wiki. These are descriptive associations for the evaluated model and benchmark combinations.

\section{Depth and Evidence-Allocation Protocol}
\label{app:depth}
\label{sec:s-depth-protocol}
\noindent\textbf{Depth configurations.}
The auxiliary protocol selects mutually disjoint question sets at different reasoning depths, excluding the pilot and original held-out question identifiers. A two-hop configuration assigns one agent a distractor-only paragraph. A four-hop configuration assigns a two-paragraph local bundle to one agent. Reference units and ownership sets follow the assigned evidence. Comparisons across these configurations therefore change question content and evidence allocation together.

\noindent\textbf{Agent-count configurations.}
The group-size protocol holds four supporting paragraphs and the question identities fixed. Two agents receive two paragraphs each. Three agents receive a $2+1+1$ allocation, and four agents receive one paragraph each. A five-agent configuration adds a distractor-only agent to the four supporting-evidence owners. Each condition retains three response states. Increasing the number of agents changes the total call budget, evidence fragmentation, and the ownership sets used for non-owned coverage.

\noindent\textbf{Matched comparison.}
Within each fixed configuration, self refinement and MAS use the same questions, evidence allocation, and response schedule. The corresponding trajectory contrast evaluates the contribution of communication under that allocation. The primary results in the paper use the fixed configuration specified in the execution protocol.

\section{GAP Architecture: Protocol and Comparisons}
\label{app:gap-protocol}
\label{sec:s-gap-protocol}
\subsection{Architecture State and Runtime Inputs}
The GAP study uses three agents, two synchronous updates after state zero, and seeds 7, 42, and 123. Each initial response emits a local fact, information-gap queries, three reasoning steps, a conclusion, and an answer. The state-zero record is physically generated once per model, question, seed, and agent, then reused by every architecture condition. The source evidence remains assigned to its original agent. The runtime directory exposes the agents' generated local facts. Gold references, aliases, labels, and evaluation scores enter only offline scoring.

The local-fact field is emitted alongside reasoning in the same initial model response. The program extracts and stores that field directly. It does not invoke a separate LLM to summarize a completed trace. This initial-generation cost is shared across architecture conditions. Later receiver calls are ordinary reasoning updates using the selected input fields.

The fact directory stores each sender's initial local fact. A receiver's current gaps are embedded with the cached MiniLM model and compared with peer facts by cosine similarity. The highest-scoring fact for each gap is selected, followed by deduplication and a cap of two facts. An empty gap set selects no facts. The candidates stay fixed across states. The receiver's queries and memory can change. The design does not recursively forward acquired claims as new source facts.

The local NLI gate assigns the most probable class to each comparison between a candidate and a memory item. A contradiction with any memory item takes priority and quarantines the candidate. Otherwise, entailment with an item marks a duplicate. All remaining candidates are accepted. GAP-Transfer and GAP-SOFT check the current reasoning-derived memory, while hard GAP-MAS checks the persistent anchor and accepted peer facts. This memory distinction is part of the hard treatment. The source reports log routing decisions and gate actions separately.

\subsection{Condition Definitions and Matched Controls}
\begin{table*}[t]
\caption{\textbf{Architecture treatments and self refinement controls.} All rows use the same initial records, three agents, and two updates. The description summarizes the reported runtime semantics.}
\label{tab:s-gap-conditions}\centering\small
\begin{tabularx}{\textwidth}{@{}lXXX@{}}\toprule
Condition & Peer input & Local treatment & Matched control\\\midrule
Self refinement (plain) & No peer messages & Frozen independent-update prompt, including original local evidence & Baseline\\
Self refinement (anchored) & No peer messages & Persistent anchor and preservation instruction & Baseline\\
Full-MAS & Previous-round peer reasoning, conclusion and answer & Full-reasoning update & self refinement (plain)\\
GAP-Transfer & Accepted gap-relevant initial peer facts & Selective update with current reasoning & self refinement (plain)\\
GAP-SOFT & Same router/gate as GAP-Transfer & Original local evidence and soft retention instruction & self refinement (plain), as reported\\
Hard GAP-MAS & Gap-relevant facts. Anchor-based gate memory & Persistent local anchor and explicit preservation & self refinement (anchored)\\
\bottomrule\end{tabularx}\end{table*}

Table~\ref{tab:s-gap-conditions} preserves the reported control assignments. The GAP-SOFT analysis reused self refinement (plain) after confirming that its frozen prompt re-shows original evidence. The soft instruction remains part of the intervention. Accordingly, the soft-versus-transfer comparison evaluates the complete soft treatment, including its local-evidence presentation and later state changes. The hard-versus-transfer comparison includes memory and control differences as well as preservation wording.

For method $m$, write $b(m)$ for its reported matched self refinement control. The architecture contrasts are $E_\Delta(m)=G_m-G_{b(m)}$, with the same definition for the non-owned and gold-recovery contrasts. Physically shared initial states cancel exactly. A difference between methods with different controls decomposes as
\begin{equation}
 E_\Delta(a)-E_\Delta(b)
 =(G_a-G_b)-(G_{b(a)}-G_{b(b)}).
\label{eq:s-control-difference}
\end{equation}
Thus the hard variant's interaction contrast and direct task performance are separate comparisons. Main-table F1 is the direct outcome. Supplementary $\Delta$F1 retains the matching baseline stated in the report. Reference scoring continues to use only emitted steps and conclusions. Gap queries, local-anchor fields, accepted-fact metadata, and sender identifiers are excluded from score inputs.

\subsection{Common-Subset Execution}
The reported retention comparison uses Qwen and Gemini on the primary MuSiQue and corrected 2Wiki sets. GAP-SOFT generates the new soft condition while reusing the other conditions and their shared initial records. Its incremental generation cost is \$2.6198. Every displayed retention comparison uses this common two-model population.

These runs use a shared-start protocol with initial local-fact and gap fields. The earlier cross-family study uses its own generation schema. The architecture results are reported separately and evaluated within their matched shared-start protocol.

\subsection{Accounting and Paired Comparisons}
Communication tokens count peer payload delivered to agents. Total tokens include complete input and output usage, including the treatment-specific context. Model selection, local embedding work, and NLI gate work are kept distinct from the quoted API token and dollar totals. Paired effects average seeds within question and resample question clusters 10,000 times. The combined follow-up weights the 100 MuSiQue and 58 2Wiki questions by their sample counts, with the same two models per question.

All paired treatment contrasts and their pointwise intervals are reported in \appref{gap-results}. The method comparisons retain their stated self-refinement controls and report answer F1 separately from the ownership-sensitive coverage changes.

\section{GAP Retention Results and Paired Effects}
\label{app:gap-results}
\label{sec:s-gap-results}
\begin{table*}[t]
\caption{\textbf{Complete follow-up results on the common Qwen/Gemini subset.} Three seeds on the primary MuSiQue and 2Wiki sets. F1 is a percentage and $\Delta$F1 is in percentage points. Hard GAP-MAS uses self refinement (anchored). Other methods use self refinement (plain). Token and cost values are per question/run.}
\label{tab:s-gap-complete}\centering\small\setlength{\tabcolsep}{4pt}
\begin{tabular}{@{}lrrrrrrrrrr@{}}\toprule
Method & $G^{\rm own}$ & $G^{\rm non}$ & $E_\Delta$ & $\Enon$ & $\Delta T_{\rm gold}$ & F1 & $\Delta$F1 & Peer tok. & Total tok. & USD\\\midrule
\multicolumn{11}{l}{\textbf{MuSiQue}}\\
FULL-MAS & -0.0709 & 0.0727 & 0.0530 & 0.1091 & 0.2017 & 37.08 & 2.64 & 1725.7 & 9029.0 & 0.0039\\
GAP-TRANSFER & -0.0490 & 0.0641 & 0.0517 & 0.1005 & 0.1781 & 35.99 & 1.55 & 435.4 & 7413.1 & 0.0036\\
GAP-SOFT & -0.0014 & 0.0556 & 0.0588 & 0.0920 & 0.1722 & 43.45 & 9.01 & 585.9 & 8550.3 & 0.0038\\
hard GAP-MAS & -0.0451 & 0.0731 & 0.0434 & 0.0744 & 0.1606 & 44.98 & 10.43 & 428.8 & 7706.8 & 0.0037\\
\addlinespace
\multicolumn{11}{l}{\textbf{2Wiki}}\\
FULL-MAS & -0.0701 & 0.1897 & 0.0432 & 0.1936 & 0.3309 & 27.51 & 12.73 & 1421.9 & 7574.8 & 0.0034\\
GAP-TRANSFER & 0.0290 & 0.1244 & 0.0350 & 0.1283 & 0.1638 & 31.12 & 16.34 & 275.7 & 5865.3 & 0.0030\\
GAP-SOFT & 0.2236 & 0.0289 & 0.0324 & 0.0327 & 0.0402 & 13.73 & -1.06 & 441.2 & 6095.0 & 0.0031\\
hard GAP-MAS & 0.1238 & 0.0406 & -0.0163 & -0.0219 & -0.0398 & 17.38 & -2.59 & 301.2 & 6161.1 & 0.0031\\
\addlinespace
\bottomrule\end{tabular}\end{table*}

\begin{table*}[t]
\caption{\textbf{Matched self refinement controls in the follow-up.} G: Gemini. Q: Qwen. Scores and tokens are reported per question/run. A zero interaction contrast for a self refinement row is a baseline convention.}
\label{tab:s-gap-self}\centering\small\setlength{\tabcolsep}{4pt}
\begin{tabular}{@{}lllrrrrrrr@{}}\toprule
Scope & Model & self refinement & $G^{\rm own}$ & $G^{\rm non}$ & $G$ & EM (\%) & F1 (\%) & Total tokens & Final gaps\\\midrule
2Wiki & G & Anchor & 0.2439 & 0.0955 & 0.1104 & 1.72 & 15.75 & 6395.71 & 0.28\\
2Wiki & G & Plain & 0.2461 & -0.0062 & 0.0602 & 1.72 & 7.63 & 6384.35 & 1.83\\
2Wiki & Q & Anchor & 0.0390 & 0.0295 & 0.0263 & 23.56 & 24.18 & 5075.00 & 2.05\\
2Wiki & Q & Plain & 0.0677 & -0.0015 & 0.0166 & 21.84 & 21.94 & 5043.03 & 2.93\\
MuSiQue & G & Anchor & -0.0438 & -0.0067 & -0.0210 & 29.00 & 38.73 & 7365.07 & 0.99\\
MuSiQue & G & Plain & -0.0417 & -0.0419 & -0.0382 & 29.00 & 38.84 & 8007.80 & 2.25\\
MuSiQue & Q & Anchor & -0.0589 & 0.0042 & -0.0090 & 22.67 & 30.37 & 6701.03 & 3.01\\
MuSiQue & Q & Plain & -0.0156 & -0.0308 & -0.0198 & 22.33 & 30.04 & 7366.00 & 3.87\\
\bottomrule\end{tabular}\end{table*}

\subsection{Common-Subset Architecture Outcomes}
Table~\ref{tab:s-gap-complete} reports every architecture mean on the primary evaluation sets in the GAP-SOFT follow-up. The model population is Qwen and Gemini in all four rows. The self refinement controls appear in Table~\ref{tab:s-gap-self}. Their baseline convention assigns zero interaction gain to self refinement. Absolute owned, non-owned, and overall changes are retained. Direct F1 and matched $\Delta$F1 should be read with the corresponding control assignment.

MuSiQue GAP-SOFT raises F1 from 35.99\% to 43.45\% relative to GAP-Transfer, alongside an owned-change increase from $-0.0490$ to $-0.0014$. On 2Wiki, owned change increases from 0.0290 to 0.2236 while non-owned gain falls from 0.1283 to 0.0327 and F1 from 31.12\% to 13.73\%. This juxtaposition motivates the task-dependent acquisition--retention interpretation in the main text.

\begin{table}[t]
\caption{\textbf{Complete paired follow-up effects: MuSiQue.} Effects are GAP-SOFT minus the comparator. Brackets are question-cluster bootstrap 95\% intervals. $p_+$ is the fraction of bootstrap resamples with positive mean. It is a resampling summary, not a posterior probability. F1 differences use the matched controls.}
\label{tab:s-gap-paired1}\centering\datatablefont\setlength{\tabcolsep}{1.8pt}
\begin{tabular}{@{}llrrr@{}}\toprule
Comparator & Outcome & Effect & 95\% interval & $p_+$\\\midrule
GAP-TRANSFER & $G^{\rm own}$ & $+0.0476$ & $[0.0311, 0.0636]$ & 1.0000\\
GAP-TRANSFER & $E_\Delta$ & $+0.0070$ & $[-0.0019, 0.0160]$ & 0.9365\\
GAP-TRANSFER & $\Enon$ & $-0.0085$ & $[-0.0195, 0.0023]$ & 0.0609\\
GAP-TRANSFER & $\Delta T_{\rm gold}$ & $-0.0058$ & $[-0.0228, 0.0111]$ & 0.2439\\
GAP-TRANSFER & $\Delta$F1 (pp) & $+7.45$ & $[3.75, 11.10]$ & 1.0000\\
GAP-TRANSFER & Peer tokens & $+150.5$ & $[140.2, 161.2]$ & 1.0000\\
\addlinespace
FULL-MAS & $G^{\rm own}$ & $+0.0695$ & $[0.0501, 0.0890]$ & 1.0000\\
FULL-MAS & $E_\Delta$ & $+0.0057$ & $[-0.0065, 0.0182]$ & 0.8152\\
FULL-MAS & $\Enon$ & $-0.0171$ & $[-0.0319, -0.0018]$ & 0.0149\\
FULL-MAS & $\Delta T_{\rm gold}$ & $-0.0294$ & $[-0.0586, 0.0008]$ & 0.0284\\
FULL-MAS & $\Delta$F1 (pp) & $+6.37$ & $[1.91, 10.80]$ & 0.9975\\
FULL-MAS & Peer tokens & $-1139.8$ & $[-1182.2, -1099.0]$ & 0.0000\\
\addlinespace
hard GAP-MAS & $G^{\rm own}$ & $+0.0437$ & $[0.0332, 0.0543]$ & 1.0000\\
hard GAP-MAS & $E_\Delta$ & $+0.0153$ & $[0.0057, 0.0253]$ & 0.9987\\
hard GAP-MAS & $\Enon$ & $+0.0176$ & $[0.0057, 0.0299]$ & 0.9984\\
hard GAP-MAS & $\Delta T_{\rm gold}$ & $+0.0117$ & $[-0.0097, 0.0328]$ & 0.8536\\
hard GAP-MAS & $\Delta$F1 (pp) & $-1.42$ & $[-4.91, 2.07]$ & 0.2172\\
hard GAP-MAS & Peer tokens & $+157.2$ & $[145.5, 169.2]$ & 1.0000\\
\bottomrule\end{tabular}\end{table}

\begin{table}[t]
\caption{\textbf{Complete paired follow-up effects: 2Wiki.} Effects are GAP-SOFT minus the comparator. Brackets are question-cluster bootstrap 95\% intervals. $p_+$ is the fraction of bootstrap resamples with positive mean. It is a resampling summary, not a posterior probability. F1 differences use the matched controls.}
\label{tab:s-gap-paired2}\centering\datatablefont\setlength{\tabcolsep}{1.8pt}
\begin{tabular}{@{}llrrr@{}}\toprule
Comparator & Outcome & Effect & 95\% interval & $p_+$\\\midrule
GAP-TRANSFER & $G^{\rm own}$ & $+0.1946$ & $[0.1666, 0.2223]$ & 1.0000\\
GAP-TRANSFER & $E_\Delta$ & $-0.0026$ & $[-0.0132, 0.0081]$ & 0.3087\\
GAP-TRANSFER & $\Enon$ & $-0.0955$ & $[-0.1101, -0.0818]$ & 0.0000\\
GAP-TRANSFER & $\Delta T_{\rm gold}$ & $-0.1236$ & $[-0.1557, -0.0905]$ & 0.0000\\
GAP-TRANSFER & $\Delta$F1 (pp) & $-17.40$ & $[-21.84, -13.07]$ & 0.0000\\
GAP-TRANSFER & Peer tokens & $+165.5$ & $[155.4, 175.8]$ & 1.0000\\
\addlinespace
FULL-MAS & $G^{\rm own}$ & $+0.2937$ & $[0.2652, 0.3229]$ & 1.0000\\
FULL-MAS & $E_\Delta$ & $-0.0108$ & $[-0.0228, 0.0014]$ & 0.0407\\
FULL-MAS & $\Enon$ & $-0.1609$ & $[-0.1785, -0.1435]$ & 0.0000\\
FULL-MAS & $\Delta T_{\rm gold}$ & $-0.2907$ & $[-0.3329, -0.2481]$ & 0.0000\\
FULL-MAS & $\Delta$F1 (pp) & $-13.79$ & $[-17.85, -9.68]$ & 0.0000\\
FULL-MAS & Peer tokens & $-980.7$ & $[-1005.3, -955.5]$ & 0.0000\\
\addlinespace
hard GAP-MAS & $G^{\rm own}$ & $+0.0998$ & $[0.0844, 0.1161]$ & 1.0000\\
hard GAP-MAS & $E_\Delta$ & $+0.0487$ & $[0.0392, 0.0580]$ & 1.0000\\
hard GAP-MAS & $\Enon$ & $+0.0547$ & $[0.0382, 0.0703]$ & 1.0000\\
hard GAP-MAS & $\Delta T_{\rm gold}$ & $+0.0800$ & $[0.0465, 0.1125]$ & 1.0000\\
hard GAP-MAS & $\Delta$F1 (pp) & $+1.53$ & $[-4.50, 7.63]$ & 0.6932\\
hard GAP-MAS & Peer tokens & $+140.0$ & $[128.2, 151.5]$ & 1.0000\\
\bottomrule\end{tabular}\end{table}

\begin{table}[t]
\caption{\textbf{Complete paired follow-up effects: Primary combined.} Effects are GAP-SOFT minus the comparator. Brackets are question-cluster bootstrap 95\% intervals. $p_+$ is the fraction of bootstrap resamples with positive mean. It is a resampling summary, not a posterior probability. F1 differences use the matched controls.}
\label{tab:s-gap-paired3}\centering\datatablefont\setlength{\tabcolsep}{1.8pt}
\begin{tabular}{@{}llrrr@{}}\toprule
Comparator & Outcome & Effect & 95\% interval & $p_+$\\\midrule
GAP-TRANSFER & $G^{\rm own}$ & $+0.1016$ & $[0.0836, 0.1200]$ & 1.0000\\
GAP-TRANSFER & $E_\Delta$ & $+0.0035$ & $[-0.0035, 0.0104]$ & 0.8376\\
GAP-TRANSFER & $\Enon$ & $-0.0405$ & $[-0.0514, -0.0296]$ & 0.0000\\
GAP-TRANSFER & $\Delta T_{\rm gold}$ & $-0.0491$ & $[-0.0673, -0.0308]$ & 0.0000\\
GAP-TRANSFER & $\Delta$F1 (pp) & $-1.67$ & $[-5.08, 1.73]$ & 0.1668\\
GAP-TRANSFER & Peer tokens & $+156.0$ & $[148.3, 163.8]$ & 1.0000\\
\addlinespace
FULL-MAS & $G^{\rm own}$ & $+0.1518$ & $[0.1281, 0.1751]$ & 1.0000\\
FULL-MAS & $E_\Delta$ & $-0.0003$ & $[-0.0095, 0.0089]$ & 0.4658\\
FULL-MAS & $\Enon$ & $-0.0699$ & $[-0.0859, -0.0539]$ & 0.0000\\
FULL-MAS & $\Delta T_{\rm gold}$ & $-0.1254$ & $[-0.1563, -0.0944]$ & 0.0000\\
FULL-MAS & $\Delta$F1 (pp) & $-1.03$ & $[-4.52, 2.60]$ & 0.2829\\
FULL-MAS & Peer tokens & $-1081.4$ & $[-1112.1, -1051.0]$ & 0.0000\\
\addlinespace
hard GAP-MAS & $G^{\rm own}$ & $+0.0643$ & $[0.0546, 0.0742]$ & 1.0000\\
hard GAP-MAS & $E_\Delta$ & $+0.0276$ & $[0.0201, 0.0351]$ & 1.0000\\
hard GAP-MAS & $\Enon$ & $+0.0312$ & $[0.0212, 0.0414]$ & 1.0000\\
hard GAP-MAS & $\Delta T_{\rm gold}$ & $+0.0367$ & $[0.0179, 0.0557]$ & 0.9999\\
hard GAP-MAS & $\Delta$F1 (pp) & $-0.34$ & $[-3.48, 2.78]$ & 0.4165\\
hard GAP-MAS & Peer tokens & $+150.9$ & $[142.2, 159.8]$ & 1.0000\\
\bottomrule\end{tabular}\end{table}

\subsection{Paired Retention Effects}
Tables~\ref{tab:s-gap-paired1}, \ref{tab:s-gap-paired2}, and~\ref{tab:s-gap-paired3} report all paired retention contrasts. For GAP-SOFT versus GAP-Transfer, the MuSiQue non-owned interval is $[-0.0195,\allowbreak0.0023]$. On 2Wiki it is $[-0.1101,\allowbreak-0.0818]$. The pooled non-owned effect is $-0.0405$, with interval $[-0.0514,\allowbreak-0.0296]$. The pooled owned effect is $0.1016$, with interval $[0.0836,\allowbreak0.1200]$. Against Full-MAS, the pooled overall gain difference is $-0.0003$, with interval $[-0.0095,\allowbreak0.0089]$.

The direct soft-versus-hard F1 means differ from their matched $\Delta$F1 contrast because the controls differ. For example, on held-out 2Wiki the raw F1 difference is $13.73-17.38=-3.65$ points, whereas the reported matched-change difference is $+1.53$ points. Equation~\eqref{eq:s-control-difference} explains this arithmetic distinction. The main text uses direct task outcomes and labels matched contrasts explicitly.

\begin{table*}[t]
\caption{\textbf{Routing and gate agreement with GAP-Transfer.} Each cell is a receiver update. The first update is identical. Later reasoning and gaps allow second-update decisions to diverge.}
\label{tab:s-gap-routing}\centering\small
\begin{tabular}{@{}llrrrr@{}}\toprule
Scope & Model & State & Receiver cells & Exact match & Divergence\\\midrule
2Wiki & Gemini & 1 & 522 & 1.0000 & 0.0000\\
2Wiki & Gemini & 2 & 522 & 0.5709 & 0.4291\\
2Wiki & Qwen & 1 & 522 & 1.0000 & 0.0000\\
2Wiki & Qwen & 2 & 522 & 0.4310 & 0.5690\\
MuSiQue & Gemini & 1 & 900 & 1.0000 & 0.0000\\
MuSiQue & Gemini & 2 & 900 & 0.7011 & 0.2989\\
MuSiQue & Qwen & 1 & 900 & 1.0000 & 0.0000\\
MuSiQue & Qwen & 2 & 900 & 0.4333 & 0.5667\\
\bottomrule\end{tabular}\end{table*}

\subsection{Routing Agreement and Later Divergence}
Table~\ref{tab:s-gap-routing} reports first- and second-update agreement. Every first-update comparison has exact agreement. The second update differs in 29.89\% of MuSiQue/Gemini receiver cells and 56.67\% of MuSiQue/Qwen cells. Held-out 2Wiki gives 42.91\% and 56.90\%. Changes to both gap queries and reasoning-derived gate memory can contribute to later decisions.

\noindent\textbf{Treatment interpretation.}
The soft treatment changes local-evidence presentation and retention instructions. The hard treatment also changes gate memory and its matched self-refinement control. The paired results describe these executed policy differences and the corresponding redistribution of expressed coverage.

\section{Jev Controller: Execution Protocol}
\label{app:jev-protocol}
\label{sec:s-jev-protocol}
\subsection{Shared-Start Extension}
The Jev experiment extends the Qwen/Gemini architecture study on MuSiQue and repaired 2Wiki, with seeds 7, 42, and 123. self refinement (plain), Full-MAS, GAP-Transfer, and every physically shared state-zero response are reused read-only. Each agent performs two synchronous updates. Jev 1.13.0 is the communication controller. Qwen and Gemini remain the reasoning models. The new conditions share the GAP-Transfer integration prompt and self refinement (plain) control. Their generation-time state contains the question, local evidence or accepted facts, current gaps, and candidate peer facts. References, gold labels, aliases, and evaluation scores enter the offline scorer.

The directory contains peers' original model-generated local fact cards. It does not retrieve the gold reference on behalf of the controller. For each receiver gap, all available peer candidates are evaluated. The three-agent setup has two distinct peer cards. Selection deduplicates candidates across gaps. Repeated decisions can reuse the append-only cache when the full controller state and decision specification agree.

\subsection{Typed Outputs and Confidence Semantics}
Each candidate is evaluated with three binary Noul questions---relevance, novelty, and conflict---and one Choice over SEND, SKIP, and QUARANTINE. Choice returns the selected action, option probabilities, and the official action-confidence statistic. Noul returns a probability without a separate confidence field~\cite{typesafe2026jev,typesafe2026confidence}. The binary diagnostic certainty $|2p-1|$ is stored separately. The fallback threshold uses only the official Choice confidence.

JEV-GAP sends up to two SEND candidates, ordered by SEND probability, and withholds SKIP and QUARANTINE candidates in its selective path. JEV-GAP-FALLBACK tests whether a gap has at least one SEND candidate with confidence at least 0.70. When this condition holds, it uses selective delivery. Otherwise it supplies the compact peer fact cards. The fallback is triggered by the absence of a confident SEND, which includes states where other actions have high confidence. The protocol broadens delivery within the compact-card pool. The supplementary source report and execution task retain the specified fallback behavior. Exact delivered-card logs are part of the experiment archive.

The binary diagnostics are parallel judgments of the supplied state. The controller study does not introduce a learned numerical utility function or a second NLI screening layer after the Choice decision. The evaluated change is the substitution of typed decision control for the heuristic relevance/NLI routing policy.

\subsection{Scoring, Pairing, and Accounting}
The frozen \texttt{roberta-large-mnli} scorer evaluates emitted reasoning steps and conclusions. For each method, coverage and independent gold-item recovery are measured at initial and final states and contrasted with self refinement (plain). Physically shared initial states cancel in comparisons between these methods. Reported $G^{\rm own}$ is the method's absolute initial-to-final owned change. $\Enon$ and $\Delta T_{\rm gold}$ are matched self refinement contrasts.

Seeds are averaged within question. Paired bootstrap inference uses 10,000 question resamples and retains model pairing. Combined estimates weight the 100 MuSiQue and 58 2Wiki questions by their counts. Their model population is the common Qwen/Gemini pair. The report supplies pooled benchmark means and paired effects. Its model-specific raw trajectories remain in the experiment archive. Direct intervals for fallback versus Full-MAS or GAP-Transfer are not derived by adding intervals from two other comparisons. The supplied tables retain the originally reported contrasts.

Delivered peer-message tokens count the payload shown to reasoning agents. Controller input tokens are recorded separately. The experiment reports 15,684 unique Jev calls, 12,299,084 Jev input tokens, \$0.516562 in Jev charges, and \$4.4267 in new LLM generation charges, for \$4.9433 in new API cost. Mean Jev API latency is 604.22 milliseconds per unique call. The 9,476.54 seconds obtained by summing serial call durations is a workload total, separate from per-question workflow latency. Historical baseline cost and latency records are reused as reported. The source summary does not provide a fresh contemporaneous baseline timing run or a full controller-traffic cost allocation per method.

\subsection{Policy Comparisons}
The controller variants are evaluated against GAP-Transfer and Full-MAS, with direct paired contrasts between selective and fallback delivery. The additional Compact-All and Random-K controls are defined in \appref{comm-protocol} and evaluated in \appref{comm-results}. These comparisons assess the executed communication policies using the same initial cards and reasoning-update schedule.

\section{Complete Jev-GAP Results}
\label{app:jev-results}
\label{sec:s-jev-results}
\begin{table}[t]
\caption{\textbf{Complete Jev-GAP quality results.} Original report precision. F1 and matched $\Delta$F1 are on the 0--1 scale. All rows use the Qwen/Gemini architecture population. Combined results weight questions by benchmark sample count.}
\label{tab:s-jev-main}\centering\datatablefont\setlength{\tabcolsep}{1.2pt}
\begin{tabular*}{\columnwidth}{@{\extracolsep{\fill}}lrrrrrr@{}}\toprule
Method & $E_\Delta$ & $\Enon$ & $G^{\rm own}$ & $\Delta T_{\rm gold}$ & F1 & $\Delta$F1\\\midrule
\multicolumn{7}{@{}l}{\textbf{MuSiQue}}\\
\shortstack[l]{Self refinement\\(plain)} & 0.00000 & 0.00000 & -0.02863 & 0.00000 & 0.34442 & 0.00000\\
Full-MAS & 0.05302 & 0.10907 & -0.07093 & 0.20167 & 0.37078 & 0.02636\\
GAP-Transfer & 0.05173 & 0.10046 & -0.04904 & 0.17806 & 0.35995 & 0.01552\\
Jev-GAP & 0.06011 & 0.09717 & -0.01270 & 0.17694 & 0.45289 & 0.10846\\
Jev-GAP-FB & 0.08547 & 0.14243 & -0.00625 & 0.30139 & 0.49197 & 0.14755\\
\addlinespace[3pt]
\multicolumn{7}{@{}l}{\textbf{2Wiki}}\\
\shortstack[l]{Self refinement\\(plain)} & 0.00000 & 0.00000 & 0.15690 & 0.00000 & 0.14784 & 0.00000\\
Full-MAS & 0.04324 & 0.19360 & -0.07013 & 0.33094 & 0.27512 & 0.12728\\
GAP-Transfer & 0.03498 & 0.12829 & 0.02897 & 0.16379 & 0.31124 & 0.16340\\
Jev-GAP & 0.10114 & 0.17179 & 0.21381 & 0.28544 & 0.30113 & 0.15329\\
Jev-GAP-FB & 0.10647 & 0.20173 & 0.17899 & 0.35489 & 0.28646 & 0.13861\\
\addlinespace[3pt]
\multicolumn{7}{@{}l}{\textbf{Primary combined (100 + 58 questions)}}\\
\shortstack[l]{Self refinement\\(plain)} & 0.00000 & 0.00000 & 0.03948 & 0.00000 & 0.27226 & 0.00000\\
Full-MAS & 0.04943 & 0.14010 & -0.07063 & 0.24912 & 0.33567 & 0.06341\\
GAP-Transfer & 0.04558 & 0.11068 & -0.02040 & 0.17282 & 0.34207 & 0.06981\\
Jev-GAP & 0.07517 & 0.12456 & 0.07045 & 0.21677 & 0.39718 & 0.12492\\
Jev-GAP-FB & 0.09318 & 0.16420 & 0.06175 & 0.32103 & 0.41653 & 0.14427\\
\addlinespace[3pt]
\bottomrule\end{tabular*}\end{table}

\subsection{Architecture Outcomes}
Table~\ref{tab:s-jev-main} reproduces all reported primary means, including self refinement (plain). Full-MAS, GAP-Transfer, and both Jev variants share the architecture-specific initial states and control. JEV-GAP-FALLBACK has pooled $E_\Delta=0.09318$, $\Enon=0.16420$, and $\Delta T_{\rm gold}=0.32103$. Its F1 is 0.41653. Mean delivered peer payload is 329.175 tokens per question and run. Benchmark values in the same table show how the combined result is composed.

\begin{table}[t]
\caption{\textbf{Complete Jev-GAP runtime results.} Delivered peer tokens, total API cost in USD, reported workflow latency in milliseconds, and fallback fraction. Each value is a mean per question/run.}
\label{tab:s-jev-runtime}\centering\datatablefont
\setlength{\tabcolsep}{2pt}\renewcommand{\arraystretch}{1.06}
\begin{tabular*}{\columnwidth}{@{\extracolsep{\fill}}lrrrr@{}}\toprule
Method & Peer tok. & Cost (\$) & Time (ms) & Fallback\\\midrule
\vband{5}{sectionband}{MuSiQue}
self refinement (plain) & 0.000 & 0.00361 & 12027.10 & 0.00000\\
Full-MAS & 1725.720 & 0.00392 & 12927.20 & 0.00000\\
GAP-Transfer & 435.433 & 0.00363 & 12499.30 & 0.00000\\
Jev-GAP & 133.795 & 0.00404 & 17552.60 & 0.00000\\
Jev-GAP-FB & 399.848 & 0.00411 & 17299.30 & 0.64861\\
\vband{5}{sectionband}{2Wiki}
self refinement (plain) & 0.000 & 0.00302 & 9970.26 & 0.00000\\
Full-MAS & 1421.870 & 0.00339 & 11235.30 & 0.00000\\
GAP-Transfer & 275.678 & 0.00302 & 9867.45 & 0.00000\\
Jev-GAP & 144.543 & 0.00325 & 13714.00 & 0.00000\\
Jev-GAP-FB & 207.325 & 0.00329 & 13966.40 & 0.44540\\
\vband{5}{sectionband}{Primary combined}
self refinement (plain) & 0.000 & 0.00339 & 11272.00 & 0.00000\\
Full-MAS & 1614.180 & 0.00373 & 12306.20 & 0.00000\\
GAP-Transfer & 376.789 & 0.00341 & 11533.20 & 0.00000\\
Jev-GAP & 137.741 & 0.00375 & 16143.50 & 0.00000\\
Jev-GAP-FB & 329.175 & 0.00381 & 16075.90 & 0.57402\\
\bottomrule\end{tabular*}\end{table}

\subsection{Communication, Cost, and Latency}
Table~\ref{tab:s-jev-runtime} reproduces all runtime means. Relative to Full-MAS, the fallback payload is 23.17\% on MuSiQue and 14.58\% on 2Wiki. The weighted combined ratio is 20.39\%. Cost is \$0.00411 on MuSiQue and \$0.00329 on 2Wiki. The corresponding workflow latencies are 17.30 and 13.97 seconds. Fallback rates are 0.64861 and 0.44540, respectively. These rates describe when compact-card access was broadened under the stated action-confidence rule.

\begin{table}[t]
\caption{\textbf{All Jev paired effects: MuSiQue.} Reported method differences and pointwise 95\% intervals. F1 is on the 0--1 scale. $B_+$ is the bootstrap fraction above zero.}
\label{tab:s-jev-paired1}\centering\datatablefont\setlength{\tabcolsep}{1.8pt}
\begin{tabular*}{\columnwidth}{@{\extracolsep{\fill}}lrrrr@{}}\toprule
Outcome & Effect & CI low & CI high & $B_+$\\\midrule
\multicolumn{5}{@{}l}{\textbf{Jev-GAP $-$ GAP-Transfer}}\\
$E_\Delta$ & 0.00838 & 0.00013 & 0.01676 & 0.9768\\
$\Enon$ & -0.00329 & -0.01428 & 0.00783 & 0.2908\\
$G^{\rm own}$ & 0.03634 & 0.01966 & 0.05267 & 0.9999\\
$\Delta T_{\rm gold}$ & -0.00111 & -0.02333 & 0.02000 & 0.4597\\
F1 & 0.09294 & 0.05658 & 0.13164 & 1.0000\\
Peer tokens & -301.638 & -318.275 & -285.564 & 0.0000\\
Cost (USD) & 0.00040 & 0.00037 & 0.00044 & 1.0000\\
Latency (ms) & 5,053.240 & 4,662.330 & 5,429.850 & 1.0000\\
\addlinespace
\multicolumn{5}{@{}l}{\textbf{Jev-GAP $-$ Full-MAS}}\\
$E_\Delta$ & 0.00709 & -0.00425 & 0.01846 & 0.8862\\
$\Enon$ & -0.01189 & -0.02500 & 0.00162 & 0.0402\\
$G^{\rm own}$ & 0.05823 & 0.03822 & 0.07855 & 1.0000\\
$\Delta T_{\rm gold}$ & -0.02472 & -0.05028 & 0.00112 & 0.0312\\
F1 & 0.08210 & 0.03892 & 0.12583 & 0.9998\\
Peer tokens & -1,591.930 & -1,641.120 & -1,542.530 & 0.0000\\
Cost (USD) & 0.00011 & 0.00008 & 0.00015 & 1.0000\\
Latency (ms) & 4,625.310 & 4,282.920 & 4,965.350 & 1.0000\\
\addlinespace
\multicolumn{5}{@{}l}{\textbf{Jev-GAP-FB $-$ Jev-GAP}}\\
$E_\Delta$ & 0.02536 & 0.01852 & 0.03218 & 1.0000\\
$\Enon$ & 0.04526 & 0.03454 & 0.05557 & 1.0000\\
$G^{\rm own}$ & 0.00645 & -0.00321 & 0.01646 & 0.8980\\
$\Delta T_{\rm gold}$ & 0.12444 & 0.10194 & 0.14778 & 1.0000\\
F1 & 0.03909 & 0.00568 & 0.07229 & 0.9883\\
Peer tokens & 266.053 & 244.922 & 286.852 & 1.0000\\
Cost (USD) & 0.00007 & 0.00004 & 0.00010 & 1.0000\\
Latency (ms) & -253.224 & -525.259 & 19.810 & 0.0346\\
\addlinespace
\bottomrule\end{tabular*}\end{table}

\begin{table}[t]
\caption{\textbf{All Jev paired effects: 2Wiki.} Reported method differences and pointwise 95\% intervals. F1 is on the 0--1 scale. $B_+$ is the bootstrap fraction above zero.}
\label{tab:s-jev-paired2}\centering\datatablefont\setlength{\tabcolsep}{1.8pt}
\begin{tabular*}{\columnwidth}{@{\extracolsep{\fill}}lrrrr@{}}\toprule
Outcome & Effect & CI low & CI high & $B_+$\\\midrule
\multicolumn{5}{@{}l}{\textbf{Jev-GAP $-$ GAP-Transfer}}\\
$E_\Delta$ & 0.06617 & 0.05027 & 0.08344 & 1.0000\\
$\Enon$ & 0.04350 & 0.02165 & 0.06523 & 1.0000\\
$G^{\rm own}$ & 0.18484 & 0.15529 & 0.21617 & 1.0000\\
$\Delta T_{\rm gold}$ & 0.12165 & 0.07663 & 0.16523 & 1.0000\\
F1 & -0.01011 & -0.05725 & 0.03743 & 0.3284\\
Peer tokens & -131.135 & -147.426 & -116.712 & 0.0000\\
Cost (USD) & 0.00022 & 0.00020 & 0.00025 & 1.0000\\
Latency (ms) & 3,846.510 & 3,478.520 & 4,220.080 & 1.0000\\
\addlinespace
\multicolumn{5}{@{}l}{\textbf{Jev-GAP $-$ Full-MAS}}\\
$E_\Delta$ & 0.05791 & 0.04364 & 0.07256 & 1.0000\\
$\Enon$ & -0.02181 & -0.04114 & -0.00140 & 0.0193\\
$G^{\rm own}$ & 0.28393 & 0.25404 & 0.31417 & 1.0000\\
$\Delta T_{\rm gold}$ & -0.04550 & -0.09483 & 0.00383 & 0.0359\\
F1 & 0.02601 & -0.02242 & 0.07562 & 0.8559\\
Peer tokens & -1,277.330 & -1,299.680 & -1,255.540 & 0.0000\\
Cost (USD) & -0.00014 & -0.00017 & -0.00012 & 0.0000\\
Latency (ms) & 2,478.620 & 2,104.470 & 2,876.820 & 1.0000\\
\addlinespace
\multicolumn{5}{@{}l}{\textbf{Jev-GAP-FB $-$ Jev-GAP}}\\
$E_\Delta$ & 0.00532 & -0.00262 & 0.01360 & 0.8972\\
$\Enon$ & 0.02994 & 0.01514 & 0.04463 & 0.9999\\
$G^{\rm own}$ & -0.03482 & -0.05243 & -0.01633 & 0.0000\\
$\Delta T_{\rm gold}$ & 0.06944 & 0.04119 & 0.09722 & 1.0000\\
F1 & -0.01468 & -0.05554 & 0.02943 & 0.2520\\
Peer tokens & 62.782 & 54.770 & 71.675 & 1.0000\\
Cost (USD) & 0.00004 & 0.00002 & 0.00006 & 1.0000\\
Latency (ms) & 252.482 & -61.728 & 585.930 & 0.9422\\
\addlinespace
\bottomrule\end{tabular*}\end{table}

\begin{table}[t]
\caption{\textbf{All Jev paired effects: Primary combined (100 + 58 questions).} Reported method differences and pointwise 95\% intervals. F1 is on the 0--1 scale. $B_+$ is the bootstrap fraction above zero.}
\label{tab:s-jev-paired3}\centering\datatablefont\setlength{\tabcolsep}{1.8pt}
\begin{tabular*}{\columnwidth}{@{\extracolsep{\fill}}lrrrr@{}}\toprule
Outcome & Effect & CI low & CI high & $B_+$\\\midrule
\multicolumn{5}{@{}l}{\textbf{Jev-GAP $-$ GAP-Transfer}}\\
$E_\Delta$ & 0.02959 & 0.02048 & 0.03881 & 1.0000\\
$\Enon$ & 0.01389 & 0.00269 & 0.02530 & 0.9916\\
$G^{\rm own}$ & 0.09085 & 0.07174 & 0.11014 & 1.0000\\
$\Delta T_{\rm gold}$ & 0.04395 & 0.02127 & 0.06733 & 1.0000\\
F1 & 0.05511 & 0.02492 & 0.08629 & 0.9999\\
Peer tokens & -239.049 & -256.275 & -221.656 & 0.0000\\
Cost (USD) & 0.00034 & 0.00031 & 0.00037 & 1.0000\\
Latency (ms) & 4,610.260 & 4,324.630 & 4,907.630 & 1.0000\\
\addlinespace
\multicolumn{5}{@{}l}{\textbf{Jev-GAP $-$ Full-MAS}}\\
$E_\Delta$ & 0.02574 & 0.01621 & 0.03566 & 1.0000\\
$\Enon$ & -0.01553 & -0.02635 & -0.00432 & 0.0031\\
$G^{\rm own}$ & 0.14109 & 0.11748 & 0.16558 & 1.0000\\
$\Delta T_{\rm gold}$ & -0.03235 & -0.05643 & -0.00756 & 0.0052\\
F1 & 0.06151 & 0.02842 & 0.09598 & 0.9998\\
Peer tokens & -1,476.440 & -1,515.690 & -1,437.960 & 0.0000\\
Cost (USD) & 0.00002 & -0.00001 & 0.00005 & 0.8943\\
Latency (ms) & 3,837.280 & 3,534.240 & 4,149.160 & 1.0000\\
\addlinespace
\multicolumn{5}{@{}l}{\textbf{Jev-GAP-FB $-$ Jev-GAP}}\\
$E_\Delta$ & 0.01800 & 0.01264 & 0.02358 & 1.0000\\
$\Enon$ & 0.03964 & 0.03100 & 0.04833 & 1.0000\\
$G^{\rm own}$ & -0.00870 & -0.01847 & 0.00083 & 0.0367\\
$\Delta T_{\rm gold}$ & 0.10425 & 0.08615 & 0.12289 & 1.0000\\
F1 & 0.01935 & -0.00724 & 0.04634 & 0.9232\\
Peer tokens & 191.435 & 171.090 & 212.219 & 1.0000\\
Cost (USD) & 0.00006 & 0.00004 & 0.00008 & 1.0000\\
Latency (ms) & -67.585 & -277.405 & 143.196 & 0.2680\\
\addlinespace
\bottomrule\end{tabular*}\end{table}

\subsection{All Paired Contrasts}
Tables~\ref{tab:s-jev-paired1}, \ref{tab:s-jev-paired2}, and~\ref{tab:s-jev-paired3} retain every paired effect from the controller report. The contrasts are JEV-GAP versus GAP-Transfer, JEV-GAP versus Full-MAS, and fallback versus JEV-GAP. Intervals are pointwise bootstrap intervals. The bootstrap fraction above zero summarizes the direction of the resampled means. Displayed intervals and effect sizes preserve the source precision.

The pooled JEV-GAP-minus-GAP-Transfer F1 effect is 0.05511 with interval [0.02492,0.08629]. The corresponding non-owned effect is 0.01389 with interval [0.00269,0.02530]. Fallback increases non-owned gain by 0.03964 (95\% interval: 0.03100--0.04833). Gold-recovery gain increases by 0.10425 (0.08615--0.12289). Its pooled F1 increment is 0.01935 with interval [$-0.00724$,0.04634]. These results quantify acquisition and answer outcomes separately.

\section{Communication-Family Baselines: Protocol and Budgets}
\label{app:comm-protocol}
\label{sec:s-comm-protocol}
The two supplementary baseline reports extend the shared-start Qwen/Gemini architecture study. Their primary scopes are the same 100 MuSiQue questions and repaired 2Wiki, with seeds 7, 42, and 123 and two communication updates after state zero. Baselines and new methods reuse the same initial records read-only. Generation is reference-free. The frozen scorer evaluates emitted reasoning after execution. The integrated tables keep \emph{topology-level routing} and \emph{fact-level communication} as separate families.

\subsection{Compact and Random Controls}
Compact-All broadcasts the two model-generated initial peer fact cards to each receiver at every update. These cards were produced within the shared state-zero calls. Broadcast only reads and serializes the stored peer fields into the update prompt. It adds no summarizer, selector, or controller model call. Each receiver still makes its scheduled reasoning-update call. The initial cards remain unchanged across rounds. It uses the factual integration interface without relevance selection or NLI gating. Random-K samples from the same initial peer-card pool, matching the number delivered by Jev-GAP-Fallback for each question, seed, receiver, and update. The schedule is inherited from the frozen target run. This controls selection content conditional on the target's message-count schedule. It does not randomize when the target requests messages. New Random-K trajectories are generated from the shared initial states.

Message-count matching leaves small token differences because selected cards have different lengths. Random-K uses 97.685\% of fallback's MuSiQue payload and 97.759\% of its 2Wiki payload. The common-state study retains two candidates per receiver. Compact-All transmits both cards. A count-matched random decision is identical to all-card delivery whenever both candidates are selected. These are properties of the evaluated candidate pool.

\subsection{Topology-Level Adaptations}
AgentPrune~\cite{zhang2025agentprune} uses a frozen development-selected mask from the official Bernoulli edge-logit procedure. The reported pruning rate is 0.833333 and retained-edge fraction is 0.166667. Surviving edges carry unchanged Full-MAS reasoning packets. Development communication is 268.05 tokens against a fallback target of 309.25, a mismatch of $-13.32\%$.

The AgentDropout baseline uses the vendored two-stage optimizer for nodes and edges~\cite{wang2025agentdropout}. Its active-agent fraction is 0.6667 and retained-edge fraction is 0.1667, with unchanged Full-MAS packet content. Its reported development budget mismatch is $-16.48\%$. The shared-start adaptation is evaluated at these reported active-agent and retained-edge fractions.

PEAR~\cite{feng2026pear} directly uses the vendored \texttt{pear\_full} routing code at degree one, with retained-edge fraction 0.5 per round. The shared state-zero schema has no confidence field, so the first routing decision uses neutral confidence 3. Subsequent decisions use self-reported confidence. Its development budget mismatch is $+44.34\%$. The reported results describe this fixed-start routing configuration.

\begin{table}[t]
\caption{\textbf{Delivered communication budgets.} Mean peer tokens per question/run and their ratio to Jev-GAP-FB within the same dataset. Every method has both measurements.}
\label{tab:s-family-budgets}\centering\small
\setlength{\tabcolsep}{4pt}\renewcommand{\arraystretch}{1.02}
\begin{tabular*}{\columnwidth}{@{\extracolsep{\fill}}lrr@{}}\toprule
Method & Peer tokens & Ratio to FB\\\midrule
\multicolumn{3}{@{}l}{\textbf{MuSiQue}}\\
Full-MAS & 1725.72 & 4.31594\\
AgentPrune & 289.473 & 0.72396\\
AgentDropout & 272.953 & 0.68264\\
PEAR & 427.983 & 1.07036\\
Compact-All & 407.427 & 1.01895\\
Random-K & 390.59 & 0.97685\\
GAP-Transfer & 435.433 & 1.089\\
Jev-GAP & 133.795 & 0.33461\\
Jev-GAP-FB & 399.848 & 1\\
\addlinespace[3pt]\multicolumn{3}{@{}l}{\textbf{2Wiki}}\\
Full-MAS & 1421.87 & 6.85818\\
AgentPrune & 236.543 & 1.14093\\
AgentDropout & 226.983 & 1.09482\\
PEAR & 421.655 & 2.03379\\
Compact-All & 279.586 & 1.34854\\
Random-K & 202.678 & 0.97759\\
GAP-Transfer & 275.678 & 1.32969\\
Jev-GAP & 144.543 & 0.69718\\
Jev-GAP-FB & 207.325 & 1\\
\addlinespace[3pt]\multicolumn{3}{@{}l}{\textbf{Primary combined}}\\
Full-MAS & 1614.18 & 4.90372\\
AgentPrune & 270.043 & 0.82036\\
AgentDropout & 256.078 & 0.77794\\
PEAR & 425.66 & 1.29311\\
Compact-All & 360.498 & 1.09516\\
Random-K & 321.61 & 0.97702\\
GAP-Transfer & 376.789 & 1.14465\\
Jev-GAP & 137.741 & 0.41844\\
Jev-GAP-FB & 329.175 & 1\\
\bottomrule\end{tabular*}\end{table}

\subsection{Executed Budget Profiles}
Table~\ref{tab:s-family-budgets} reports the complete peer-token and ratio measurements for every method. AgentPrune uses 72.40\% of fallback's MuSiQue payload and 114.09\% of its 2Wiki payload. AgentDropout uses 68.26\% and 109.48\%, while PEAR uses 107.04\% and 203.38\%. Each external method uses one development-selected configuration. The comparison therefore evaluates these realized operating points.

Runtime cost and latency are reported per question/run, reusing historical baseline measurements. The resource comparison follows the reported per-run API and latency accounting. The baseline summaries leave topology-optimization amortization unspecified, so the resource claims refer to the recorded runtime measurements. Paired uncertainty is reported where supplied by the original question-level analysis.

\subsection{Statistics and Comparison Scope}
Metrics are calculated per question/seed, then seeds are averaged within question. Paired intervals use 10,000 question-cluster bootstrap resamples. Combined estimates average the matched models within a question and weight tasks by 100 and 58 questions. Primary analyses retain benchmark-specific effects. Pooled statistics refer to this fixed two-model population.

Compact-All leads the pooled acquisition and F1 means, while Jev-GAP provides the smallest peer payload. The paired tables report acquisition, retention, gold recovery, F1, and resource differences together for the full comparison set.

\section{Complete Communication-Family Results}
\label{app:comm-results}
\label{sec:s-comm-results}
\begin{table}[t]
\caption{\textbf{Complete communication-family quality results.} Source-report precision is retained. Each task averages the same Qwen/Gemini runs and three seeds. Combined means weight the 100 and 58 questions.}
\label{tab:s-family-quality}\centering\datatablefont\setlength{\tabcolsep}{1.2pt}
\begin{tabular*}{\columnwidth}{@{\extracolsep{\fill}}lrrrrr@{}}\toprule
Method & $E_\Delta$ & $\Enon$ & $G^{\rm own}$ & $\Delta T_{\rm gold}$ & F1\\\midrule
\multicolumn{6}{@{}l}{\textbf{MuSiQue}}\\
Full-MAS & 0.05302 & 0.10907 & -0.07093 & 0.20167 & 0.37078\\
\multicolumn{6}{@{}l}{\textit{Topology-level routing}}\\
AgentPrune & 0.02132 & 0.06009 & -0.08082 & 0.09528 & 0.31003\\
AgentDropout & 0.03621 & 0.08286 & -0.06799 & 0.11028 & 0.32086\\
PEAR & 0.00732 & 0.05131 & -0.11862 & 0.08444 & 0.32018\\
\multicolumn{6}{@{}l}{\textit{Fact-level communication}}\\
Compact-All & 0.10394 & 0.17283 & -0.00745 & 0.37778 & 0.54485\\
Random-K & 0.08744 & 0.14547 & -0.00969 & 0.29889 & 0.50038\\
GAP-Transfer & 0.05173 & 0.10046 & -0.04904 & 0.17806 & 0.35995\\
Jev-GAP & 0.06011 & 0.09717 & -0.0127 & 0.17694 & 0.45289\\
Jev-GAP-FB & 0.08547 & 0.14243 & -0.00625 & 0.30139 & 0.49197\\
\midrule
\multicolumn{6}{@{}l}{\textbf{2Wiki}}\\
Full-MAS & 0.04324 & 0.1936 & -0.07013 & 0.33094 & 0.27512\\
\multicolumn{6}{@{}l}{\textit{Topology-level routing}}\\
AgentPrune & 0.02375 & 0.14263 & -0.04637 & 0.2227 & 0.31334\\
AgentDropout & 0.02974 & 0.14866 & -0.03194 & 0.21312 & 0.27533\\
PEAR & 0.00814 & 0.13173 & -0.09552 & 0.23372 & 0.45656\\
\multicolumn{6}{@{}l}{\textit{Fact-level communication}}\\
Compact-All & 0.11315 & 0.34778 & -0.08451 & 0.62596 & 0.32737\\
Random-K & 0.09228 & 0.20165 & 0.12323 & 0.38506 & 0.2721\\
GAP-Transfer & 0.03498 & 0.12829 & 0.02897 & 0.16379 & 0.31124\\
Jev-GAP & 0.10114 & 0.17179 & 0.21381 & 0.28544 & 0.30113\\
Jev-GAP-FB & 0.10647 & 0.20173 & 0.17899 & 0.35489 & 0.28646\\
\midrule
\multicolumn{6}{@{}l}{\textbf{Primary combined}}\\
Full-MAS & 0.04943 & 0.1401 & -0.07063 & 0.24912 & 0.33567\\
\multicolumn{6}{@{}l}{\textit{Topology-level routing}}\\
AgentPrune & 0.02221 & 0.09039 & -0.06818 & 0.14205 & 0.31124\\
AgentDropout & 0.03384 & 0.10702 & -0.05476 & 0.14803 & 0.30415\\
PEAR & 0.00762 & 0.08083 & -0.11014 & 0.13924 & 0.37024\\
\multicolumn{6}{@{}l}{\textit{Fact-level communication}}\\
Compact-All & 0.10732 & 0.23705 & -0.03574 & 0.46888 & 0.46501\\
Random-K & 0.08922 & 0.1661 & 0.03911 & 0.33052 & 0.41658\\
GAP-Transfer & 0.04558 & 0.11068 & -0.0204 & 0.17282 & 0.34207\\
Jev-GAP & 0.07517 & 0.12456 & 0.07045 & 0.21677 & 0.39718\\
Jev-GAP-FB & 0.09318 & 0.1642 & 0.06175 & 0.32103 & 0.41653\\
\bottomrule\end{tabular*}\end{table}

\begin{table}[t]
\caption{\textbf{Complete communication-family runtime results.} Source-report precision is retained. Each task averages the same Qwen/Gemini runs and three seeds. Combined means weight the 100 and 58 questions.}
\label{tab:s-family-runtime}\centering\datatablefont\setlength{\tabcolsep}{1.2pt}
\begin{tabular*}{\columnwidth}{@{\extracolsep{\fill}}lrrr@{}}\toprule
Method & Peer tokens & Cost (\$) & Latency (ms)\\\midrule
\multicolumn{4}{@{}l}{\textbf{MuSiQue}}\\
Full-MAS & 1725.72 & 0.00392 & 12927.2\\
\multicolumn{4}{@{}l}{\textit{Topology-level routing}}\\
AgentPrune & 289.473 & 0.00371 & 13136.1\\
AgentDropout & 272.953 & 0.0029 & 9375.91\\
PEAR & 427.983 & 0.0057 & 24884.3\\
\multicolumn{4}{@{}l}{\textit{Fact-level communication}}\\
Compact-All & 407.427 & 0.00381 & 12354.3\\
Random-K & 390.59 & 0.00379 & 12537.6\\
GAP-Transfer & 435.433 & 0.00363 & 12499.3\\
Jev-GAP & 133.795 & 0.00404 & 17552.6\\
Jev-GAP-FB & 399.848 & 0.00411 & 17299.3\\
\midrule
\multicolumn{4}{@{}l}{\textbf{2Wiki}}\\
Full-MAS & 1421.87 & 0.00339 & 11235.3\\
\multicolumn{4}{@{}l}{\textit{Topology-level routing}}\\
AgentPrune & 236.543 & 0.00309 & 9953.32\\
AgentDropout & 226.983 & 0.00241 & 7095.01\\
PEAR & 421.655 & 0.00496 & 20255.3\\
\multicolumn{4}{@{}l}{\textit{Fact-level communication}}\\
Compact-All & 279.586 & 0.00314 & 10110.8\\
Random-K & 202.678 & 0.00313 & 10194.8\\
GAP-Transfer & 275.678 & 0.00302 & 9867.45\\
Jev-GAP & 144.543 & 0.00325 & 13714\\
Jev-GAP-FB & 207.325 & 0.00329 & 13966.4\\
\midrule
\multicolumn{4}{@{}l}{\textbf{Primary combined}}\\
Full-MAS & 1614.18 & 0.00373 & 12306.2\\
\multicolumn{4}{@{}l}{\textit{Topology-level routing}}\\
AgentPrune & 270.043 & 0.00348 & 11967.7\\
AgentDropout & 256.078 & 0.00272 & 8538.62\\
PEAR & 425.66 & 0.00543 & 23185\\
\multicolumn{4}{@{}l}{\textit{Fact-level communication}}\\
Compact-All & 360.498 & 0.00356 & 11530.7\\
Random-K & 321.61 & 0.00355 & 11677.6\\
GAP-Transfer & 376.789 & 0.00341 & 11533.2\\
Jev-GAP & 137.741 & 0.00375 & 16143.5\\
Jev-GAP-FB & 329.175 & 0.00381 & 16075.9\\
\bottomrule\end{tabular*}\end{table}

Tables~\ref{tab:s-family-quality} and~\ref{tab:s-family-runtime} combine the two new reports with their common baselines. Overlapping cells were checked numerically before integration. They cover nine methods across three scopes: the two primary benchmarks and their combined estimate. The source CSVs preserve decimal precision and all original budget fields.

Compact-All has pooled $E_\Delta=0.10732$, $\Enon=0.23705$, gold-recovery gain 0.46888, and F1 0.46501. Jev-GAP has the minimum pooled peer payload, 137.741 tokens, and owned change 0.07045. PEAR achieves the highest 2Wiki F1, 0.45656, while Compact-All achieves the highest MuSiQue F1, 0.54485. These patterns support content-level and answer-level evaluation as complementary views of the same execution.

\subsection{Random-K, Compact-All, and AgentPrune Contrasts}
\begin{table}[t]
\caption{\textbf{Compact/random/pruning controls --- MuSiQue.} Paired effects, pointwise 95\% intervals, and bootstrap fractions above zero are transcribed from the frozen report. F1 differences use the 0--1 scale.}
\label{tab:s-communication-paired1}\centering\datatablefont\setlength{\tabcolsep}{1.8pt}
\begin{tabular*}{\columnwidth}{@{\extracolsep{\fill}}lrrrr@{}}\toprule
Outcome & Effect & CI low & CI high & \vhead{Boot.\\$>0$}\\\midrule
\multicolumn{5}{@{}l}{\textbf{Jev-GAP-FB $-$ Random-K}}\\
$E_\Delta$ & -0.00197 & -0.00766 & 0.00366 & 0.2427\\
$\Enon$ & -0.00304 & -0.0112 & 0.00509 & 0.2243\\
$G^{\rm own}$ & 0.00344 & -0.00648 & 0.01346 & 0.7548\\
$\Delta T_{\rm gold}$ & 0.0025 & -0.01056 & 0.01611 & 0.6313\\
F1 & -0.00841 & -0.03598 & 0.01954 & 0.2709\\
Peer tokens & 9.25833 & 7.39162 & 11.2483 & 1\\
\addlinespace[3pt]
\multicolumn{5}{@{}l}{\textbf{Jev-GAP-FB $-$ Compact-All}}\\
$E_\Delta$ & -0.01847 & -0.0268 & -0.01038 & 0\\
$\Enon$ & -0.0304 & -0.04331 & -0.01778 & 0\\
$G^{\rm own}$ & 0.0012 & -0.00798 & 0.01042 & 0.6003\\
$\Delta T_{\rm gold}$ & -0.07639 & -0.09222 & -0.06083 & 0\\
F1 & -0.05288 & -0.07912 & -0.02862 & 0\\
Peer tokens & -7.57833 & -19.3187 & 3.95204 & 0.0944\\
\addlinespace[3pt]
\multicolumn{5}{@{}l}{\textbf{Jev-GAP-FB $-$ AgentPrune}}\\
$E_\Delta$ & 0.06415 & 0.05301 & 0.07562 & 1\\
$\Enon$ & 0.08234 & 0.06711 & 0.09787 & 1\\
$G^{\rm own}$ & 0.07457 & 0.05736 & 0.09193 & 1\\
$\Delta T_{\rm gold}$ & 0.20611 & 0.17528 & 0.2375 & 1\\
F1 & 0.18194 & 0.13558 & 0.2312 & 1\\
Peer tokens & 110.375 & 96.3724 & 124.083 & 1\\
\bottomrule\end{tabular*}\end{table}

\begin{table}[t]
\caption{\textbf{Compact/random/pruning controls --- 2Wiki.} Paired effects, pointwise 95\% intervals, and bootstrap fractions above zero are transcribed from the frozen report. F1 differences use the 0--1 scale.}
\label{tab:s-communication-paired2}\centering\datatablefont\setlength{\tabcolsep}{1.8pt}
\begin{tabular*}{\columnwidth}{@{\extracolsep{\fill}}lrrrr@{}}\toprule
Outcome & Effect & CI low & CI high & \vhead{Boot.\\$>0$}\\\midrule
\multicolumn{5}{@{}l}{\textbf{Jev-GAP-FB $-$ Random-K}}\\
$E_\Delta$ & 0.01418 & 0.00613 & 0.02209 & 0.9992\\
$\Enon$ & 8e-05 & -0.01516 & 0.01463 & 0.4964\\
$G^{\rm own}$ & 0.05575 & 0.03537 & 0.07536 & 1\\
$\Delta T_{\rm gold}$ & -0.03017 & -0.0546 & -0.00623 & 0.0073\\
F1 & 0.01436 & -0.02557 & 0.05417 & 0.754\\
Peer tokens & 4.64655 & 3.17529 & 6.15532 & 1\\
\addlinespace[3pt]
\multicolumn{5}{@{}l}{\textbf{Jev-GAP-FB $-$ Compact-All}}\\
$E_\Delta$ & -0.00669 & -0.01724 & 0.00393 & 0.1075\\
$\Enon$ & -0.14604 & -0.16121 & -0.13088 & 0\\
$G^{\rm own}$ & 0.2635 & 0.23301 & 0.29349 & 1\\
$\Delta T_{\rm gold}$ & -0.27107 & -0.30268 & -0.23994 & 0\\
F1 & -0.04091 & -0.09561 & 0.01281 & 0.0686\\
Peer tokens & -72.2615 & -81.3249 & -62.9483 & 0\\
\addlinespace[3pt]
\multicolumn{5}{@{}l}{\textbf{Jev-GAP-FB $-$ AgentPrune}}\\
$E_\Delta$ & 0.08272 & 0.06552 & 0.10098 & 1\\
$\Enon$ & 0.05911 & 0.03428 & 0.08379 & 1\\
$G^{\rm own}$ & 0.22536 & 0.19633 & 0.25569 & 1\\
$\Delta T_{\rm gold}$ & 0.13218 & 0.08764 & 0.1772 & 1\\
F1 & -0.02688 & -0.08571 & 0.03326 & 0.1882\\
Peer tokens & -29.2184 & -38.1524 & -20.1493 & 0\\
\bottomrule\end{tabular*}\end{table}

\begin{table}[t]
\caption{\textbf{Compact/random/pruning controls --- Primary combined.} Paired effects, pointwise 95\% intervals, and bootstrap fractions above zero are transcribed from the frozen report. F1 differences use the 0--1 scale.}
\label{tab:s-communication-paired3}\centering\datatablefont\setlength{\tabcolsep}{1.8pt}
\begin{tabular*}{\columnwidth}{@{\extracolsep{\fill}}lrrrr@{}}\toprule
Outcome & Effect & CI low & CI high & \vhead{Boot.\\$>0$}\\\midrule
\multicolumn{5}{@{}l}{\textbf{Jev-GAP-FB $-$ Random-K}}\\
$E_\Delta$ & 0.00396 & -0.00087 & 0.00884 & 0.9461\\
$\Enon$ & -0.00189 & -0.0093 & 0.00554 & 0.3057\\
$G^{\rm own}$ & 0.02264 & 0.01218 & 0.03323 & 1\\
$\Delta T_{\rm gold}$ & -0.00949 & -0.02198 & 0.00299 & 0.0683\\
F1 & -5e-05 & -0.02319 & 0.02346 & 0.4934\\
Peer tokens & 7.5654 & 6.21411 & 9.01582 & 1\\
\addlinespace[3pt]
\multicolumn{5}{@{}l}{\textbf{Jev-GAP-FB $-$ Compact-All}}\\
$E_\Delta$ & -0.01415 & -0.0207 & -0.00766 & 0\\
$\Enon$ & -0.07285 & -0.08617 & -0.0598 & 0\\
$G^{\rm own}$ & 0.09749 & 0.07512 & 0.12122 & 1\\
$\Delta T_{\rm gold}$ & -0.14786 & -0.16966 & -0.12711 & 0\\
F1 & -0.04848 & -0.07447 & -0.02317 & 0\\
Peer tokens & -31.3228 & -40.803 & -22.2109 & 0\\
\addlinespace[3pt]
\multicolumn{5}{@{}l}{\textbf{Jev-GAP-FB $-$ AgentPrune}}\\
$E_\Delta$ & 0.07097 & 0.06146 & 0.08091 & 1\\
$\Enon$ & 0.07381 & 0.06048 & 0.08767 & 1\\
$G^{\rm own}$ & 0.12993 & 0.1105 & 0.14946 & 1\\
$\Delta T_{\rm gold}$ & 0.17897 & 0.1526 & 0.20534 & 1\\
F1 & 0.10529 & 0.0657 & 0.14668 & 1\\
Peer tokens & 59.1319 & 44.9578 & 72.9146 & 1\\
\bottomrule\end{tabular*}\end{table}

The 54 effects in Tables~\ref{tab:s-communication-paired1}, \ref{tab:s-communication-paired2}, and~\ref{tab:s-communication-paired3} reproduce every required comparison from the first baseline report. Against Random-K, fallback has a non-owned difference of $-0.00189$. Its interval is [$-0.00930$, 0.00554]. The F1 difference is $-0.00005$ with interval [$-0.02319$, 0.02346]. The owned contrast is 0.02264, with interval [0.01218, 0.03323]. These results separate ownership retention from selection-specific acquisition.

Compared with Compact-All, fallback's pooled differences are $-0.01415$ in overall gain, $-0.07285$ in non-owned gain, $-0.14786$ in gold recovery, and $-0.04848$ in F1. All corresponding intervals are below zero. Its owned difference is 0.09749 with interval [0.07512,0.12122], and peer payload is lower by 31.3228 tokens. Compared with AgentPrune, fallback's pooled non-owned gain is higher by 0.07381 with interval [0.06048,0.08767]. Its peer payload is also higher by 59.1319 tokens.

\subsection{AgentDropout and PEAR Contrasts}
\begin{table}[t]
\caption{\textbf{AgentDropout and PEAR contrasts --- MuSiQue.} Paired effects, pointwise 95\% intervals, and bootstrap fractions above zero are transcribed from the frozen report. F1 differences use the 0--1 scale.}
\label{tab:s-topology-paired1}\centering\datatablefont\setlength{\tabcolsep}{1.8pt}
\begin{tabular*}{\columnwidth}{@{\extracolsep{\fill}}lrrrr@{}}\toprule
Outcome & Effect & CI low & CI high & \vhead{Boot.\\$>0$}\\\midrule
\multicolumn{5}{@{}l}{\textbf{Jev-GAP-FB $-$ AgentDropout}}\\
$E_\Delta$ & 0.04925 & 0.03978 & 0.05893 & 1\\
$\Enon$ & 0.05957 & 0.04595 & 0.07336 & 1\\
$G^{\rm own}$ & 0.06174 & 0.04469 & 0.07919 & 1\\
$\Delta T_{\rm gold}$ & 0.19111 & 0.16167 & 0.22278 & 1\\
F1 & 0.17111 & 0.12363 & 0.22194 & 1\\
Peer tokens & 126.895 & 114.392 & 139.642 & 1\\
\addlinespace[3pt]
\multicolumn{5}{@{}l}{\textbf{Jev-GAP-FB $-$ PEAR}}\\
$E_\Delta$ & 0.07815 & 0.06389 & 0.09254 & 1\\
$\Enon$ & 0.09112 & 0.07248 & 0.10978 & 1\\
$G^{\rm own}$ & 0.11237 & 0.09146 & 0.13322 & 1\\
$\Delta T_{\rm gold}$ & 0.21694 & 0.18111 & 0.25361 & 1\\
F1 & 0.17179 & 0.12486 & 0.21898 & 1\\
Peer tokens & -28.135 & -44.8564 & -11.4613 & 0.0002\\
\addlinespace[3pt]
\multicolumn{5}{@{}l}{\textbf{Compact-All $-$ AgentDropout}}\\
$E_\Delta$ & 0.06773 & 0.05313 & 0.08228 & 1\\
$\Enon$ & 0.08997 & 0.07079 & 0.1095 & 1\\
$G^{\rm own}$ & 0.06054 & 0.04167 & 0.07905 & 1\\
$\Delta T_{\rm gold}$ & 0.2675 & 0.23333 & 0.3025 & 1\\
F1 & 0.22399 & 0.1735 & 0.27669 & 1\\
Peer tokens & 134.473 & 126.01 & 142.705 & 1\\
\addlinespace[3pt]
\multicolumn{5}{@{}l}{\textbf{Compact-All $-$ PEAR}}\\
$E_\Delta$ & 0.09662 & 0.07908 & 0.11488 & 1\\
$\Enon$ & 0.12152 & 0.09882 & 0.145 & 1\\
$G^{\rm own}$ & 0.11117 & 0.08832 & 0.1335 & 1\\
$\Delta T_{\rm gold}$ & 0.29333 & 0.25333 & 0.33556 & 1\\
F1 & 0.22467 & 0.17157 & 0.27897 & 1\\
Peer tokens & -20.5567 & -35.1671 & -6.45321 & 0.0023\\
\bottomrule\end{tabular*}\end{table}

\begin{table}[t]
\caption{\textbf{AgentDropout and PEAR contrasts --- 2Wiki.} Paired effects, pointwise 95\% intervals, and bootstrap fractions above zero are transcribed from the frozen report. F1 differences use the 0--1 scale.}
\label{tab:s-topology-paired2}\centering\datatablefont\setlength{\tabcolsep}{1.8pt}
\begin{tabular*}{\columnwidth}{@{\extracolsep{\fill}}lrrrr@{}}\toprule
Outcome & Effect & CI low & CI high & \vhead{Boot.\\$>0$}\\\midrule
\multicolumn{5}{@{}l}{\textbf{Jev-GAP-FB $-$ AgentDropout}}\\
$E_\Delta$ & 0.07673 & 0.0603 & 0.09428 & 1\\
$\Enon$ & 0.05307 & 0.02873 & 0.07844 & 1\\
$G^{\rm own}$ & 0.21093 & 0.18351 & 0.23892 & 1\\
$\Delta T_{\rm gold}$ & 0.14176 & 0.09674 & 0.1887 & 1\\
F1 & 0.01113 & -0.04468 & 0.06925 & 0.6457\\
Peer tokens & -19.658 & -29.0433 & -10.221 & 0\\
\addlinespace[3pt]
\multicolumn{5}{@{}l}{\textbf{Jev-GAP-FB $-$ PEAR}}\\
$E_\Delta$ & 0.09832 & 0.08108 & 0.11601 & 1\\
$\Enon$ & 0.07001 & 0.04456 & 0.09693 & 1\\
$G^{\rm own}$ & 0.27451 & 0.24788 & 0.30085 & 1\\
$\Delta T_{\rm gold}$ & 0.12117 & 0.07039 & 0.17243 & 1\\
F1 & -0.1701 & -0.25128 & -0.08767 & 0\\
Peer tokens & -214.33 & -227.94 & -201.078 & 0\\
\addlinespace[3pt]
\multicolumn{5}{@{}l}{\textbf{Compact-All $-$ AgentDropout}}\\
$E_\Delta$ & 0.08342 & 0.06202 & 0.10536 & 1\\
$\Enon$ & 0.19911 & 0.1693 & 0.22972 & 1\\
$G^{\rm own}$ & -0.05257 & -0.08542 & -0.02044 & 0.0007\\
$\Delta T_{\rm gold}$ & 0.41284 & 0.35248 & 0.4727 & 1\\
F1 & 0.05203 & 0.00094 & 0.10374 & 0.9766\\
Peer tokens & 52.6035 & 47.1581 & 57.9483 & 1\\
\addlinespace[3pt]
\multicolumn{5}{@{}l}{\textbf{Compact-All $-$ PEAR}}\\
$E_\Delta$ & 0.10501 & 0.08202 & 0.12796 & 1\\
$\Enon$ & 0.21605 & 0.1851 & 0.24747 & 1\\
$G^{\rm own}$ & 0.01101 & -0.01834 & 0.04146 & 0.7636\\
$\Delta T_{\rm gold}$ & 0.39224 & 0.32854 & 0.4545 & 1\\
F1 & -0.1292 & -0.21107 & -0.04781 & 0.0006\\
Peer tokens & -142.069 & -152.839 & -131.247 & 0\\
\bottomrule\end{tabular*}\end{table}

\begin{table}[t]
\caption{\textbf{AgentDropout and PEAR contrasts --- Primary combined.} Paired effects, pointwise 95\% intervals, and bootstrap fractions above zero are transcribed from the frozen report. F1 differences use the 0--1 scale.}
\label{tab:s-topology-paired3}\centering\datatablefont\setlength{\tabcolsep}{1.8pt}
\begin{tabular*}{\columnwidth}{@{\extracolsep{\fill}}lrrrr@{}}\toprule
Outcome & Effect & CI low & CI high & \vhead{Boot.\\$>0$}\\\midrule
\multicolumn{5}{@{}l}{\textbf{Jev-GAP-FB $-$ AgentDropout}}\\
$E_\Delta$ & 0.05934 & 0.05048 & 0.06841 & 1\\
$\Enon$ & 0.05718 & 0.04478 & 0.07007 & 1\\
$G^{\rm own}$ & 0.11651 & 0.09736 & 0.13536 & 1\\
$\Delta T_{\rm gold}$ & 0.173 & 0.14645 & 0.19972 & 1\\
F1 & 0.11238 & 0.07424 & 0.15285 & 1\\
Peer tokens & 73.097 & 58.838 & 87.0257 & 1\\
\addlinespace[3pt]
\multicolumn{5}{@{}l}{\textbf{Jev-GAP-FB $-$ PEAR}}\\
$E_\Delta$ & 0.08555 & 0.07484 & 0.0966 & 1\\
$\Enon$ & 0.08337 & 0.06815 & 0.09895 & 1\\
$G^{\rm own}$ & 0.17189 & 0.15201 & 0.19265 & 1\\
$\Delta T_{\rm gold}$ & 0.18179 & 0.15137 & 0.21308 & 1\\
F1 & 0.04629 & -0.00366 & 0.09554 & 0.9651\\
Peer tokens & -96.4852 & -114.6 & -77.9839 & 0\\
\addlinespace[3pt]
\multicolumn{5}{@{}l}{\textbf{Compact-All $-$ AgentDropout}}\\
$E_\Delta$ & 0.07349 & 0.06118 & 0.08555 & 1\\
$\Enon$ & 0.13003 & 0.11214 & 0.14902 & 1\\
$G^{\rm own}$ & 0.01902 & -0.0001 & 0.03812 & 0.9742\\
$\Delta T_{\rm gold}$ & 0.32085 & 0.28797 & 0.35478 & 1\\
F1 & 0.16087 & 0.12223 & 0.20145 & 1\\
Peer tokens & 104.42 & 96.0081 & 112.947 & 1\\
\addlinespace[3pt]
\multicolumn{5}{@{}l}{\textbf{Compact-All $-$ PEAR}}\\
$E_\Delta$ & 0.0997 & 0.08565 & 0.11376 & 1\\
$\Enon$ & 0.15622 & 0.13658 & 0.17705 & 1\\
$G^{\rm own}$ & 0.0744 & 0.05504 & 0.0942 & 1\\
$\Delta T_{\rm gold}$ & 0.32964 & 0.29518 & 0.36498 & 1\\
F1 & 0.09477 & 0.03945 & 0.14745 & 0.9997\\
Peer tokens & -65.1625 & -78.8654 & -51.368 & 0\\
\bottomrule\end{tabular*}\end{table}

Tables~\ref{tab:s-topology-paired1}, \ref{tab:s-topology-paired2}, and~\ref{tab:s-topology-paired3} retain all 72 paired effects from the topology follow-up. Jev-GAP-Fallback has positive non-owned differences over AgentDropout and PEAR in both tasks. Compact-All also has positive non-owned differences over both methods in both tasks. The gold-recovery differences follow the same direction, with positive pointwise intervals.

Answer outcomes are task-dependent. On MuSiQue, fallback improves F1 over AgentDropout by 0.17111 and over PEAR by 0.17179, with positive intervals. On 2Wiki its difference from AgentDropout is 0.01113 with interval [$-0.04468$,0.06925], while its difference from PEAR is $-0.17010$ with interval [$-0.25128$,$-0.08767$]. Compact-All also trails PEAR on 2Wiki F1 by 0.12920, with a negative interval, while exceeding it on non-owned gain by 0.21605.

All displayed intervals are the original pointwise paired bootstrap intervals. The fraction of resampled means above zero is a resampling diagnostic. The package includes aggregate tables and source reports, with no synthetic per-question records or reconstructed tests. The additional baselines add 126 reported paired effects to the prior supplementary analyses.

\FloatBarrier
\section{Capability and Family Heterogeneity}
\label{app:heterogeneity}
\label{sec:s-heterogeneity}
\subsection{Design and Tier Notation}
The follow-up examines six homogeneous controls and eight mixed compositions. The six same-tier mixtures are $W_1W_1W_2$, $W_1W_2W_2$, $M_1M_1M_2$, $M_1M_2M_2$, $S_1S_1S_2$, and $S_1S_2S_2$. The two capability-spread mixtures are $W_1M_1S_1$ and $W_2M_2S_2$. The execution plan assigns capability tiers using frozen MuSiQue self-refinement F1 and rotates model-to-evidence positions across questions.

\noindent\textbf{Tier identifiers.}
\tierlegend. The six selected models are grouped by their frozen MuSiQue self-refinement F1 in the main homogeneous experiment. The assignments recorded in the execution plan are $W_1$ = Mistral Small 3.2 24B, $W_2$ = Claude Haiku 4.5, $M_1$ = DeepSeek V3.2, $M_2$ = GPT-5.6 Terra, $S_1$ = Qwen3.5-35B-A3B, and $S_2$ = Gemini 3.1 Flash-Lite. W, M, and S denote relative capability in this selected study population. Same-colour circles distinguish models within a tier, and a three-colour team represents a W--M--S mixture. The assignments are frozen before the composition comparison and retained on 2Wiki.

The experiments use three seeds, three agents, and two communication rounds on MuSiQue and 2Wiki. Round-0 and self-refinement outputs are reused where available. New MAS outputs enforce complete prior-round dependencies. Each reported effect is the mixed-team value minus the homogeneous-control value. Seeds are averaged within question, and paired question-cluster intervals use 10,000 resamples. The intervals are pointwise.

\subsection{Source-Trace Audit}
Reused Mistral Round-0, self-refinement, and homogeneous outputs were served through DeepInfra fp8, while new Mistral MAS updates use Parasail bf16. Comparisons involving this model therefore combine a composition change with a serving-backend change. The source report records unchanged providers for the medium- and strong-tier same-tier mixtures and for $W_2M_2S_2$.

The inherited frozen traces contain context mismatches on 69 MuSiQue and 29 2Wiki questions. A paired sensitivity analysis on the context-consistent question subset shared by all compositions is pending. The supplied archive contains the frozen-report contrasts but lacks the completed common-clean-subset analysis and the executed question-level audit mask. The composition and directional estimates retain the exploratory status stated in the main paper.

\subsection{Complete Same-Tier Comparisons}
Tables~\ref{tab:s-het-effects1}, \ref{tab:s-het-effects2}, \ref{tab:s-het-effects3}, and~\ref{tab:s-het-effects4} report all 96 effects for the six same-tier mixtures against both homogeneous endpoints, across the two benchmarks and four outcomes. These tables supply the full paired intervals underlying the main-text heterogeneous comparison and add the independent gold-recovery differences. Both homogeneous endpoints and all reported effect signs are retained.

\begin{table*}[t]
\caption{\textbf{Same-tier paired contrasts: $\delta E_\Delta$.} Mixed team minus homogeneous component control, with pointwise 95\% intervals for every reported endpoint comparison.}
\label{tab:s-het-effects1}
\centering\datatablefont\setlength{\tabcolsep}{3pt}\renewcommand{\arraystretch}{1.1}
\begin{tabular*}{\textwidth}{@{\extracolsep{\fill}}llrrrrrr@{}}
\toprule
 & & \multicolumn{3}{c}{MuSiQue} & \multicolumn{3}{c}{2Wiki}\\
\cmidrule(lr){3-5}\cmidrule(l){6-8}
Mixed team & Homogeneous control & Effect & CI low & CI high & Effect & CI low & CI high\\
\midrule
\mainband{8}{topologyband}{Weak-tier mixtures (backend change)}
\tierdisc{W}{1}\,\tierdisc{W}{1}\,\tierdisc{W}{2} & \tierdisc{W}{1}\,\tierdisc{W}{1}\,\tierdisc{W}{1} & +0.0120 & 0.0022 & 0.0220 & \ensuremath{-}0.0461 & \ensuremath{-}0.0633 & \ensuremath{-}0.0289\\
\tierdisc{W}{1}\,\tierdisc{W}{1}\,\tierdisc{W}{2} & \tierdisc{W}{2}\,\tierdisc{W}{2}\,\tierdisc{W}{2} & +0.0254 & 0.0141 & 0.0366 & +0.0650 & 0.0465 & 0.0832\\
\tierdisc{W}{1}\,\tierdisc{W}{2}\,\tierdisc{W}{2} & \tierdisc{W}{1}\,\tierdisc{W}{1}\,\tierdisc{W}{1} & \ensuremath{-}0.0062 & \ensuremath{-}0.0194 & 0.0068 & \ensuremath{-}0.0864 & \ensuremath{-}0.1098 & \ensuremath{-}0.0618\\
\tierdisc{W}{1}\,\tierdisc{W}{2}\,\tierdisc{W}{2} & \tierdisc{W}{2}\,\tierdisc{W}{2}\,\tierdisc{W}{2} & +0.0072 & \ensuremath{-}0.0005 & 0.0149 & +0.0246 & 0.0134 & 0.0355\\
\mainband{8}{factband}{Medium-tier mixtures}
\tierdisc{M}{1}\,\tierdisc{M}{1}\,\tierdisc{M}{2} & \tierdisc{M}{1}\,\tierdisc{M}{1}\,\tierdisc{M}{1} & \ensuremath{-}0.0038 & \ensuremath{-}0.0188 & 0.0110 & \ensuremath{-}0.0333 & \ensuremath{-}0.0471 & \ensuremath{-}0.0206\\
\tierdisc{M}{1}\,\tierdisc{M}{1}\,\tierdisc{M}{2} & \tierdisc{M}{2}\,\tierdisc{M}{2}\,\tierdisc{M}{2} & \ensuremath{-}0.0143 & \ensuremath{-}0.0287 & \ensuremath{-}0.0005 & +0.0623 & 0.0485 & 0.0765\\
\tierdisc{M}{1}\,\tierdisc{M}{2}\,\tierdisc{M}{2} & \tierdisc{M}{1}\,\tierdisc{M}{1}\,\tierdisc{M}{1} & +0.0034 & \ensuremath{-}0.0155 & 0.0216 & \ensuremath{-}0.0695 & \ensuremath{-}0.0881 & \ensuremath{-}0.0509\\
\tierdisc{M}{1}\,\tierdisc{M}{2}\,\tierdisc{M}{2} & \tierdisc{M}{2}\,\tierdisc{M}{2}\,\tierdisc{M}{2} & \ensuremath{-}0.0072 & \ensuremath{-}0.0184 & 0.0040 & +0.0261 & 0.0184 & 0.0337\\
\mainband{8}{topologyband}{Strong-tier mixtures}
\tierdisc{S}{1}\,\tierdisc{S}{1}\,\tierdisc{S}{2} & \tierdisc{S}{1}\,\tierdisc{S}{1}\,\tierdisc{S}{1} & \ensuremath{-}0.0031 & \ensuremath{-}0.0142 & 0.0085 & \ensuremath{-}0.1162 & \ensuremath{-}0.1329 & \ensuremath{-}0.0993\\
\tierdisc{S}{1}\,\tierdisc{S}{1}\,\tierdisc{S}{2} & \tierdisc{S}{2}\,\tierdisc{S}{2}\,\tierdisc{S}{2} & +0.0174 & 0.0026 & 0.0320 & +0.1951 & 0.1825 & 0.2084\\
\tierdisc{S}{1}\,\tierdisc{S}{2}\,\tierdisc{S}{2} & \tierdisc{S}{1}\,\tierdisc{S}{1}\,\tierdisc{S}{1} & \ensuremath{-}0.0159 & \ensuremath{-}0.0308 & \ensuremath{-}0.0009 & \ensuremath{-}0.2188 & \ensuremath{-}0.2366 & \ensuremath{-}0.2011\\
\tierdisc{S}{1}\,\tierdisc{S}{2}\,\tierdisc{S}{2} & \tierdisc{S}{2}\,\tierdisc{S}{2}\,\tierdisc{S}{2} & +0.0046 & \ensuremath{-}0.0080 & 0.0171 & +0.0925 & 0.0818 & 0.1033\\
\bottomrule\end{tabular*}
\end{table*}

\begin{table*}[t]
\caption{\textbf{Same-tier paired contrasts: $\delta\Enon$.} Mixed team minus homogeneous component control, with pointwise 95\% intervals for every reported endpoint comparison.}
\label{tab:s-het-effects2}
\centering\datatablefont\setlength{\tabcolsep}{3pt}\renewcommand{\arraystretch}{1.1}
\begin{tabular*}{\textwidth}{@{\extracolsep{\fill}}llrrrrrr@{}}
\toprule
 & & \multicolumn{3}{c}{MuSiQue} & \multicolumn{3}{c}{2Wiki}\\
\cmidrule(lr){3-5}\cmidrule(l){6-8}
Mixed team & Homogeneous control & Effect & CI low & CI high & Effect & CI low & CI high\\
\midrule
\mainband{8}{topologyband}{Weak-tier mixtures (backend change)}
\tierdisc{W}{1}\,\tierdisc{W}{1}\,\tierdisc{W}{2} & \tierdisc{W}{1}\,\tierdisc{W}{1}\,\tierdisc{W}{1} & +0.0070 & \ensuremath{-}0.0079 & 0.0219 & \ensuremath{-}0.0479 & \ensuremath{-}0.0782 & \ensuremath{-}0.0175\\
\tierdisc{W}{1}\,\tierdisc{W}{1}\,\tierdisc{W}{2} & \tierdisc{W}{2}\,\tierdisc{W}{2}\,\tierdisc{W}{2} & +0.0539 & 0.0360 & 0.0714 & +0.0428 & 0.0131 & 0.0715\\
\tierdisc{W}{1}\,\tierdisc{W}{2}\,\tierdisc{W}{2} & \tierdisc{W}{1}\,\tierdisc{W}{1}\,\tierdisc{W}{1} & \ensuremath{-}0.0244 & \ensuremath{-}0.0447 & \ensuremath{-}0.0040 & \ensuremath{-}0.0774 & \ensuremath{-}0.1190 & \ensuremath{-}0.0339\\
\tierdisc{W}{1}\,\tierdisc{W}{2}\,\tierdisc{W}{2} & \tierdisc{W}{2}\,\tierdisc{W}{2}\,\tierdisc{W}{2} & +0.0225 & 0.0105 & 0.0344 & +0.0132 & \ensuremath{-}0.0029 & 0.0286\\
\mainband{8}{factband}{Medium-tier mixtures}
\tierdisc{M}{1}\,\tierdisc{M}{1}\,\tierdisc{M}{2} & \tierdisc{M}{1}\,\tierdisc{M}{1}\,\tierdisc{M}{1} & \ensuremath{-}0.0092 & \ensuremath{-}0.0290 & 0.0107 & \ensuremath{-}0.0389 & \ensuremath{-}0.0604 & \ensuremath{-}0.0190\\
\tierdisc{M}{1}\,\tierdisc{M}{1}\,\tierdisc{M}{2} & \tierdisc{M}{2}\,\tierdisc{M}{2}\,\tierdisc{M}{2} & \ensuremath{-}0.0134 & \ensuremath{-}0.0362 & 0.0081 & +0.0780 & 0.0536 & 0.1024\\
\tierdisc{M}{1}\,\tierdisc{M}{2}\,\tierdisc{M}{2} & \tierdisc{M}{1}\,\tierdisc{M}{1}\,\tierdisc{M}{1} & \ensuremath{-}0.0063 & \ensuremath{-}0.0317 & 0.0184 & \ensuremath{-}0.0862 & \ensuremath{-}0.1167 & \ensuremath{-}0.0561\\
\tierdisc{M}{1}\,\tierdisc{M}{2}\,\tierdisc{M}{2} & \tierdisc{M}{2}\,\tierdisc{M}{2}\,\tierdisc{M}{2} & \ensuremath{-}0.0105 & \ensuremath{-}0.0285 & 0.0072 & +0.0307 & 0.0175 & 0.0443\\
\mainband{8}{topologyband}{Strong-tier mixtures}
\tierdisc{S}{1}\,\tierdisc{S}{1}\,\tierdisc{S}{2} & \tierdisc{S}{1}\,\tierdisc{S}{1}\,\tierdisc{S}{1} & \ensuremath{-}0.0063 & \ensuremath{-}0.0224 & 0.0099 & \ensuremath{-}0.1682 & \ensuremath{-}0.1998 & \ensuremath{-}0.1366\\
\tierdisc{S}{1}\,\tierdisc{S}{1}\,\tierdisc{S}{2} & \tierdisc{S}{2}\,\tierdisc{S}{2}\,\tierdisc{S}{2} & +0.0289 & 0.0084 & 0.0487 & +0.2477 & 0.2260 & 0.2694\\
\tierdisc{S}{1}\,\tierdisc{S}{2}\,\tierdisc{S}{2} & \tierdisc{S}{1}\,\tierdisc{S}{1}\,\tierdisc{S}{1} & \ensuremath{-}0.0289 & \ensuremath{-}0.0495 & \ensuremath{-}0.0088 & \ensuremath{-}0.3011 & \ensuremath{-}0.3335 & \ensuremath{-}0.2686\\
\tierdisc{S}{1}\,\tierdisc{S}{2}\,\tierdisc{S}{2} & \tierdisc{S}{2}\,\tierdisc{S}{2}\,\tierdisc{S}{2} & +0.0063 & \ensuremath{-}0.0111 & 0.0234 & +0.1147 & 0.0974 & 0.1327\\
\bottomrule\end{tabular*}
\end{table*}

\begin{table*}[t]
\caption{\textbf{Same-tier paired contrasts: $\delta(\Delta T_{\rm gold})$.} Mixed team minus homogeneous component control, with pointwise 95\% intervals for every reported endpoint comparison.}
\label{tab:s-het-effects3}
\centering\datatablefont\setlength{\tabcolsep}{3pt}\renewcommand{\arraystretch}{1.1}
\begin{tabular*}{\textwidth}{@{\extracolsep{\fill}}llrrrrrr@{}}
\toprule
 & & \multicolumn{3}{c}{MuSiQue} & \multicolumn{3}{c}{2Wiki}\\
\cmidrule(lr){3-5}\cmidrule(l){6-8}
Mixed team & Homogeneous control & Effect & CI low & CI high & Effect & CI low & CI high\\
\midrule
\mainband{8}{topologyband}{Weak-tier mixtures (backend change)}
\tierdisc{W}{1}\,\tierdisc{W}{1}\,\tierdisc{W}{2} & \tierdisc{W}{1}\,\tierdisc{W}{1}\,\tierdisc{W}{1} & \ensuremath{-}0.0144 & \ensuremath{-}0.0417 & 0.0144 & +0.0527 & 0.0326 & 0.0728\\
\tierdisc{W}{1}\,\tierdisc{W}{1}\,\tierdisc{W}{2} & \tierdisc{W}{2}\,\tierdisc{W}{2}\,\tierdisc{W}{2} & +0.0572 & 0.0206 & 0.0928 & \ensuremath{-}0.0460 & \ensuremath{-}0.0738 & \ensuremath{-}0.0182\\
\tierdisc{W}{1}\,\tierdisc{W}{2}\,\tierdisc{W}{2} & \tierdisc{W}{1}\,\tierdisc{W}{1}\,\tierdisc{W}{1} & \ensuremath{-}0.0417 & \ensuremath{-}0.0789 & \ensuremath{-}0.0039 & +0.0718 & 0.0431 & 0.1006\\
\tierdisc{W}{1}\,\tierdisc{W}{2}\,\tierdisc{W}{2} & \tierdisc{W}{2}\,\tierdisc{W}{2}\,\tierdisc{W}{2} & +0.0300 & 0.0033 & 0.0561 & \ensuremath{-}0.0268 & \ensuremath{-}0.0421 & \ensuremath{-}0.0125\\
\mainband{8}{factband}{Medium-tier mixtures}
\tierdisc{M}{1}\,\tierdisc{M}{1}\,\tierdisc{M}{2} & \tierdisc{M}{1}\,\tierdisc{M}{1}\,\tierdisc{M}{1} & \ensuremath{-}0.0444 & \ensuremath{-}0.0750 & \ensuremath{-}0.0144 & \ensuremath{-}0.0048 & \ensuremath{-}0.0134 & 0.0038\\
\tierdisc{M}{1}\,\tierdisc{M}{1}\,\tierdisc{M}{2} & \tierdisc{M}{2}\,\tierdisc{M}{2}\,\tierdisc{M}{2} & +0.0656 & 0.0300 & 0.1011 & +0.0086 & \ensuremath{-}0.0077 & 0.0259\\
\tierdisc{M}{1}\,\tierdisc{M}{2}\,\tierdisc{M}{2} & \tierdisc{M}{1}\,\tierdisc{M}{1}\,\tierdisc{M}{1} & \ensuremath{-}0.0694 & \ensuremath{-}0.1094 & \ensuremath{-}0.0300 & \ensuremath{-}0.0086 & \ensuremath{-}0.0230 & 0.0048\\
\tierdisc{M}{1}\,\tierdisc{M}{2}\,\tierdisc{M}{2} & \tierdisc{M}{2}\,\tierdisc{M}{2}\,\tierdisc{M}{2} & +0.0406 & 0.0150 & 0.0667 & +0.0048 & \ensuremath{-}0.0077 & 0.0172\\
\mainband{8}{topologyband}{Strong-tier mixtures}
\tierdisc{S}{1}\,\tierdisc{S}{1}\,\tierdisc{S}{2} & \tierdisc{S}{1}\,\tierdisc{S}{1}\,\tierdisc{S}{1} & \ensuremath{-}0.0294 & \ensuremath{-}0.0561 & \ensuremath{-}0.0033 & \ensuremath{-}0.0843 & \ensuremath{-}0.1121 & \ensuremath{-}0.0556\\
\tierdisc{S}{1}\,\tierdisc{S}{1}\,\tierdisc{S}{2} & \tierdisc{S}{2}\,\tierdisc{S}{2}\,\tierdisc{S}{2} & +0.0906 & 0.0622 & 0.1183 & +0.1772 & 0.1322 & 0.2251\\
\tierdisc{S}{1}\,\tierdisc{S}{2}\,\tierdisc{S}{2} & \tierdisc{S}{1}\,\tierdisc{S}{1}\,\tierdisc{S}{1} & \ensuremath{-}0.0839 & \ensuremath{-}0.1156 & \ensuremath{-}0.0517 & \ensuremath{-}0.1628 & \ensuremath{-}0.2059 & \ensuremath{-}0.1188\\
\tierdisc{S}{1}\,\tierdisc{S}{2}\,\tierdisc{S}{2} & \tierdisc{S}{2}\,\tierdisc{S}{2}\,\tierdisc{S}{2} & +0.0361 & 0.0133 & 0.0600 & +0.0987 & 0.0680 & 0.1293\\
\bottomrule\end{tabular*}
\end{table*}

\begin{table*}[t]
\caption{\textbf{Same-tier paired contrasts: $\delta(\Delta\mathrm{F1})$.} Mixed team minus homogeneous component control, with pointwise 95\% intervals for every reported endpoint comparison. F1 differences use the source 0--1 scale.}
\label{tab:s-het-effects4}
\centering\datatablefont\setlength{\tabcolsep}{3pt}\renewcommand{\arraystretch}{1.1}
\begin{tabular*}{\textwidth}{@{\extracolsep{\fill}}llrrrrrr@{}}
\toprule
 & & \multicolumn{3}{c}{MuSiQue} & \multicolumn{3}{c}{2Wiki}\\
\cmidrule(lr){3-5}\cmidrule(l){6-8}
Mixed team & Homogeneous control & Effect & CI low & CI high & Effect & CI low & CI high\\
\midrule
\mainband{8}{topologyband}{Weak-tier mixtures (backend change)}
\tierdisc{W}{1}\,\tierdisc{W}{1}\,\tierdisc{W}{2} & \tierdisc{W}{1}\,\tierdisc{W}{1}\,\tierdisc{W}{1} & +0.0524 & 0.0002 & 0.1081 & +0.2917 & 0.1853 & 0.4022\\
\tierdisc{W}{1}\,\tierdisc{W}{1}\,\tierdisc{W}{2} & \tierdisc{W}{2}\,\tierdisc{W}{2}\,\tierdisc{W}{2} & +0.1095 & 0.0426 & 0.1780 & \ensuremath{-}0.3154 & \ensuremath{-}0.4208 & \ensuremath{-}0.2183\\
\tierdisc{W}{1}\,\tierdisc{W}{2}\,\tierdisc{W}{2} & \tierdisc{W}{1}\,\tierdisc{W}{1}\,\tierdisc{W}{1} & +0.0334 & \ensuremath{-}0.0240 & 0.0954 & +0.4921 & 0.3785 & 0.6071\\
\tierdisc{W}{1}\,\tierdisc{W}{2}\,\tierdisc{W}{2} & \tierdisc{W}{2}\,\tierdisc{W}{2}\,\tierdisc{W}{2} & +0.0906 & 0.0414 & 0.1408 & \ensuremath{-}0.1149 & \ensuremath{-}0.1667 & \ensuremath{-}0.0690\\
\mainband{8}{factband}{Medium-tier mixtures}
\tierdisc{M}{1}\,\tierdisc{M}{1}\,\tierdisc{M}{2} & \tierdisc{M}{1}\,\tierdisc{M}{1}\,\tierdisc{M}{1} & \ensuremath{-}0.0602 & \ensuremath{-}0.1233 & \ensuremath{-}0.0006 & +0.0000 & 0.0000 & 0.0000\\
\tierdisc{M}{1}\,\tierdisc{M}{1}\,\tierdisc{M}{2} & \tierdisc{M}{2}\,\tierdisc{M}{2}\,\tierdisc{M}{2} & +0.0383 & \ensuremath{-}0.0295 & 0.1050 & +0.0000 & 0.0000 & 0.0000\\
\tierdisc{M}{1}\,\tierdisc{M}{2}\,\tierdisc{M}{2} & \tierdisc{M}{1}\,\tierdisc{M}{1}\,\tierdisc{M}{1} & \ensuremath{-}0.0503 & \ensuremath{-}0.1351 & 0.0305 & +0.0000 & 0.0000 & 0.0000\\
\tierdisc{M}{1}\,\tierdisc{M}{2}\,\tierdisc{M}{2} & \tierdisc{M}{2}\,\tierdisc{M}{2}\,\tierdisc{M}{2} & +0.0482 & \ensuremath{-}0.0102 & 0.1090 & +0.0000 & 0.0000 & 0.0000\\
\mainband{8}{topologyband}{Strong-tier mixtures}
\tierdisc{S}{1}\,\tierdisc{S}{1}\,\tierdisc{S}{2} & \tierdisc{S}{1}\,\tierdisc{S}{1}\,\tierdisc{S}{1} & \ensuremath{-}0.0047 & \ensuremath{-}0.0575 & 0.0488 & \ensuremath{-}0.0644 & \ensuremath{-}0.1322 & 0.0046\\
\tierdisc{S}{1}\,\tierdisc{S}{1}\,\tierdisc{S}{2} & \tierdisc{S}{2}\,\tierdisc{S}{2}\,\tierdisc{S}{2} & +0.0785 & 0.0144 & 0.1422 & +0.1598 & 0.0402 & 0.2793\\
\tierdisc{S}{1}\,\tierdisc{S}{2}\,\tierdisc{S}{2} & \tierdisc{S}{1}\,\tierdisc{S}{1}\,\tierdisc{S}{1} & \ensuremath{-}0.0766 & \ensuremath{-}0.1424 & \ensuremath{-}0.0120 & \ensuremath{-}0.1609 & \ensuremath{-}0.2471 & \ensuremath{-}0.0747\\
\tierdisc{S}{1}\,\tierdisc{S}{2}\,\tierdisc{S}{2} & \tierdisc{S}{2}\,\tierdisc{S}{2}\,\tierdisc{S}{2} & +0.0065 & \ensuremath{-}0.0506 & 0.0626 & +0.0632 & \ensuremath{-}0.0460 & 0.1782\\
\bottomrule\end{tabular*}
\end{table*}

\noindent\textbf{Reading the component contrasts.}
Each row fixes a mixed composition and one component homogeneous control. Its effect describes that particular endpoint comparison. The two rows for a mixture expose how its gain changes with the selected control. The F1 contrasts in this appendix retain the source 0--1 scale. The main-text table expresses the same F1 differences in percentage points.

\subsection{Directional Recovery and Reciprocal Contrasts}
\begin{table*}[t]
\caption{\textbf{Directional endpoint statistics.} Counts and rates are copied from the frozen report. Raw recovery is new recovery among initially absent, non-owned facts. self-refinement-adjusted gold gain uses the report\textquotesingle s separate normalization. It is not the raw recovery rate.}
\label{tab:s-het-directions}
\centering\datatablefont\setlength{\tabcolsep}{2.8pt}\renewcommand{\arraystretch}{1.1}
\begin{tabular*}{\textwidth}{@{\extracolsep{\fill}}lrrrrrrr@{}}
\toprule
Owner $\to$ receiver & Cells & Eligible & Recovered & \vhead{Raw\\recovery} & \vhead{Final gold\\recovery} & \vhead{self-refinement-adjusted\\gold gain} & \vhead{self-refinement-adjusted\\coverage gain}\\
\midrule
\mainband{8}{topologyband}{MuSiQue. Two W--M--S teams, three seeds}
\tierdisc{M}{}\,$\to$\,\tierdisc{S}{} & 600 & 251 & 93 & 0.3705 & 0.6767 & 0.0817 & 0.0330\\
\tierdisc{M}{}\,$\to$\,\tierdisc{W}{} & 600 & 305 & 165 & 0.5410 & 0.7417 & 0.2533 & 0.0765\\
\tierdisc{S}{}\,$\to$\,\tierdisc{M}{} & 600 & 315 & 118 & 0.3746 & 0.6500 & 0.2100 & 0.0863\\
\tierdisc{S}{}\,$\to$\,\tierdisc{W}{} & 600 & 265 & 141 & 0.5321 & 0.7800 & 0.2233 & 0.0707\\
\tierdisc{W}{}\,$\to$\,\tierdisc{M}{} & 600 & 328 & 94 & 0.2866 & 0.5900 & 0.1717 & 0.0864\\
\tierdisc{W}{}\,$\to$\,\tierdisc{S}{} & 600 & 288 & 77 & 0.2674 & 0.6100 & 0.0850 & 0.0051\\
\mainband{8}{factband}{2Wiki. Two W--M--S teams, three seeds}
\tierdisc{M}{}\,$\to$\,\tierdisc{S}{} & 348 & 225 & 220 & 0.9778 & 0.9856 & 0.6264 & 0.3905\\
\tierdisc{M}{}\,$\to$\,\tierdisc{W}{} & 348 & 274 & 270 & 0.9854 & 0.9885 & 0.7931 & 0.4648\\
\tierdisc{S}{}\,$\to$\,\tierdisc{M}{} & 348 & 310 & 308 & 0.9935 & 0.9943 & 0.8879 & 0.4511\\
\tierdisc{S}{}\,$\to$\,\tierdisc{W}{} & 348 & 255 & 254 & 0.9961 & 0.9971 & 0.7443 & 0.4752\\
\tierdisc{W}{}\,$\to$\,\tierdisc{M}{} & 348 & 293 & 283 & 0.9659 & 0.9713 & 0.8190 & 0.4526\\
\tierdisc{W}{}\,$\to$\,\tierdisc{S}{} & 348 & 210 & 207 & 0.9857 & 0.9885 & 0.6034 & 0.3703\\
\bottomrule\end{tabular*}
\end{table*}

A direction $a\!\to\!b$ links the tier initially owning a fact to the receiving tier. The raw transfer rate divides the number of subsequently recovered facts by the number of initially absent non-owned facts. Table~\ref{tab:s-het-directions} reports the recovered and eligible counts alongside the corresponding raw rates. The self-refinement-adjusted gold-gain statistic uses its own normalization, which is retained in a separate column.

Ownership identifies the original evidence allocation. The symmetric MAS protocol permits indirect transmission. Directional outcomes also depend on receiver ability and initial coverage. The reciprocal contrasts compare the same tiers in reversed ownership and receiver roles, with their self-refinement differences subtracted.

\begin{table*}[t]
\caption{\textbf{Original reciprocal-direction contrasts.} Signs and intervals follow the source report. The main-text table reverses these contrasts for readability using $(d,l,u)\mapsto(-d,-u,-l)$. $B_+$ is the bootstrap fraction above zero.}
\label{tab:s-het-reciprocal}
\centering\datatablefont\setlength{\tabcolsep}{3pt}\renewcommand{\arraystretch}{1.1}
\begin{tabular*}{\textwidth}{@{\extracolsep{\fill}}llrrrr@{}}
\toprule
Direction contrast & Outcome & Effect & CI low & CI high & $B_+$\\
\midrule
\mainband{6}{topologyband}{MuSiQue}
\tierdisc{W}{}\,$\to$\,\tierdisc{M}{}\,$-$\,\tierdisc{M}{}\,$\to$\,\tierdisc{W}{} & self-refinement-adjusted gold & \ensuremath{-}0.0817 & \ensuremath{-}0.1467 & \ensuremath{-}0.0167 & 0.0059\\
\tierdisc{W}{}\,$\to$\,\tierdisc{M}{}\,$-$\,\tierdisc{M}{}\,$\to$\,\tierdisc{W}{} & Non-owned coverage & \ensuremath{-}0.0329 & \ensuremath{-}0.0611 & \ensuremath{-}0.0048 & 0.0105\\
\tierdisc{W}{}\,$\to$\,\tierdisc{S}{}\,$-$\,\tierdisc{S}{}\,$\to$\,\tierdisc{W}{} & self-refinement-adjusted gold & \ensuremath{-}0.1383 & \ensuremath{-}0.2083 & \ensuremath{-}0.0700 & 0.0001\\
\tierdisc{W}{}\,$\to$\,\tierdisc{S}{}\,$-$\,\tierdisc{S}{}\,$\to$\,\tierdisc{W}{} & Non-owned coverage & \ensuremath{-}0.0707 & \ensuremath{-}0.1019 & \ensuremath{-}0.0396 & 0.0001\\
\tierdisc{M}{}\,$\to$\,\tierdisc{S}{}\,$-$\,\tierdisc{S}{}\,$\to$\,\tierdisc{M}{} & self-refinement-adjusted gold & \ensuremath{-}0.1283 & \ensuremath{-}0.1934 & \ensuremath{-}0.0633 & 0.0001\\
\tierdisc{M}{}\,$\to$\,\tierdisc{S}{}\,$-$\,\tierdisc{S}{}\,$\to$\,\tierdisc{M}{} & Non-owned coverage & \ensuremath{-}0.0183 & \ensuremath{-}0.0477 & 0.0122 & 0.1205\\
\mainband{6}{factband}{2Wiki}
\tierdisc{W}{}\,$\to$\,\tierdisc{M}{}\,$-$\,\tierdisc{M}{}\,$\to$\,\tierdisc{W}{} & self-refinement-adjusted gold & +0.0259 & \ensuremath{-}0.0086 & 0.0661 & 0.9257\\
\tierdisc{W}{}\,$\to$\,\tierdisc{M}{}\,$-$\,\tierdisc{M}{}\,$\to$\,\tierdisc{W}{} & Non-owned coverage & \ensuremath{-}0.0156 & \ensuremath{-}0.0617 & 0.0343 & 0.2512\\
\tierdisc{W}{}\,$\to$\,\tierdisc{S}{}\,$-$\,\tierdisc{S}{}\,$\to$\,\tierdisc{W}{} & self-refinement-adjusted gold & \ensuremath{-}0.1408 & \ensuremath{-}0.2011 & \ensuremath{-}0.0776 & 0.0000\\
\tierdisc{W}{}\,$\to$\,\tierdisc{S}{}\,$-$\,\tierdisc{S}{}\,$\to$\,\tierdisc{W}{} & Non-owned coverage & \ensuremath{-}0.0847 & \ensuremath{-}0.1329 & \ensuremath{-}0.0367 & 0.0002\\
\tierdisc{M}{}\,$\to$\,\tierdisc{S}{}\,$-$\,\tierdisc{S}{}\,$\to$\,\tierdisc{M}{} & self-refinement-adjusted gold & \ensuremath{-}0.2615 & \ensuremath{-}0.3218 & \ensuremath{-}0.2040 & 0.0000\\
\tierdisc{M}{}\,$\to$\,\tierdisc{S}{}\,$-$\,\tierdisc{S}{}\,$\to$\,\tierdisc{M}{} & Non-owned coverage & \ensuremath{-}0.0476 & \ensuremath{-}0.0847 & \ensuremath{-}0.0098 & 0.0060\\
\bottomrule\end{tabular*}
\end{table*}

Table~\ref{tab:s-het-reciprocal} preserves the source report's low-to-high minus high-to-low signs and paired intervals. The main-text table reverses these contrasts so that a positive value indicates a higher-tier-owner to lower-tier-receiver advantage. The exact transformation is
\[
(d,[l,u])\longmapsto(-d,[-u,-l]).
\]
The two presentations therefore describe the same estimated contrasts. Raw recovery rates and adjusted effects retain their distinct denominators and interpretations. The source-trace audit applies to both the same-tier and directional results.

\noindent\textbf{Source records.}
The original report, same-tier paired effects, directional recovery counts, and reciprocal contrasts are retained under \texttt{data/heterogeneity\_v15/}. Displayed tables use the recorded estimates and intervals without reconstructing unavailable per-composition absolute means or seed standard deviations.
\FloatBarrier

\raggedbottom
\section{Manuscript Artifacts and Research Scope}
\label{app:artifacts}
\noindent\textbf{Included artifacts.}
The project contains the main manuscript with its appendices, the AAMAS class and bibliography style, editable figure sources, aggregate CSV tables, source reports, and build scripts. The CSV files preserve the supplied experimental values. Displayed confidence intervals come from the corresponding question-level analyses recorded in those reports. Method illustrations are labelled as examples.

\noindent\textbf{Experimental records.}
The manuscript package provides aggregate results. Raw generation traces, verbatim prompts, exact reference files, provider metadata, and per-question statistical scripts belong to the separate experiment archive. The reporting protocol specifies logical-slot deduplication, shared initial states for architecture comparisons, and seed-averaged question-bootstrap inference.

\noindent\textbf{AI assistance.}
ChatGPT assisted with evaluation-design discussions, experimental-analysis instructions, manuscript preparation, and diagram and aggregate-plot code. The experimental values originate from the supplied frozen reports.

\noindent\textbf{Research scope.}
The experiments evaluate expressed reasoning in factual question answering using public benchmarks. Reference alignment is defined by the annotated intermediate information and final answers. Interpretation depends on reference quality, evidence allocation, and the specified matching procedure.

\section{Prompt Templates}
\label{app:prompts}
\label{sec:s-prompts}

\noindent\textbf{Scope and provenance.}
These protocol-level templates normalize the interfaces in \apprefs{protocol}{gap-protocol} and \appref{jev-protocol}. Wording and placeholder names are reconstructed from the recorded protocol. Verbatim API requests are absent from the supplied project and remain necessary for byte-exact replication. Braced placeholders are populated from runtime state. References, gold labels, answer aliases, and evaluation scores are excluded from generation and routing inputs.

\subsection{Matched Reasoning Trajectories}
The cross-family and composition experiments emit exactly three reasoning steps, a conclusion, and an answer. Self refinement uses the question, local source, and own previous response. MAS additionally receives prior-round peer responses. Updates are synchronous.

\noindent\begin{minipage}{\columnwidth}
\noindent\textbf{P1. Independent initial response.}
\begin{lstlisting}[style=ceaprompt]
Question:
{question}

Your assigned evidence:
{local_evidence}

Reason about the question using your assigned evidence. Return a JSON object with exactly three reasoning steps, a conclusion, and a short final answer:
{"steps": ["...", "...", "..."],
 "conclusion": "...",
 "answer": "..."}
\end{lstlisting}
\end{minipage}\par\smallskip
For corrected 2Wiki, the recorded protocol additionally instructs agents to answer \texttt{unknown} when their available evidence is insufficient. MuSiQue supplies the assigned source paragraph. Corrected 2Wiki supplies the rendered evidence fact and adds the requirement to state that local fact explicitly. The shown output keys are those documented for the reasoning trajectories.

\noindent\begin{minipage}{\columnwidth}
\noindent\textbf{P2. Individual self-refinement update.}
\begin{lstlisting}[style=ceaprompt]
Question:
{question}

Your original local evidence:
{local_evidence}

Your previous response:
{own_previous_response}

Review and revise your reasoning, conclusion, and answer. Return the same JSON schema with exactly three steps, a conclusion, and a short answer.
\end{lstlisting}
\end{minipage}\par\smallskip

\noindent\begin{minipage}{\columnwidth}
\noindent\textbf{P3. Full-MAS update.}
\begin{lstlisting}[style=ceaprompt]
Question:
{question}

Your original local evidence:
{local_evidence}

Your previous response:
{own_previous_response}

Other agents' previous responses:
{peer_previous_responses}

Use the available evidence and peer responses to revise your reasoning, conclusion, and answer. Return the same JSON schema with exactly three steps, a conclusion, and a short answer.
\end{lstlisting}
\end{minipage}\par\smallskip
Each peer response contains its reasoning, conclusion, and answer. The corrected 2Wiki protocol additionally makes the peer-reported local fact available for integration. Peer responses follow the frozen ordering in the corresponding request record.

\subsection{Fact-Level Communication}
Architecture conditions share one initial response containing reasoning, a local fact, and gap queries. Source-identified fact cards remain fixed across updates. The field names below normalize this extended interface.

\noindent\begin{minipage}{\columnwidth}
\noindent\textbf{P4. Shared initial reasoning and fact card.}
\begin{lstlisting}[style=ceaprompt]
Question:
{question}

Your assigned evidence:
{local_evidence}

Produce a compact local fact supported by your evidence and identify information gaps relevant to the question. Also produce exactly three reasoning steps, a conclusion, and a short answer in the same response.

Return these fields:
{"local_fact": "...",
 "gaps": ["..."],
 "steps": ["...", "...", "..."],
 "conclusion": "...",
 "answer": "..."}
\end{lstlisting}
\end{minipage}\par\smallskip
The runtime attaches the sender/source identifier to the emitted fact. It extracts the fact directly from this response. Fact-card generation and initial reasoning therefore share one model call.

\noindent\begin{minipage}{\columnwidth}
\noindent\textbf{P5. Fact-level integration update.}
\begin{lstlisting}[style=ceaprompt]
Question:
{question}

Your current reasoning state:
{current_reasoning_and_answer}

Your current information gaps:
{current_gaps}

Delivered peer facts and source IDs:
{delivered_fact_cards}

Use the delivered facts to address your information gaps and update your reasoning. Return updated gaps, exactly three reasoning steps, a conclusion, and a short answer.
\end{lstlisting}
\end{minipage}\par\smallskip
All fact-level variants share the integration interface. Compact-All delivers both peer cards. Random-K matches fallback's card count, while GAP and Jev supply their selected cards. Treatment-specific serialized memory fields require the original request archive.

\noindent\begin{minipage}{\columnwidth}
\noindent\textbf{P6. Retention-treatment additions.}
The following additions summarize the recorded soft and hard retention semantics. They are applied with the corresponding memory construction and matched self-refinement control in \appref{gap-protocol}.
\begin{lstlisting}[style=ceaprompt]
SOFT RETENTION
Original local evidence:
{local_evidence}

Preserve the supported local content when integrating new peer information.

HARD RETENTION
Persistent local anchor:
{local_anchor}

Previously accepted peer facts:
{accepted_peer_facts}

Preserve the persistent anchor while integrating the accepted peer facts.
\end{lstlisting}
\end{minipage}\par\smallskip
GAP-SOFT re-shows the original source and adds the preservation instruction. Hard GAP-MAS uses the persistent anchor and accepted peer facts in its gate memory, together with the anchored self-refinement control. Their differences include the memory state and evidence presentation described in the protocol.

\subsection{Typed Jev Queries and Delivery Policy}
The Jev controller receives the question, receiver memory, current gap, and candidate peer fact. Three binary queries and one action choice use a common state. The questions below express the recorded decision semantics, with normalized wording. Typed responses follow the Jev interface~\cite{typesafe2026jev,typesafe2026confidence}.

\noindent\begin{minipage}{\columnwidth}
\noindent\textbf{P7. Candidate state and typed queries.}
\begin{lstlisting}[style=ceaprompt]
STATE
Question: {question}
Receiver memory: {receiver_memory}
Current gap: {gap}
Candidate peer fact: {candidate_fact}

NOUL: RELEVANCE
Is this fact relevant to the current gap?
NOUL: NOVELTY
Does it add information absent from memory?
NOUL: CONFLICT
Does it contradict receiver memory?
CHOICE: ACTION
Select an action using the state above.
Options: SEND, SKIP, QUARANTINE
\end{lstlisting}
\end{minipage}\par\smallskip
SEND selects useful information for delivery, SKIP excludes the candidate, and QUARANTINE isolates conflicting information. Choice supplies the action confidence used by the fallback policy. Noul supplies a probability. The separately logged binary certainty $|2p-1|$ is a diagnostic. These typed queries are distinct from the receiving LLM's reasoning-update prompt.

\noindent\textbf{Runtime selection and fallback.}
Candidate selection, deduplication, the two-card cap, and fallback expansion are programmatic operations. Fallback broadens sharing when no SEND candidate meets the action-confidence threshold $\theta$, recorded as $0.70$ in the run protocol. GAP uses a frozen NLI classifier, checking contradiction before duplicate entailment. These operations populate P5 using reference-free runtime inputs.

\noindent\textbf{Reproduction record.}
Each populated template should be released with model/provider identifiers, evidence allocations, prior-round dependencies, request parameters, parsers, and treatment-specific memory fields. Shared question, seed, and agent identifiers link the matched traces and resource logs.

\end{document}